\documentclass[pdflatex,sn-mathphys-num,referee]{sn-jnl}

\usepackage{graphicx}%

\usepackage{float}

\usepackage{multirow}%
\usepackage{amsmath,amssymb,amsfonts}%
\usepackage{amsthm}%
\usepackage{mathrsfs}%
\usepackage[title]{appendix}%
\usepackage{xcolor}%
\usepackage{textcomp}%
\usepackage{manyfoot}%
\usepackage{booktabs}%
\usepackage{caption}
\usepackage{subcaption}
\usepackage{algorithm}%
\usepackage{algorithmicx}%
\usepackage{algpseudocode}%
\usepackage{listings}%

\usepackage{caption}
\usepackage{subcaption}

\UseRawInputEncoding

\theoremstyle{thmstyleone}%
\theoremstyle{thmstyletwo}%

\theoremstyle{thmstylethree}%

\newcommand{\projectname}{StormNet\ignorespaces}
\begin{document}

\title[Article Title]{\projectname{}: Improving storm surge predictions with a GNN-based spatio-temporal offset forecasting model}


\author*[1]{\fnm{Noujoud} \sur{Nader} \orcid{https://orcid.org/0009-0000-4687-1416}
}\email{nnader@lsu.edu}\phone{+1-509-957-5330}
\equalcont{These authors contributed equally to this work.}

\author[2]{\fnm{Stefanos} \sur{Giaremis} \orcid{https://orcid.org/0000-0002-0107-3127} }\email{sgiaremi@physics.auth.gr}
\equalcont{These authors contributed equally to this work.}

\author[3]{\fnm{Clint} \sur{Dawson} \orcid{https://orcid.org/0000-0001-7273-0684} }\email{clint@oden.utexas.edu}

\author[1]{\fnm{Carola} \sur{Kaiser} }\email{ckaiser@cct.lsu.edu}

\author[1]{\fnm{Karame} \sur{Mohammadiporshokooh} \orcid{https://orcid.org/0009-0000-8349-3389} }\email{kmoham6@lsu.edu}

\author[1]{\fnm{Hartmut} \sur{Kaiser} \orcid{https://orcid.org/0000-0002-8712-2806} }\email{hkaiser@cct.lsu.edu}

\affil[1]{\orgdiv{Center for Computation and Technology}, \orgname{Louisiana State University}, \orgaddress{\city{Baton Rouge}, \postcode{70803}, \state{LA}, \country{US}}}

\affil[2]{\orgdiv{Department of Physics}, \orgname{Aristotle University of Thessaloniki}, \orgaddress{\city{Thessaloniki}, \postcode{54124}, \country{Greece}}}

\affil[3]{\orgdiv{Oden Institute for Computational Engineering and Sciences}, \orgname{The University of Texas at Austin}, \orgaddress{\city{Austin}, \postcode{78712}, \state{TX}, \country{US}}}


\abstract{Storm surge forecasting remains a critical challenge in mitigating the impacts of tropical cyclones on coastal regions, particularly given recent trends of rapid intensification and increasing nearshore storm activity. Traditional high fidelity numerical models such as ADCIRC, while robust, are often hindered by inevitable uncertainties arising from various sources. To address these challenges, this study introduces StormNet, a spatio-temporal graph neural network (GNN) designed for bias correction of storm surge forecasts. StormNet integrates graph convolutional (GCN) and graph attention (GAT) mechanisms with long short-term memory (LSTM) components to capture complex spatial and temporal dependencies among water-level gauge stations. The model was trained using historical hurricane data from the U.S. Gulf Coast and evaluated on Hurricane Idalia (2023). Results demonstrate that StormNet can effectively reduce the root mean square error (RMSE) in water-level predictions by more than 70\% for 48-hour forecasts and above 50\% for 72-hour forecasts, as well as outperform a sequential LSTM baseline, particularly for longer prediction horizons. The model also exhibits low training time, enhancing its applicability in real-time operational forecasting systems. Overall, StormNet provides a computationally efficient and physically meaningful framework for improving storm surge prediction accuracy and reliability during extreme weather events.}




\maketitle

\newpage
\section*{Highlights}
\begin{itemize}
    \item StormNet, a spatio-temporal GNN combining GCN, GAT, and LSTM, is proposed for storm surge bias correction.
    \item  Graph nodes represent gauge stations; edges are defined by high water-level correlation and proximity, enabling physically meaningful connectivity.
    \item StormNet reduces RMSE by $>$70\% for 48-hour and $>$50\% for 72-hour storm surge forecasts over ADCIRC baselines.
    \item Low training cost and real-time compatibility make StormNet suitable for operational forecasting systems such as CERA.
\end{itemize}

\newpage
\section*{Graphical Abstract}
\begin{figure}[h]
    \centering
    \includegraphics[width=1\linewidth]{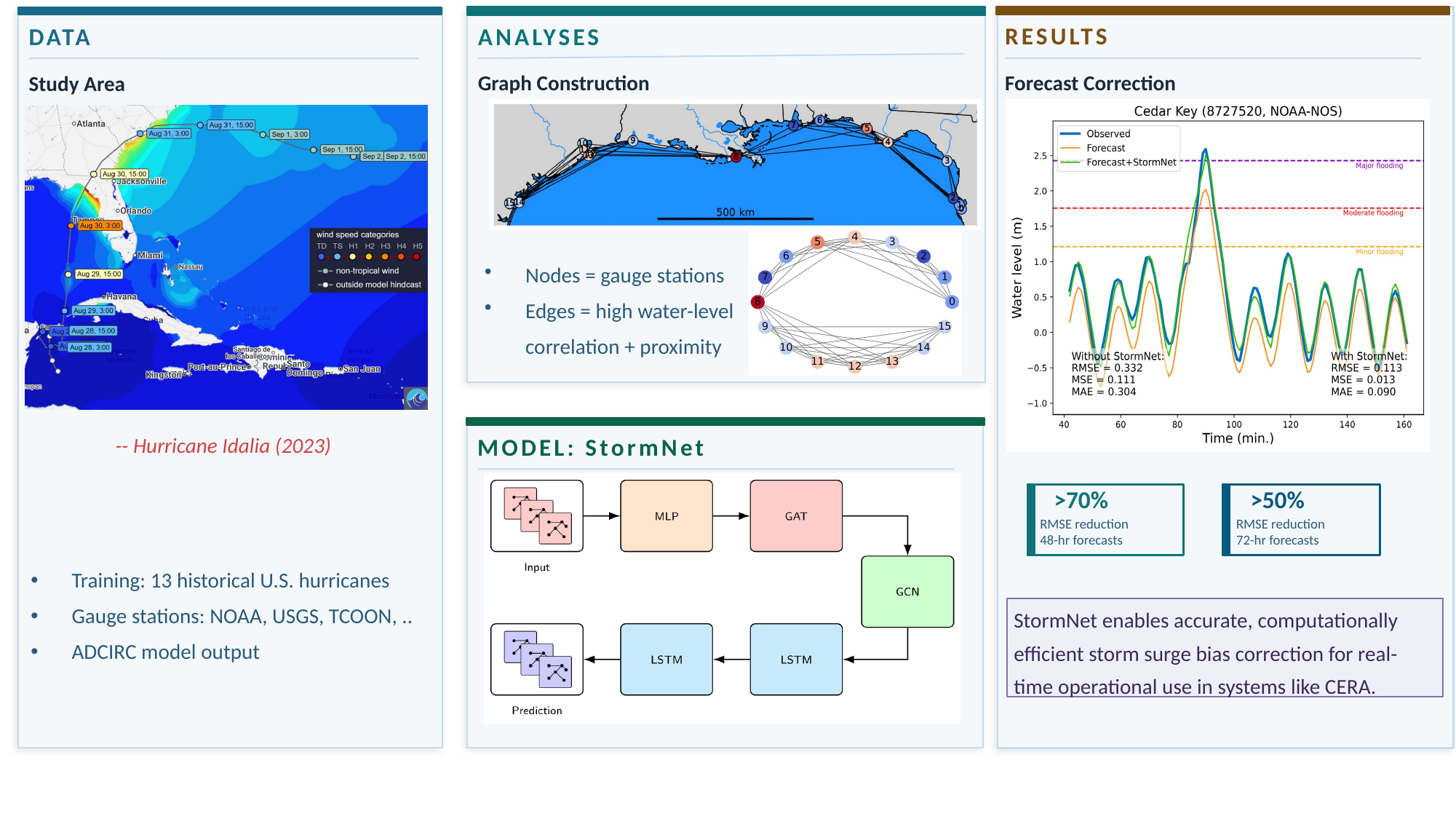}
    \caption*{}
    \label{fig:placeholder}
\end{figure}
\textbf{Graphical Abstract Description:} Based on the graphical abstract, this study develops and evaluates \textbf{StormNet}, a spatio-temporal graph neural network framework designed for bias correction of storm surge forecasts along the U.S. Gulf Coast. This work captures the complex spatial and temporal dependencies among coastal water-level gauge stations and leverages them to systematically reduce forecast errors produced by the high-fidelity ADCIRC physics-based model. Observed water level data were collected from multiple gauge stations, and forecast water levels were obtained from ADCIRC. Forecast errors were estimated and subsequently used as training data. Data from 13 historical hurricane events were utilized for model training, while Hurricane Idalia (2023) was reserved as an independent test case to assess generalization performance. The graph construction strategy builds $\mathcal{G}(\mathcal{V}, \mathcal{E})$, where nodes $\mathcal{V}$ represent gauge stations and edges $\mathcal{E}$ are defined based on pairwise water-level correlation and geographic proximity between stations. StormNet integrates multilayer perceptron (MLP), graph attention network (GAT), graph convolutional network (GCN), and long short-term memory (LSTM) components to jointly capture spatial correlations and temporal dynamics across the gauge network. Model performance was evaluated using root mean square error (RMSE), mean square error (MSE), and mean absolute error (MAE) metrics. Results demonstrate that StormNet reduces RMSE by more than 70\% for 48-hour forecasts and more than 50\% for 72-hour forecasts relative to uncorrected ADCIRC output, while outperforming a sequential LSTM baseline, particularly at extended prediction horizons. The framework exhibits low training cost and is designed for seamless integration into real-time operational systems such as the Coastal Emergency Risks Assessment (CERA) platform. The findings offer critical insights for coastal flood risk managers and emergency response stakeholders seeking computationally efficient tools for improving the accuracy and reliability of storm surge predictions during extreme weather events.

\newpage
\section{Introduction}
The likelihood and severity of intense tropical cyclones (TC) and extreme storm surge events have been shown to be quite susceptible to climate variability and follow increasing trends over the last few decades~\cite{NCEI, Calafat2022}. Furthermore, the behavior of TC is becoming more detrimental to coastal regions. Specifically, the maximum intensity of TC has been found to shift towards shorelines by 30 km per year, and the frequency of nearshore TC has been increased by factor of 2, over the period 1982–2018~\cite{Wang2021}. TCs are also found to be more likely to stall in coastal regions, cause increased rainfall, and decay more slowly over land~\cite{Hall2019, Patricola2018, Li2020}. The rapid intensification (RI) of TC shortly before landfall is another increasingly prevalent phenomenon, threatening coastal communities worldwide~\cite{Balaguru2024, Li2023}. RI has been a major characteristic of many severe TCs, with recent examples being Hurricanes Idalia (2023)~\cite{shi2025intensification}, Ian (2022)~\cite{Liu2024} and Ida (2021)~\cite{Zhu2022} in the United States (U.S.), Cyclone Mocha (2023) in the Bay of Bengal ~\cite{Kotal2024}, Typhoon Rai (2021) in Philippines~\cite{Petilla2025} and Typhoon Hato (2017) in the South China Sea~\cite{Zhang2019}. RI has been attributed and/or correlated to circulation events within the TC, such as warm-core eddies in the inner core, eyewall mesovortices and convective bursts, and their interplay with larger-scale environmental conditions, such as vertical wind shear, sea surface temperature, moisture, stratospheric gravity wave activity, sea spray-induced heat fluxes and river plumes~\cite{Chang2017, Wu2015, Zagrodnik2014, Wu2025, Yang2024, shi2025intensification, Kim2024}. TC intensification rate has been found to increase in coastal regions in the last four decades, due to increasing relative humidity and decreasing vertical wind shear in such regions, compared to the open ocean~\cite{Balaguru2024}. Consequently, while forecasting accuracy has increased in terms of predicting TC paths over the last years, the presence of events with such rapid intensity changes and the complexity of the multiscale underlying processes, pose a significant challenge for predicting TC intensity~\cite{Cangialosi2020, Trabing2020, Cyriac2018}. 

From the above, it is evident that the intensification of TC-related events requires augmented efforts towards the development and improvement of operational forecasting frameworks and early warning systems for hazard mitigation, emergency planning, and quantifying the impacts of extreme weather events on coastal communities. Storm surge models, based on numerical solutions of hydrodynamic equations and forced by meteorological inputs, constitute the cornerstone of such frameworks~\cite{Turner2024, Penny2023, Suh2015}. The advanced circulation model for oceanic, coastal and estuarine waters (ADCIRC)~\cite{Westerink1992} is among the most broadly used storm surge models, and it is employed by U.S. federal agencies, academic institutions, and private organizations for both research and operational purposes. It is commonly used in tight coupling with SWAN, a third-generation spectral wave model, with the two models sharing the same triangular mesh and exchanging water depth, current, and wave radiation stress information~\cite{Booij1999, Dietrich2011}. The Coastal Emergency Risk Assessment (CERA) framework is an interactive online visualization platform that combines real-time water level measurements with numerical predictions based on ADCIRC~\cite{CERA}. It is designed to provide first responders, decision-makers, and the general public with critical insights during hurricanes and other extreme weather events. The aforementioned intensification of extreme weather events has motivated numerous and continuous developments in storm surge prediction and response frameworks such as the above, for instance in terms of improving mesh design, computational efficiency and descriptions of natural processes, facilitating model comparisons, enabling the coupling of multiscale approaches, using storm surge model data for traffic and risk assessment, and others~\cite{Turner2024, Chen2025, Pringle2021, Khani2023, Blakely2022, Loveland2024, Dawson2024, Bernier2024, Loveland2021, Wei2024, Zhang2023, Huang2021}.

Nevertheless, inherent uncertainties remain inevitable in storm surge models. These primarily arise from errors in the specification of hurricane characteristics, such as track, wind speed, and size, and inaccuracies in input data, including coastal topography, bathymetry, and land cover representation. Additional sources of bias include inaccuracies in atmospheric forcing from parent meteorological models, simplified physical parameterizations like bottom friction, and the use of coarse-resolution topographic and bathymetric data that fail to capture critical coastal features~\cite{Ozkan2025, Munoz2022, Torres2019, Gallien2018, Ferreira2014}. Moreover, physical drivers such as rainfall-runoff processes, large-scale oceanic circulations, and hydrological inputs are often omitted to increase computational performance or stability~\cite{Asher2019}. The combined effect of such uncertainties has been shown to introduce substantial biases that impact the accuracy of storm surge predictions~\cite{Gonzalez2019, Resio2012}. This is crucial for the design and implementation of response and prevention strategies, as for instance, regarding water levels, only 0.3 to 0.7 m have been shown to separate minor, high-tide flooding, from severe and destructive floods~\cite{Sweet2018}. Typical methods to quantify and treat uncertainties and biases in storm surge models in the context of either operational forecasting or research purposes include probabilistic methods, data assimilation and other statistical approaches~\cite{Feng2023, Munoz2022, Asher2019, Resio2017, Butler2012}. Furthermore, significant efforts have been made for the systematic collection of data from historical measurements and hindcasts in curated datasets (e.g., \cite{Muis2016, Tadesse2021, CERA2023}), which is crucial for extracting trends, analyzing model accuracy, and enabling data-driven forecasting and bias correction approaches.

The vast and rapid developments in the field of machine learning (ML) in the last decade have also motivated numerous efforts for the implementation of ML algorithms for enhancing forecasting capabilities. Such models are typically trained on historical, hindcast and/or synthetic data, motivating efforts for global datasets, such as the Global Extreme Sea Level Analysis (GESLA)~\cite{Haigh2022}, the European Centre of Medium-Range Weather Forecasts Reanalysis (ERA)~\cite{Soci2024}, and the Coastal Dataset for the Evaluation of Climate Impact (CoDEC)~\cite{Muis2020}, but also TC- and location-specific data collections, such as the HURDAT dataset by the National Oceanic and Atmospheric Administration (NOAA)~\cite{noaaHURDATReanalysis}, the Historical Storm Archive by CERA~\cite{CERAarchive, CERA2023}, the Western North Pacific Tropical Cyclone Database~\cite{Lu2021}, and the Digital Typhoon dataset~\cite{Kitamoto2023}. ML models for enhancing forecasting accuracy can be generally split into two main categories: surrogate and bias correction models. Surrogate models are trained to reproduce the outputs of high-fidelity but computationally demanding physics-based solvers, promising comparable accuracy at a significantly lower computational cost (at least regarding the inference part). Recent examples of ML surrogate models in the context of TC forecasting include the use of models such as artificial neural networks (ANN), convolutional neural networks (CNN), hierarchical deep neural networks (HDNN), long short-term memory (LSTM) and convolutional LSTM (ConvLSTM) networks, vision transformers (ViT) and physics-informed neural networks (PINN), for predicting quantities such as storm surge, TC intensity, tidal levels, rainfall, and winds~\cite{Zhao2025, Han2025, SavizNaeini2025, Zhu2025, Sreeraj2025, Kim2024, Huang2024, Shi2024, Pachev2023, Dotse2023}. On the other hand, bias correction models aim to mitigate the systemic errors of physics-based models by learning and eliminating their deviations from target (usually observational) data. This approach has been followed by implementing similar ML algorithms such as the ones used for surrogate models to predict biases in forecasting storm surge, significant wave height, tides, wind, temperature, humidity, precipitation and other relevant quantities~\cite{Wei2025, Li2025, Cerrone2025, CarneiroBarros2025, giaremis2024storm, Tedesco2024, Liao2024, Zhang2024, Zhang2024b, Kao2024, Liu2023}.

Graph Neural Networks (GNN) constitute an emerging route for modeling physical processes with ML, as graph representations using nodes and edges can be often corresponded to elements of the system in hand and reflect its properties in the model in a more meaningful and controlled manner. Additionally, spatiotemporal GNNs (ST-GNN) leverage the strengths of approaches such as convolution and attention networks (with the Spatiotemporal Graph Convolution Networks, ST-GCN~\cite{Yan2018}, and ATtention networks, ST-GAT architectures~\cite{Zhang2019b}), for simultaneously learning both spatial and temporal correlations. This approach has been explored for surrogate models for storm surge and flood predictions (e.g., \cite{Jiang2024, Kazadi2024}), while similar GNN-based architectures, based on an encoder-processor-decoder scheme, have been shown positive results for global weather forecasting~\cite{Lam2023}. While a similar approach has been also reported for bias correcting numerical forecasts in the same context~\cite{Wu2023}, the implementation of this family of methods for bias correcting  storm-specific scenarios has not been yet widely explored. Thus, in this work, we demonstrate the performance of \projectname{}, a GNN-based approach for bias correcting storm surge forecasts. We employ our model for bias correcting ADCIRC-based predictions during storm events in the U.S. Gulf Coast and we compare the results with our previous LSTM-based model~\cite{giaremis2024storm}. 

\paragraph{Contributions.}
The main contributions of this work are summarized as follows:

\begin{itemize}
    \item We propose \projectname{}, a spatio-temporal graph neural network architecture that integrates GCN, GAT, and LSTM components to jointly capture spatial correlations and temporal dynamics for storm surge bias correction. 

    \item We introduce a physically informed graph construction strategy that leverages inter-station correlations and coastal geometry to encode meaningful spatial relationships between gauge locations, enabling better spatio-temporal learning.

    \item We develop an end-to-end forecasting framework that predicts future model biases and applies them to correct ADCIRC water-level forecasts. This hybrid physics–ML approach leads to accuracy improvements, especially for longer (48–72 hour) prediction horizons.

    \item We conduct a comprehensive evaluation using real hurricane events from the U.S. Gulf Coast, including an independent test on Hurricane Idalia (2023).

    \item We perform an ablation analysis to quantify the contribution of each architectural component (GCN, GAT, LSTM) and demonstrate the robustness of the model, low training cost, and suitability for a real-time operational forecasting system (CERA).
\end{itemize}

The paper is organized as follows: Data acquisition, preprocessing, and GNN model architecture, along with its underlying principles, are presented in Section 2. The results of applying \projectname{} in the considered region of interest, along with a comparison with the LSTM-based model of \cite{giaremis2024storm}, and the results of the ablation study are shown in Section 3. Finally, the concluding remarks and the limitations of this approach are included in Section 4.

\section{Methodology}
\subsection{Overview}
\label{sec:Overview}
\begin{figure}
    \centering
    \includegraphics[width=1\linewidth]{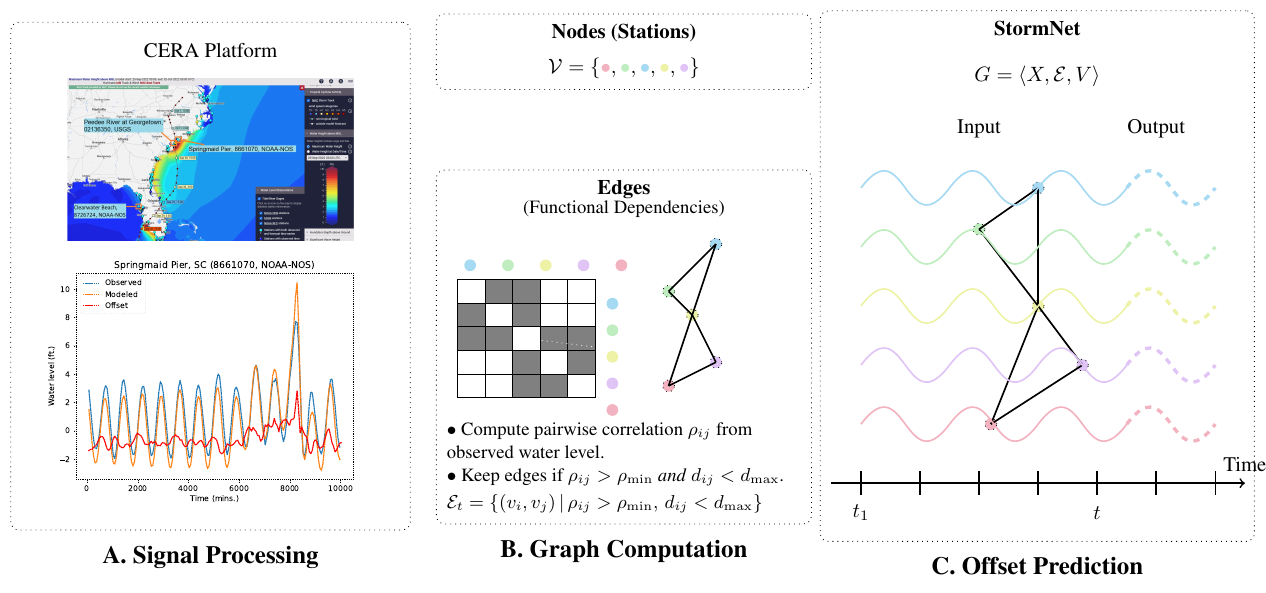}
    \caption{Overview of the proposed spatio-temporal learning pipeline for bias correction in hurricane storm surge forecasting \textbf{\projectname{}}. 
(A)~Data preprocessing: include computation  of offsets between modeled and observed water levels at each gauge station. 
(B)~Graph computation: construction of the spatio-temporal graph $\mathcal{G} = (\mathcal{V}, \mathcal{E})$, where nodes represent stations and edges represent correlations betweeen stations. 
(C) Offset prediction: the proposed \textit{StormNet} model forecasts future offsets $\hat{o}_i(t)$, which are subsequently used to correct the physics-based (ADCIRC) modeled water levels.
}
    \label{fig:pipeline}
\end{figure}

The proposed pipeline integrates spatial and temporal learning components to correct biases in physics-based hurricane storm surge predictions, as illustrated in Figure \ref{fig:pipeline}. It comprises three key steps, namely data preprocessing, graph computation, and offset prediction. The first stage (Figure~\ref{fig:pipeline}.A) involves the systematic extraction of the offsets (Eq. \ref{eq:offsets}) between the modeled and observed water level time series from each gauge station in the available dataset. The offset time series ($o_i$) for each gauge station $i$ is defined as follows: 
\begin{equation}
    o_i(t) = x_i^{\text{modeled}}(t) - x_i(t)
    \label{eq:offsets}
\end{equation}
where $x_i^{\text{modeled}}(t)$ and $x_i(t)$ denote the forecast (via ADCIRC) and observed (from gauge stations) water levels, respectively, and $o_i(t)$ is the water level offset (i.e., the bias), at each timestep $t$ for each station $i$. More details on this phase are explained in Section \ref{ssec:data-acquisition}. Given the graph-like nature of our problem, the data can be reformulated as a graph (Stage~2, Figure~\ref{fig:pipeline}.B). 
A graph $\mathcal{G}$ is an abstract representation of a network, consisting of a set of nodes $\mathcal{V}$ and edges $\mathcal{E}$ that indicate the presence of interactions between nodes~\cite{nader2015classification, nader2016node, al2015detecting}. 
In our formulation, each node $v_i \in \mathcal{V}$ corresponds to a gauge station and each edge $(v_i, v_j) \in \mathcal{E}$ represents the correlation between stations $i$ and $j$. 
Further details about this stage are described in Section~\ref{ssec:graph-construction}. Finally, in the third stage, we employ the proposed \projectname{} model to forecast offsets (biases) at all stations. 
These predicted offsets are then used to correct the physics-based (ADCIRC) modeled water levels, thereby improving the accuracy and reliability of storm surge forecasts within CERA platform.

\subsection{Data Acquisition and Preprocessing}
\label{ssec:data-acquisition}

Observed and modeled water elevation time series data at an hourly resolution from 13 historical U.S. hurricanes (Table \ref{tbl:hurricanes}) are extracted from the Historical Storm Archive by CERA~\cite{CERAarchive, CERA2023} and visualized with the online interactive CERA framework~\cite{CERA}. Gauge station data from NOAA, U.S. Geological Survey, Texas Coastal Ocean Observation Network and other U.S. agencies were used as observed values, while modeled water elevation for each gauge station location was produced by ADCIRC~\cite{Westerink1992}. The offset timeseries (Eq. \ref{eq:offsets}) is calculated by the difference between the modeled values from their observed counterparts at each time interval for each station and TC. Offset values outside the range of three standard deviations (calculated on the whole dataset) were eliminated. 

Our study focuses on the Gulf Coast of the U.S., due to its high susceptibility to hurricane events; therefore, only data from gauge stations in this region are included. Furthermore, as explained in more detail in the following sections, offset time series for all the considered stations are treated in parallel by the ML model. Thus, it is necessary to construct a single time series for each station by concatenating the data for all the considered hurricanes. The resulting time series for all stations must also have the same length. As a result, stations with a large amount of missing data for at least one of the considered storms are excluded from the dataset. This procedure leads to the selection of 16 gauge stations, listed in Table~\ref{tbl:gauge_stations}.
Out of the 13 total TCs, data for Hurricanes Ian (2022) and Idalia (2023) are separated from the full dataset and used as validation and test sets, respectively, while the remaining TC data constitute the training set. All sets were further divided into windows using the windowing technique, a common method in time series forecasting. This technique converts continuous time series data into a supervised learning problem by creating input-output pairs. It involves defining a fixed-size window that slides through the data, forming input sequences from a set of consecutive past values ("input window"), and their corresponding target values ("prediction window"). This approach is effective for capturing temporal patterns and dependencies in the data, making it a useful strategy for time series forecasting~\cite{Tedesco2024, giaremis2024storm}.
To address the issue of disparate data scales, data are transformed using the \texttt{MinMax} scaler of the Scikit-Learn Python package~\cite{pedregosa2011scikit}. Training, validation, and test sets are rescaled separately, based on the attributes of the training set, to avoid data leakage~\cite{geron2022hands}.

\begin{table}[h]
\caption{Hurricanes considered in this study (in chronological order), their category based on the Saffir-Simpson hurricane scale~\cite{SaffirSimpson} (H1-H5: Hurricane Category 1-5, TS: Tropical Storm) and the total amount of hourly data in each.}
\label{tbl:hurricanes}
\begin{tabular}{@{}lcc@{}}
\toprule
Hurricane (Year) & Category & No. of hourly data points \\ \midrule
Charley (2004)   & H4       & 2304                      \\
Dennis (2005)    & H4       & 3056                      \\
Wilma (2005)     & H5       & 2974                      \\
Debby (2012)     & TS       & 1536                      \\
Hermine (2016)   & H1       & 2768                      \\
Michael (2018)   & H5       & 2384                      \\
Eta (2020)       & H4       & 2672                      \\
Fred (2021)      & TS       & 3056                      \\
Idalia (2023)    & H4       & 2688                      \\
Ian (2022)       & H5       & 2288                      \\
Debby (2024)     & H1       & 2861                      \\
Helene (2024)    & H4       & 2208                      \\
Milton (2024)    & H5       & 2208                      \\ \bottomrule
\end{tabular}
\end{table}

\subsection{Graph-Based Representation of Gauge Stations}
\label{ssec:graph-construction}
To capture the spatial dependencies among water‐level gauge stations, we model the entire network of stations as an undirected graph. A graph ($\mathcal{G}$) is an
abstract representation of a network, consisting of a set of
nodes ($\mathcal{V}$) and edges ($\mathcal{E}$), indicating the presence of an
interaction between nodes~\cite{nader2015classification, nader2016node, al2015detecting}. For our problem, we present the graph as
\(\mathcal{G} = (\mathcal{V}, \mathcal{E})\), where each node \(v_i \in \mathcal{V}\) corresponds to a gauge station and each edge \((v_i, v_j)\in\mathcal{E}\) represents a strong correlation between stations \(i\) and \(j\).

\paragraph{Node features.}  
Each node \(v_i\) is associated with a time series 
\[
\mathbf{x}_i = \bigl[x_i(1),\,x_i(2),\,\dots,\,x_i(T)\bigr]^\top,
\]
where \(x_i(t)\) is the observed water level at station \(i\) and time \(t\).  

\paragraph{Edge construction.}  
We place an undirected edge between nodes \(i\) and \(j\) if and only if both
\[
\rho_{ij} \;>\;\rho_{\min}
\quad\text{and}\quad
d_{ij}\;<\;d_{\max},
\]
where
\[
\rho_{ij}
=
\frac{\sum_{t=1}^T \bigl(x_i(t)-\bar{x}_i\bigr)\,\bigl(x_j(t)-\bar{x}_j\bigr)}
     {\sqrt{\sum_{t=1}^T \bigl(x_i(t)-\bar{x}_i\bigr)^2}\;\sqrt{\sum_{t=1}^T \bigl(x_j(t)-\bar{x}_j\bigr)^2}}
\]
is the Pearson correlation between the two observed water‐level series, and
\[
d_{ij}
=\;
\mathrm{haversine}\bigl((\phi_i,\lambda_i),\,(\phi_j,\lambda_j)\bigr)
\]
is the great‐circle distance between station locations where each station has geographic coordinates \((\phi_i,\lambda_i)\).  

\paragraph{Adjacency matrix.}  
From the edge set \(\mathcal{E}\), we build the weighted adjacency matrix \(A\in\mathbb{R}^{N\times N}\):
\[
A_{ij}
=
\begin{cases}
\rho_{ij}, & (v_i,v_j)\in\mathcal{E},\\
0,          & \text{otherwise.}
\end{cases}
\]
\begin{table}[h!]
\centering
\caption{Symbol meaning.}
\begin{tabular}{@{}lll@{}}
\toprule
\textbf{Type} & \textbf{Symbols} & \textbf{Meaning} \\ \midrule
\textbf{Graph Related} & $\mathcal{G}$ & Graph structure \\
                       & $\mathcal{V}$ & Vertices \\
                       & $\mathcal{E}$ & Edges \\
                       & $A$ & Adjacency matrix \\
\textbf{Data Related}  & $x_i$ & Observed water-level signal at gauge station $i$ \\
                       &$x_i^{modeled}$ & Modeled water-level (via ADCIRC) signal at gauge station $i$ \\
\bottomrule
\end{tabular}
\end{table}
\subsection{Graph neural network}
Our proposed model integrates GCN and Graph Attention Networks (GAT) to capture both global topological dependencies and adaptive local feature importance within the graph. The overall architecture is described in detail in Section \ref{sec:proposed_model}.
\subsubsection{Graph Convolution Network (GCN)}
 The GCN layer aggregates feature information from a node’s local neighborhood through a normalized summation. It captures global structural information efficiently through a spectral-based convolution. It can be presented  as follows: \begin{equation}
h_i^{l+1} = \sigma \left( \sum_{j \in \mathcal{N}(i)} \frac{1}{c_{ij}} \phi^l h_j^{l} \right),
\end{equation}
 where $\mathcal{N}(i)$ denotes the set of nodes directly connected to node $i$, $\sigma$ is the activation function.
\subsubsection{Graph Attention Network (GAT)}
Following a self-attention strategy~\cite{vaswani2017attention},
GAT learns the hidden features of each node by iteratively
using node features to compute similarities. The GAT layer extends GCNs by applying an attention mechanism that learns to assign different weights to neighboring nodes, allowing the model to focus on more relevant features.

At this stage, each node $i$ is represented by a feature vector $h_i^l \in \mathbb{R}^F$, all node features can be expressed as $h^l = \{h_1^l, h_2^l, \ldots, h_N^l\}$, where $N$ and $F$ denote the total number of nodes and features, respectively. 
To project these input features into a higher-dimensional latent space, a shared linear transformation is applied using the weight matrix $W \in \mathbb{R}^{F' \times F}$. 
The transformed features are then used to compute pairwise attention scores between the connected nodes. 
The attention coefficient $e_{ij}$ between the nodes $i$ and $j$ is obtained as:

\begin{equation}
e_{ij} = a\!\left( W h_i^l, W h_j^l \right),
\quad a : \mathbb{R}^{F'} \times \mathbb{R}^{F'} \rightarrow \mathbb{R},
\label{eq:eij}
\end{equation}

where $a(\cdot,\cdot)$ denotes the attention mechanism and $e_{ij}$ quantifies the importance of node $j$'s features to node $i$. 
To preserve the local topology of the graph, attention scores are computed only between each node and its direct neighbors. 
Next, a LeakyReLU activation is applied, and the resulting attention coefficients are normalized using the softmax function to ensure comparability across neighboring nodes:

\begin{equation}
\alpha_{ij} = \text{softmax}\!\left(\text{LeakyReLU}(e_{ij})\right).
\label{eq:alphaij}
\end{equation}

Finally, these normalized attention weights are employed to aggregate information from neighboring nodes and update the node features according to:

\begin{equation}
h_i^{l+1} = \sigma\!\left( \sum_{j \in \mathcal{N}(i)} \alpha_{ij}\,W h_j^l \right),
\label{eq:h_update}
\end{equation}

where $\sigma(\cdot)$ represents a non-linear activation function (e.g., ReLU), and $\mathcal{N}(i)$ indicates the neighbors of node $i$.

\noindent\textbf{Multi-Head Attention Mechanism.} To enhance the expressiveness and stability of the self-attention process, a multi-head attention mechanism is employed. 
This mechanism allows the model to learn attention coefficients from multiple independent representation subspaces, thereby capturing diverse relational patterns between nodes. 
Formally, $K$ independent attention heads are used, each performing its own attention transformation and feature aggregation as described in Equations~\eqref{eq:eij}–\eqref{eq:h_update}. 
The resulting feature maps from all heads are then either concatenated or averaged to form the final node representation.

When concatenation is used, the updated node representation is given by:

\begin{equation}
h_i^{l+1} = \bigg\Vert_{K=1}^{K} 
\sigma \left( \sum_{j \in \mathcal{N}(i)} 
\alpha_{ij}^{K} \, W^{K} h_j^{l} \right),
\label{eq:multihead_concat}
\end{equation}

where $\Vert$ denotes the concatenation operation across all $K$ attention heads.  
Alternatively, when averaging is applied, the multi-head outputs are aggregated as:

\begin{equation}
h_i^{l+1} = 
\sigma \left( 
\frac{1}{K} \sum_{K=1}^{K} 
\sum_{j \in \mathcal{N}(i)} 
\alpha_{ij}^{K} \, W^{K} h_j^{l} 
\right).
\label{eq:multihead_avg}
\end{equation}

The concatenation strategy is typically adopted in intermediate layers to enrich the feature diversity, while averaging is often used in the final layer to stabilize the output and reduce variance across attention heads.
\begin{figure}
    \centering
    \includegraphics[width=0.5\linewidth]{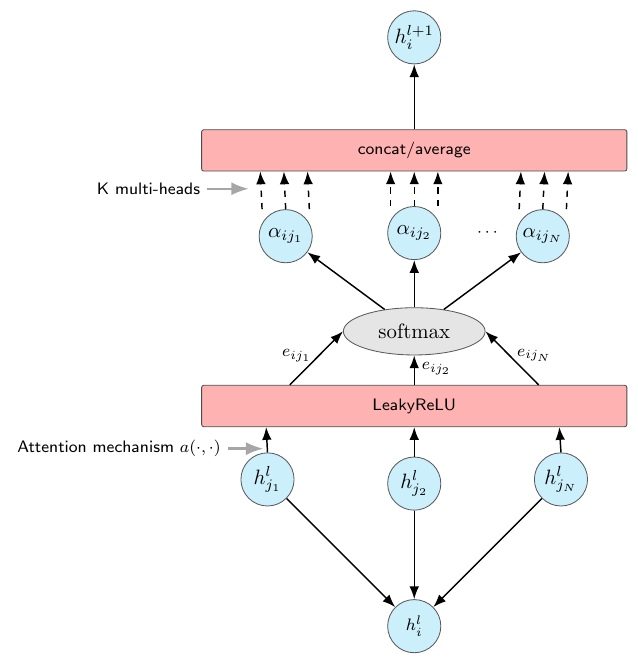}
    \caption{A graph attentional layer with multi-head attention
mechanism, involving $K$ heads. $N$ denotes the number of nodes
connected to node $i$ .}
    \label{fig:gat}
\end{figure}

\subsection{Proposed Model: \projectname} \label{sec:proposed_model}
\begin{figure}
    \centering
    \includegraphics[width=0.8\linewidth]{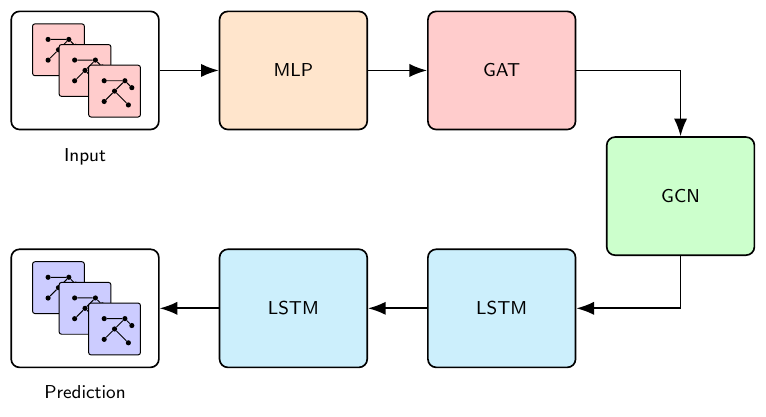}
    \caption{Architecture of the proposed model \projectname{}}
    \label{fig:StormNet}
\end{figure}

The proposed \projectname{} architecture integrates spatial and temporal learning mechanisms into a unified framework to forecast future offsets $\hat{o}_i(t)$, which are subsequently used to correct the physics-based (ADCIRC) storm surge predictions. \projectname{} combines multiple graph-based and sequence-based components to exploit both inter-station dependencies and temporal dynamics in coastal water-level time series. As illustrated in Figure~\ref{fig:StormNet}, the model operates in three main phases. 
First, the input signals, including historical offsets from each station, are preprocessed through a lightweight multilayer perceptron (MLP). 
Next, \projectname{} processes the data through two spatial encoders. The first is a GCN, which captures broad spatial correlations by aggregating information from neighboring stations according to the graph structure. The second is GAT, which applies an attention mechanism to adaptively weight the importance of each neighbor. While the GCN extracts general spatial patterns, the GAT refines these features by emphasizing the most relevant inter-station relationships during storm events. After that, we use two LSTM layers that are commonly leveraged to learn temporal dependencies
and time-series prediction. The final output consists of predicted offsets $\hat{o}$ in future per station.

\subsubsection{Experimental Details}
\projectname{} was trained using the Adam optimizer \cite{Kingma2014} for 200 epochs. The initial learning rate is set to $3e^{-5}$, and the weight decay is $5e^{-7}$ with a batch size of 20. Regarding the past time window, by explicitly examining values in the range of 6 to 48 h, we found the lower error values were obtained with past window values in the range of 42 to 48 hours (see Supplementary Information, Table 1). Training and inference were performed on a single NVIDIA A100 graphical processing unit.

We use Root Mean Square Error (RMSE), Mean Squared Error (MSE), and Mean Absolute Error (MAE) to evaluate the performance of the model, which are defined as follows:
\begin{equation}
\text{RMSE} = \sqrt{\frac{1}{n} \sum_{i=1}^{n} (o_i - \hat{o}_i)^2}
\end{equation}

\begin{equation}
\text{MSE} = \frac{1}{n} \sum_{i=1}^{n} (o_i - \hat{o}_i)^2
\end{equation}

\begin{equation}
\text{MAE} = \frac{1}{n} \sum_{i=1}^{n} \lvert o_i - \hat{o}_i \rvert
\end{equation}

where $o_i$ denotes the true offset at time $i$, and $\hat{o}_i$ denotes the predicted offset at time $i$.

\section{Results and Discussion}
\begin{figure}[H]
\centering
\begin{subfigure}[c]{0.9\textwidth}
    \includegraphics[width=\textwidth]{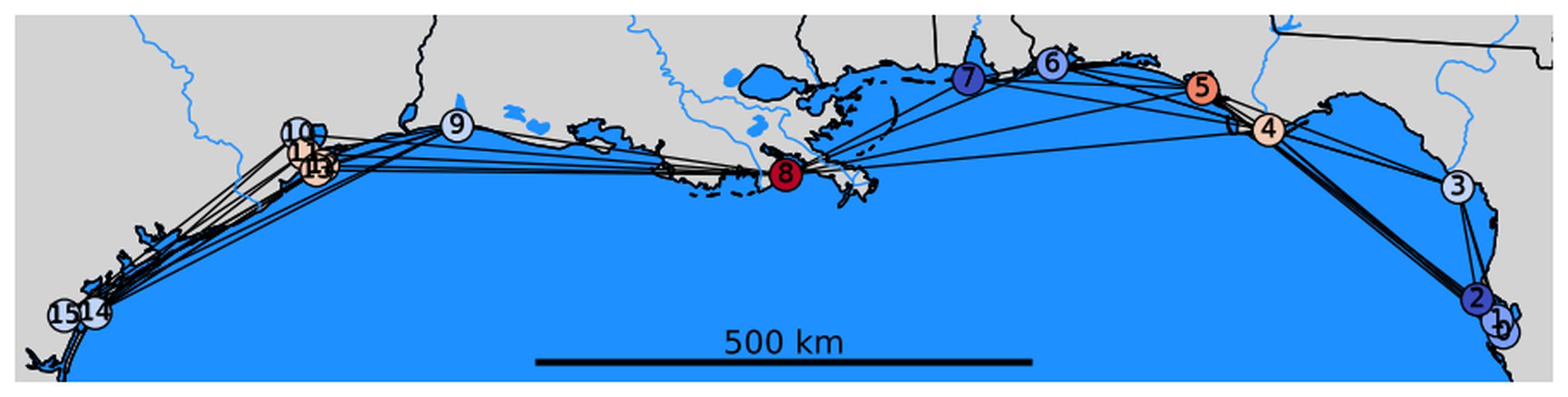}
    \caption{}
\end{subfigure}
\hfill
\begin{subfigure}[c]{0.9\textwidth}
    \centering
    \includegraphics[width=0.4\textwidth]{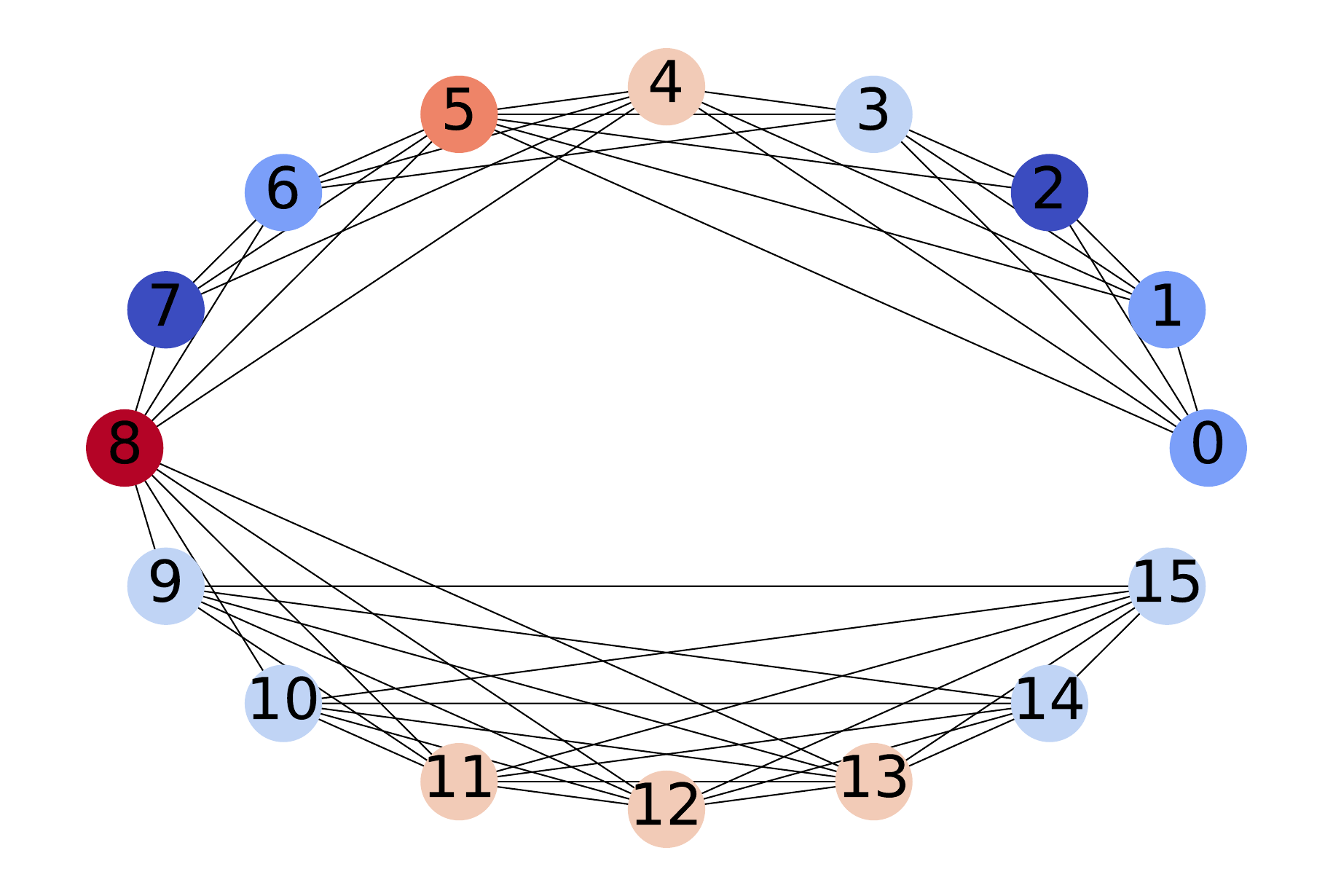}
    \caption{}
\end{subfigure}
\hfill
\begin{subfigure}[c]{0.9\textwidth}
    \centering
    \includegraphics[width=0.6\textwidth]{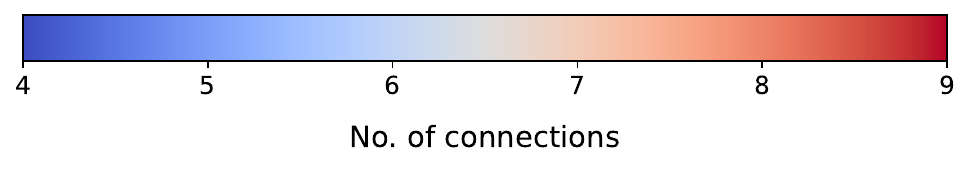}
\end{subfigure}
\hfill
\caption{(a): Geospatial representation of the \projectname{} graph used for this study. (b): Circular representation of the \projectname{} graph. Node indices in both schematics indicate gauge stations as listed in Table~\ref{tbl:gauge_stations}. Color scale in both schematics indicates the degree of each node (i.e., the total number of nodes each node is connected to via edges).}
\label{fig:graph}
\end{figure}

Applying the data processing steps explained in Section~\ref{ssec:data-acquisition} led to the selection of 16 gauge stations in the region of interest (the Gulf Coast region of the southern U.S.) that contained both observed and forecast water level data for all of the considered storms in Table~\ref{tbl:hurricanes}. 
Subsequently, implementing the methodology explained in Section~\ref{ssec:graph-construction} for constructing a graph model of these 16 gauge stations, with correlation and radius thresholds, $\rho_{\min}$ and $d_{\max}$, set to 0.8 and 500 km, respectively, leads to the graph shown in Figure~\ref{fig:graph}, with the gauge stations corresponding to each node listed in Table~\ref{tbl:gauge_stations}. Node connectivity was calculated based on the Pearson correlation coefficient between water level observations at each node for all the hurricanes in the training set (see Section~\ref{ssec:graph-construction}). The values chosen for $\rho_{\min}$ and $d_{\max}$ represent the lowest limits that avoid disconnected stations while also ensuring that connections for each station are made within a reasonable geographic proximity. Node no. 8 (Grand Isle, LA, 8761724, NOAA-NOS) has the highest number of connections (i.e., 9). On the other hand, even the least connected nodes have no fewer than four connections. This ensures that effective message passing can occur throughout the graph. This graph was used without modification for predicting water level bias values for the specific set of stations for Hurricane Idalia (2023) comprising the test set. No data from the test set were used for the construction of the graph in Figure~\ref{fig:graph}.


\begin{table}[h!]
\centering
\caption{List of stations considered for constructing the graph model of \projectname{} in its schematic representation in Figure~\ref{fig:graph}.}
\begin{tabular}{@{}lclll@{}}
\toprule
Node number & Station Information                                            &  &  &  \\ \midrule
0:          & Port Manatee, FL (8726384, NOAA-NOS)                    &  &  &  \\
1:          & St. Petersburg Tampa Bay, FL (8726520, NOAA-NOS)        &  &  &  \\
2:          & Clearwater Beach, FL (8726724, NOAA-NOS)                &  &  &  \\
3:          & Cedar Key, FL (8727520, NOAA-NOS)                       &  &  &  \\
4:          & Apalachicola, FL (8728690, NOAA-NOS)                    &  &  &  \\
5:          & Panama City, FL (8729108, NOAA-NOS)                     &  &  &  \\
6:          & Pensacola, FL (8729840, NOAA-NOS)                       &  &  &  \\
7:          & Dauphin Island, AL (8735180, NOAA-NOS)                  &  &  &  \\
8:          & Grand Isle, LA (8761724, NOAA-NOS)                      &  &  &  \\
9:          & Calcasieu Pass, LA (8768094, NOAA-NOS)                  &  &  &  \\
10:         & Morgans Point Barbours Cut, TX (8770613, NOAA-NOS)      &  &  &  \\
11:         & Eagle Point Galveston Bay, TX (8771013, NOAA-NOS)       &  &  &  \\
12:         & Galveston Bay Entrance North Jetty, TX (8771341, TCOON) &  &  &  \\
13:         & Galveston Pier 21, TX (8771450, NOAA-NOS)               &  &  &  \\
14:         & Port Aransas, TX (8775237, TCOON)                       &  &  &  \\
15:         & USS Lexington Corpus Christi Bay, TX (8775296, TCOON)   &  &  &  \\ \bottomrule
\end{tabular}
\label{tbl:gauge_stations}
\end{table}

\subsection{Case study: Hurricane Idalia (2023)}
\begin{figure}[H]
    \centering
    \includegraphics[width=0.9\linewidth]{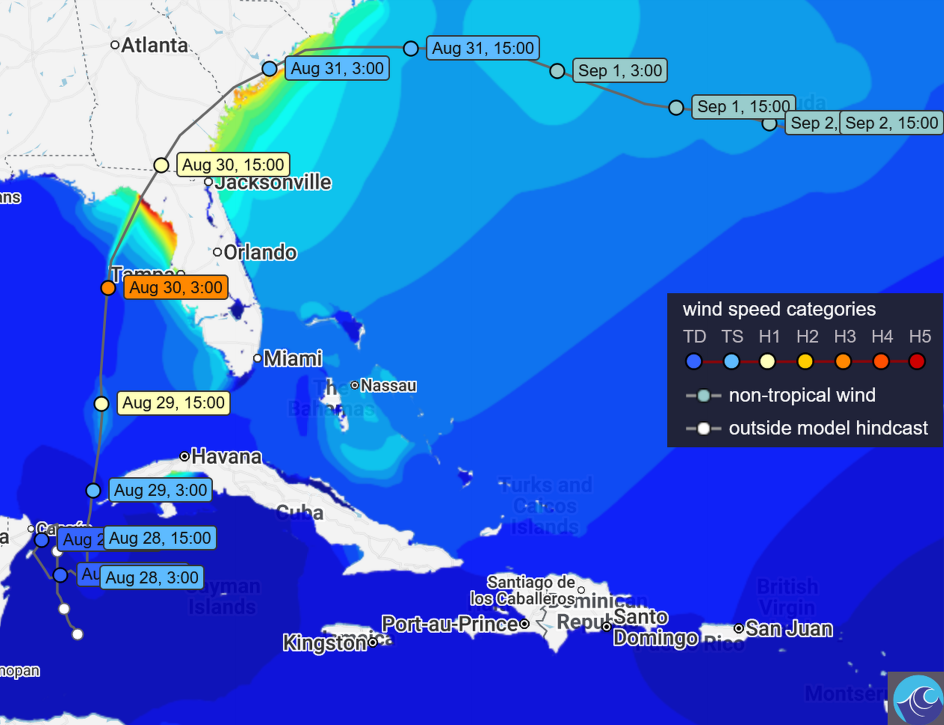}
    \caption{Official best track for Hurricane Idalia (2023) as provided by the National Hurricane Center~\cite{NHC} and visualized via the CERA online interactive visualization framework~\cite{CERA}. Wind speed categories are based on the Saffier-Simpson hurricane scale~\cite{SaffirSimpson} (H1-H5: Hurricane Category 1-5, TD: Tropical Depression, TS: Tropical Storm).}
    \label{fig:Idalia}
\end{figure}

The \projectname{} model, based on the geospatial graph representation of Figure~\ref{fig:graph} is applied to predict biases in water level forecasts for Hurricane Idalia (2023). Hurricane Idalia (2023) formed as a Tropical Depression (TD) on August 26 from a disturbance over the eastern Pacific, evolving into a Tropical Storm (TS) the next day. By 29 August, it had developed a partial eyewall and reached Category 1 intensity near western Cuba, showing a well-organized central dense overcast and strong convective bands. Later that day, a closed eye formed and became increasingly defined~\cite{Cangialosi2023}. Idalia underwent RI over the West Florida Shelf, west of Tampa Bay, as it moved toward Florida’s Big Bend region. It peaked as a Category 4 hurricane with 115-knot (213 km/h) winds and a central pressure of 942 mbar just hours before landfall near Keaton Beach, Florida, on August 30, marking the first major hurricane to hit Florida’s Big Bend region in over 70 years~\cite{Cangialosi2023}. After landfall, Idalia weakened over Georgia and the Carolinas, spawning tornadoes and heavy rainfall before dissipating over the Atlantic by September 8 (Figure~\ref{fig:Idalia}). Storm surge due to Hurricane Idalia (2023) reached 7–12 feet ($\sim 2-3.7$ m) in coastal areas, leading to catastrophic flooding in Taylor, Dixie, and Levy counties~\cite{Cangialosi2023}. The storm caused \$3.5–3.6 billion in damages, devastating agriculture, infrastructure, and homes, and resulted in 12 fatalities~\cite{Cangialosi2023}. Therefore, given Hurricane Idalia’s rapid intensification --- a process that recent studies suggest may have been influenced by river plumes~\cite{shi2025intensification}, which are typically not described explicitly in physics-based storm surge models --- and its impact on a region historically unaffected by major hurricanes, Idalia serves as a suitable case study for assessing the performance of a predictive bias correction model when applied to a real hurricane with unseen characteristics, to some extent.

\begin{figure}[H]
    \centering
    \includegraphics[width=0.9\linewidth]{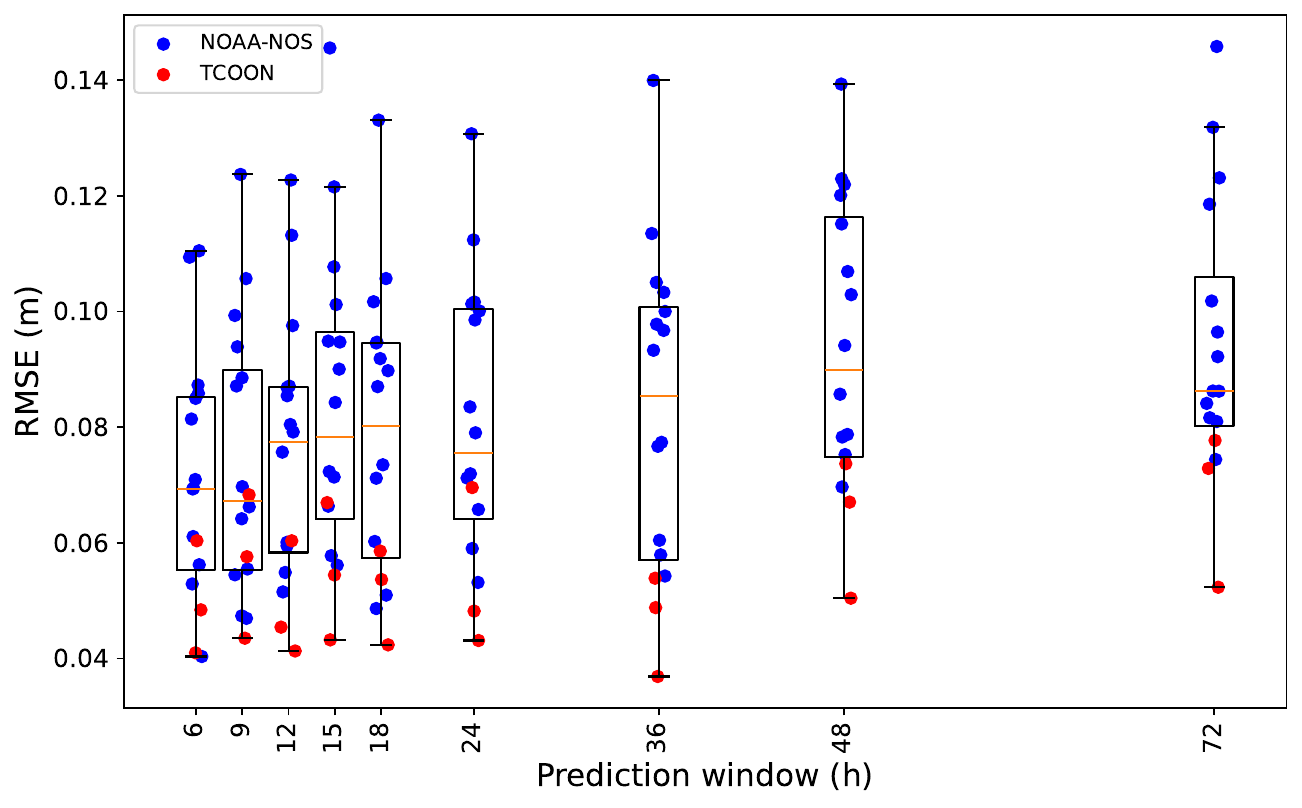}
    \caption{Distribution of the RMSE of \projectname{} bias correction predictions across all stations for different values of the prediction window. The median is denoted with an orange line segment. Blue (red) data points represent the RMSE values for each individual NOAA-NOS (TCOON) station (Table~\ref{tbl:gauge_stations}) in each case.}
    \label{fig:boxplot}
\end{figure}

The overall performance of \projectname{} to predict water level biases for different prediction window values in the range of 6-72 h, is presented in Figure~\ref{fig:boxplot}. Individual points represent the RMSE between real and predicted offsets for the bias predictions in each station, overlaid on a boxplot representing their distribution, for each value of the prediction window. Data points are color coded based on the agency that the corresponding station belongs to. It can be observed that median RMSE (shown as orange line segments in the boxplots in Figure~\ref{fig:boxplot}) increases, as expected, with prediction window up to 18 h. However, further increasing prediction window above 18h and up to 72h does not lead to a similar trend for increasing RMSE, as the median fluctuates around 0.08-0.09 m. The variance of RMSE, corresponding to the length of the box and whisker elements of the boxplots in Figure~\ref{fig:boxplot} is typically in the range of 0.08-0.10 m and does not appear to be significantly affected by the size of the prediction window. This behavior is promising for the application of GNN-based strategies such as \projectname{} for bias prediction in real-time operational forecasting systems (OFS), for which the desirable length of prediction windows is typically up to 3 days or longer~\cite{Penny2023}. Moreover, RMSE of bias predictions for the TCOON stations (red) appear to be lower compared to NOAA-NOS ones (blue), which could be attributed to the negligible impact of Hurricane Idalia in the Texas region, where these stations are located.

\begin{figure}[H]
\centering
\begin{subfigure}{0.325\textwidth}
    \includegraphics[width=1.05\textwidth]{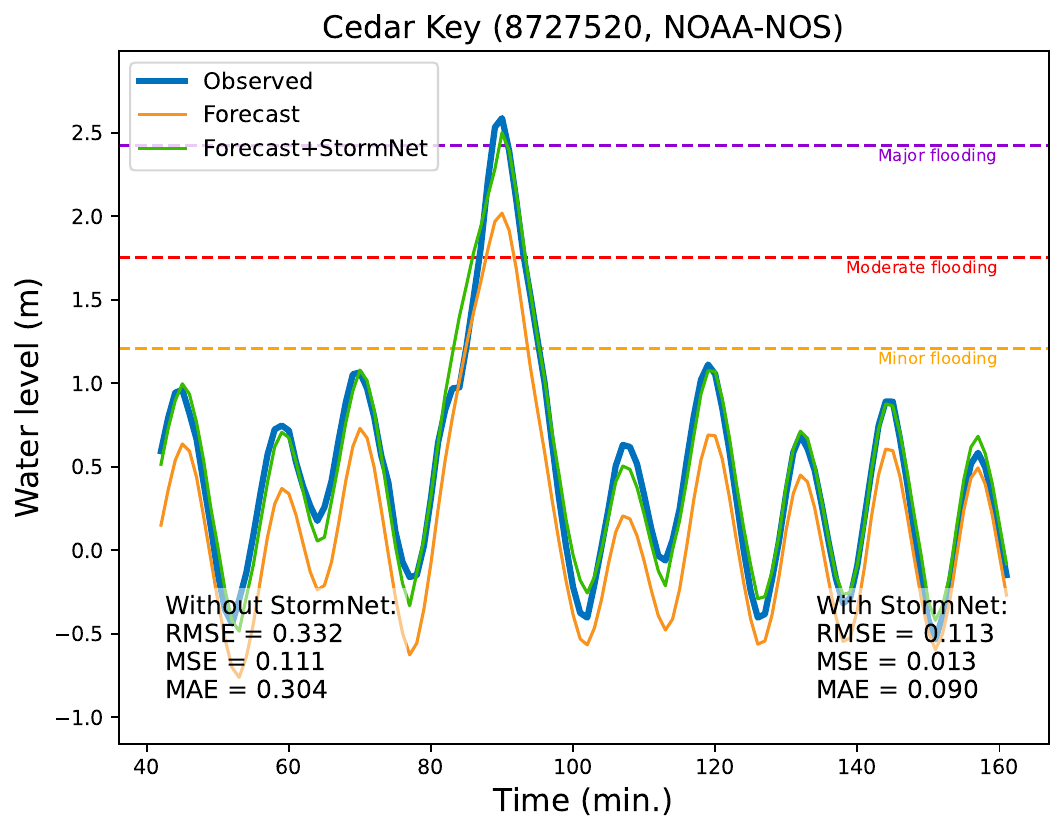}
    \caption{}
\end{subfigure}
\hfill
\begin{subfigure}{0.325\textwidth}
    \centering
    \mbox{\centering Prediction window: 12h}    
    \includegraphics[width=1.05\textwidth]{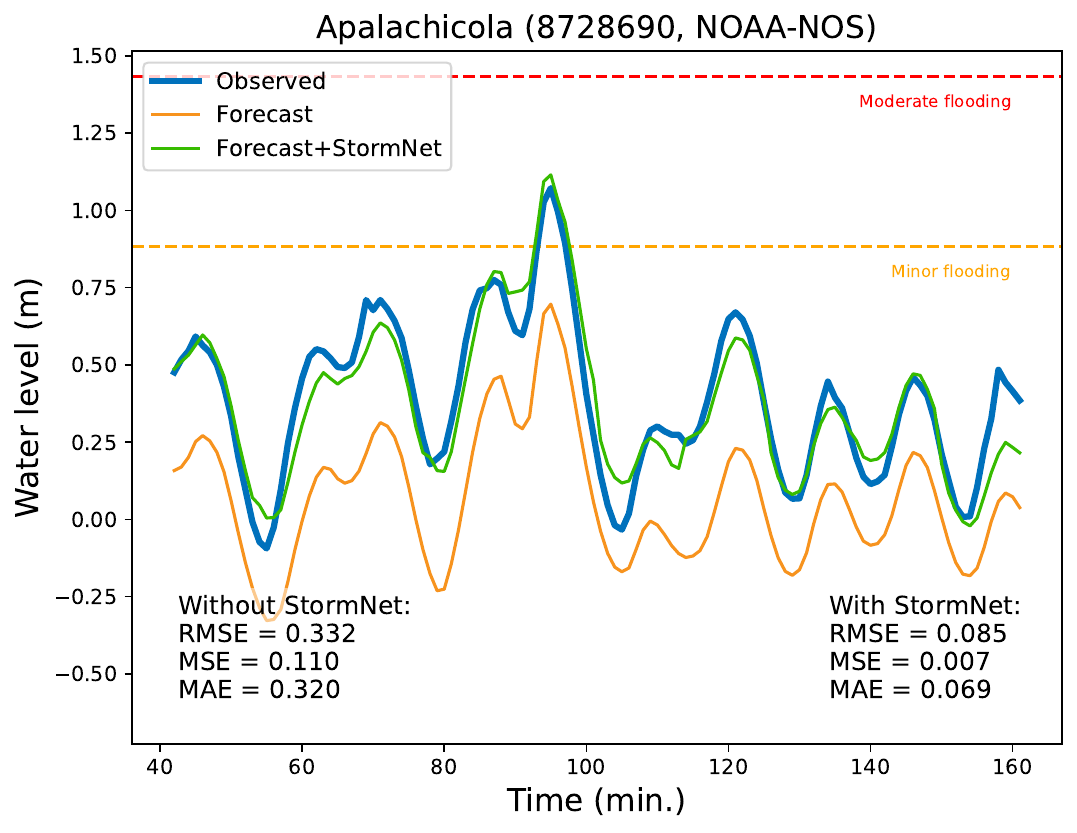}
    \caption{}
\end{subfigure}
\hfill
\begin{subfigure}{0.325\textwidth}
    \includegraphics[width=1.05\textwidth]{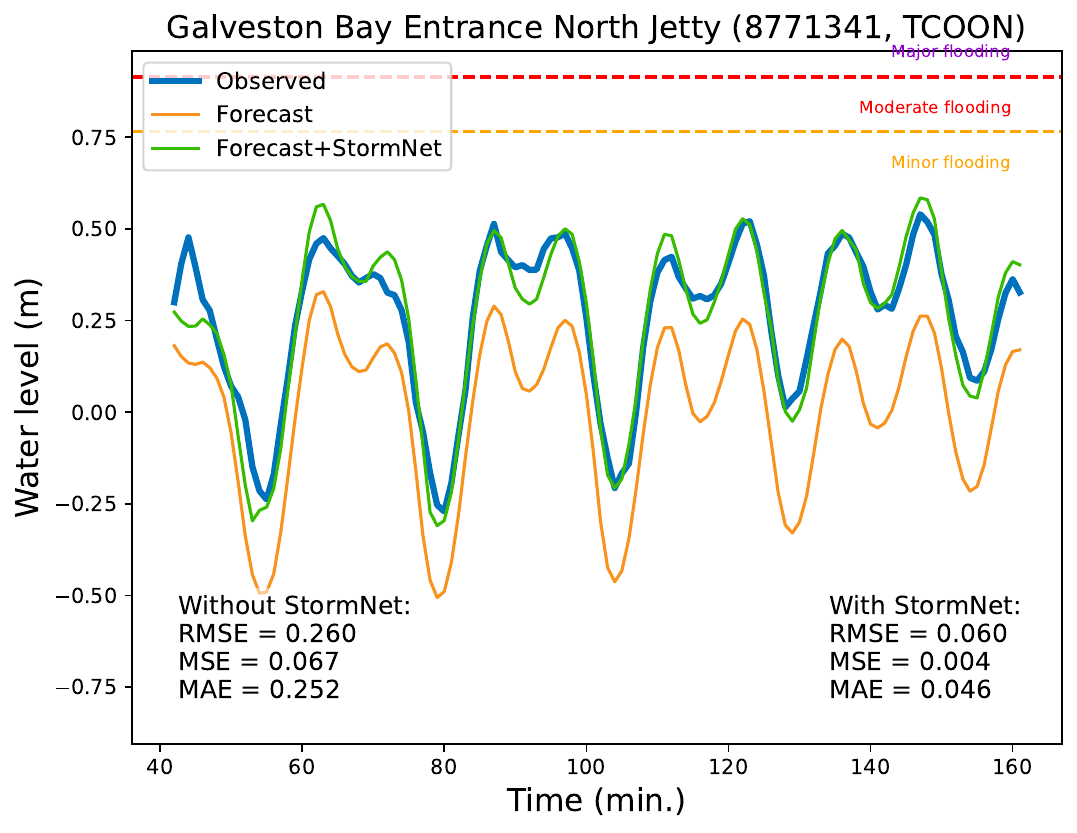}
    \caption{}
\end{subfigure}
\hfill
\begin{subfigure}{0.325\textwidth}
    \includegraphics[width=1.05\textwidth]{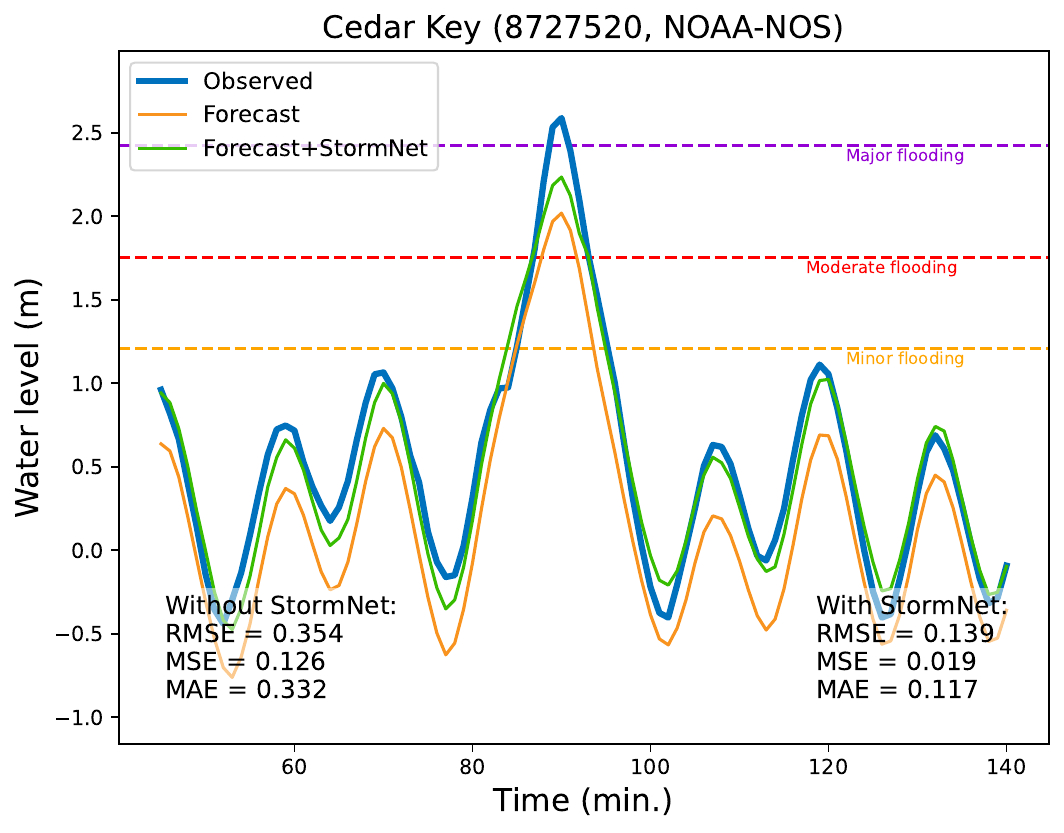}
    \caption{}
\end{subfigure}
\hfill
\begin{subfigure}{0.325\textwidth}
    \centering
    \mbox{\centering Prediction window: 48h}    
    \includegraphics[width=1.05\textwidth]{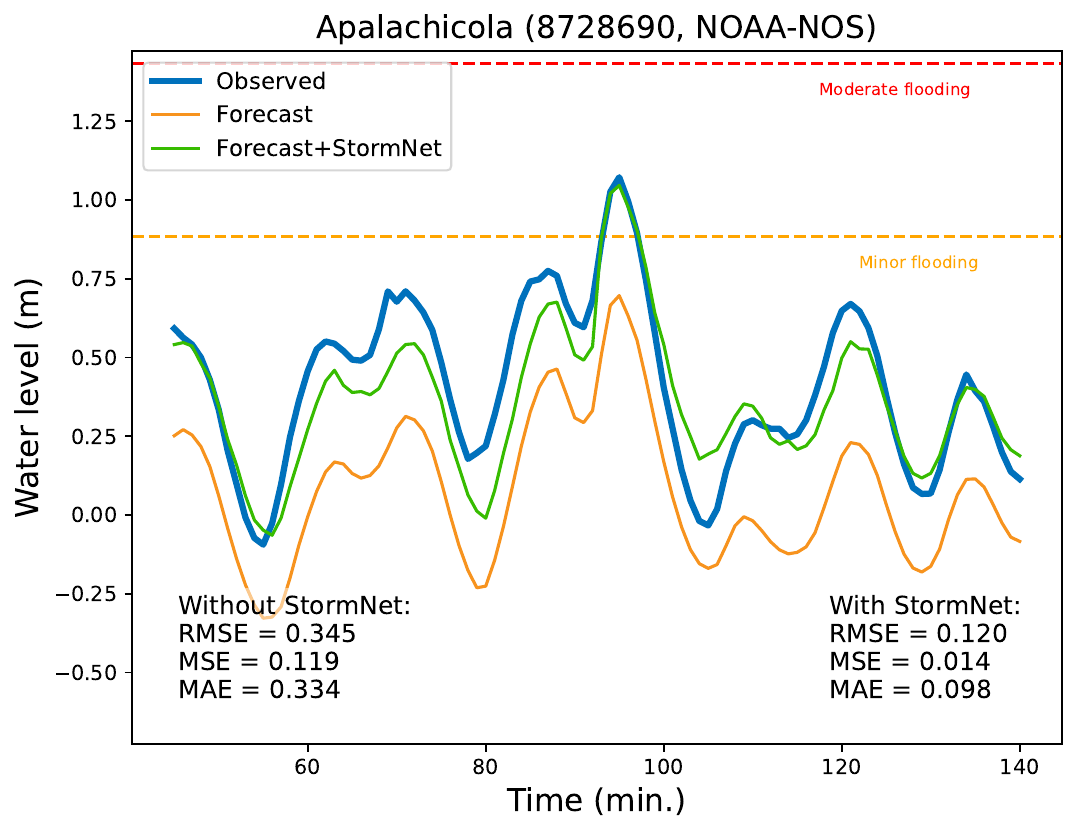}
    \caption{}
\end{subfigure}
\hfill
\begin{subfigure}{0.325\textwidth}
    \includegraphics[width=1.05\textwidth]{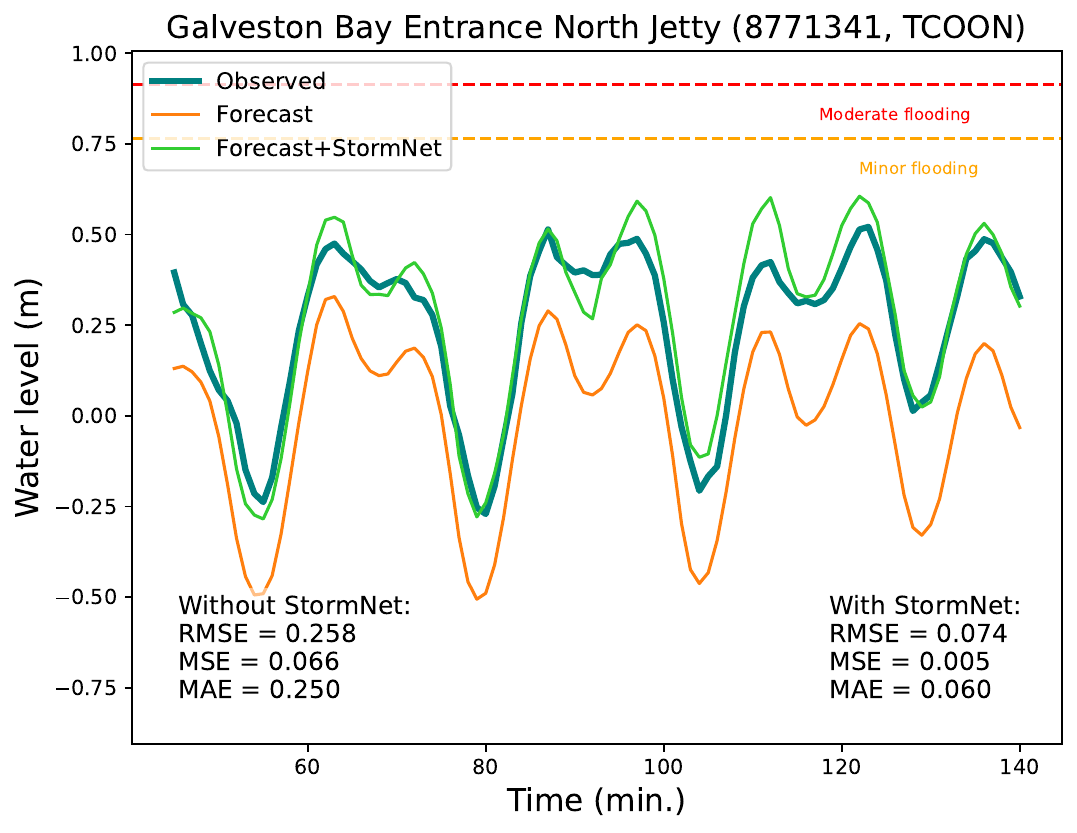}
    \caption{}
\end{subfigure}
\hfill
\begin{subfigure}{0.325\textwidth}
    \includegraphics[width=1.05\textwidth]{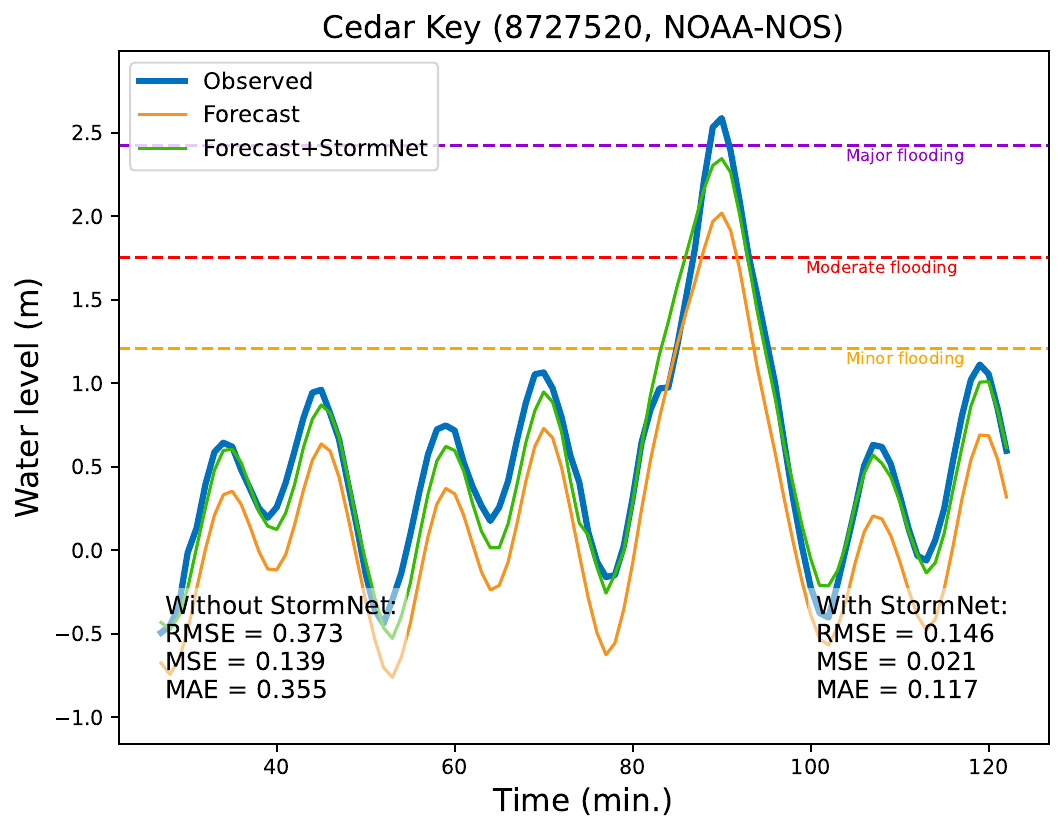}
    \caption{}
\end{subfigure}
\hfill
\begin{subfigure}{0.325\textwidth}
    \centering
    \mbox{\centering Prediction window: 72h}
    \includegraphics[width=1.05\textwidth]{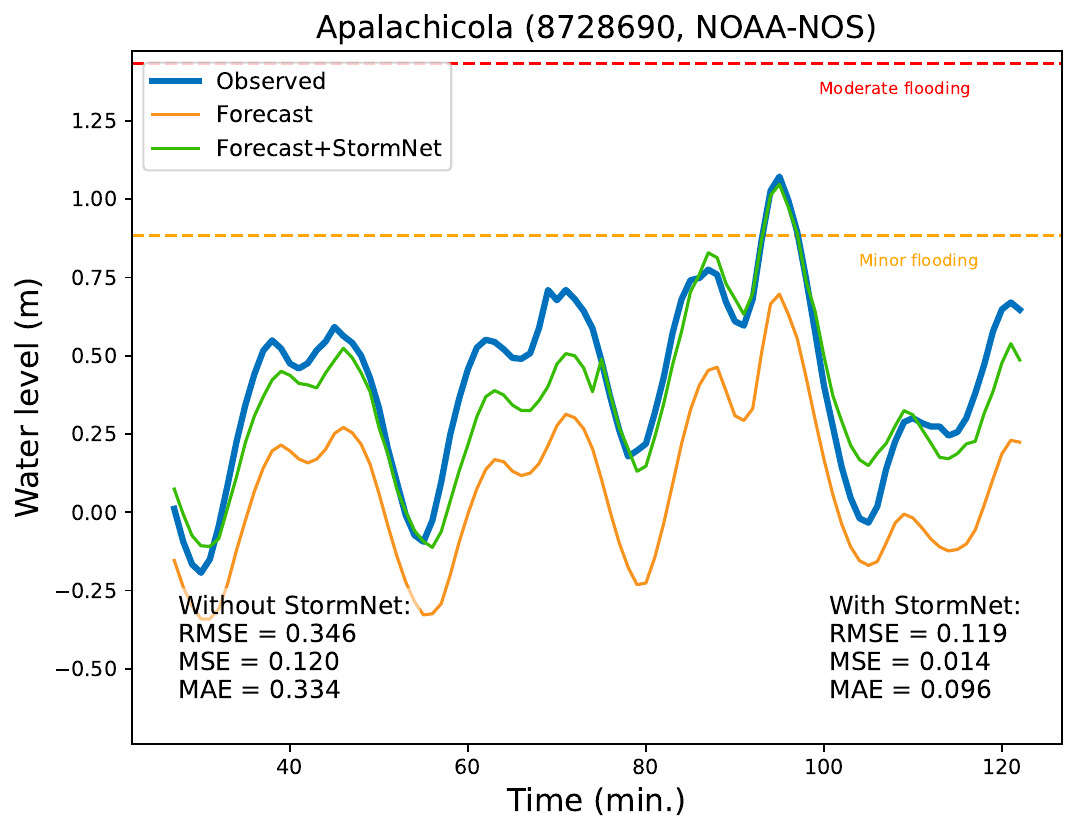}
    \caption{}
\end{subfigure}
\hfill
\begin{subfigure}{0.325\textwidth}
    \includegraphics[width=1.05\textwidth]{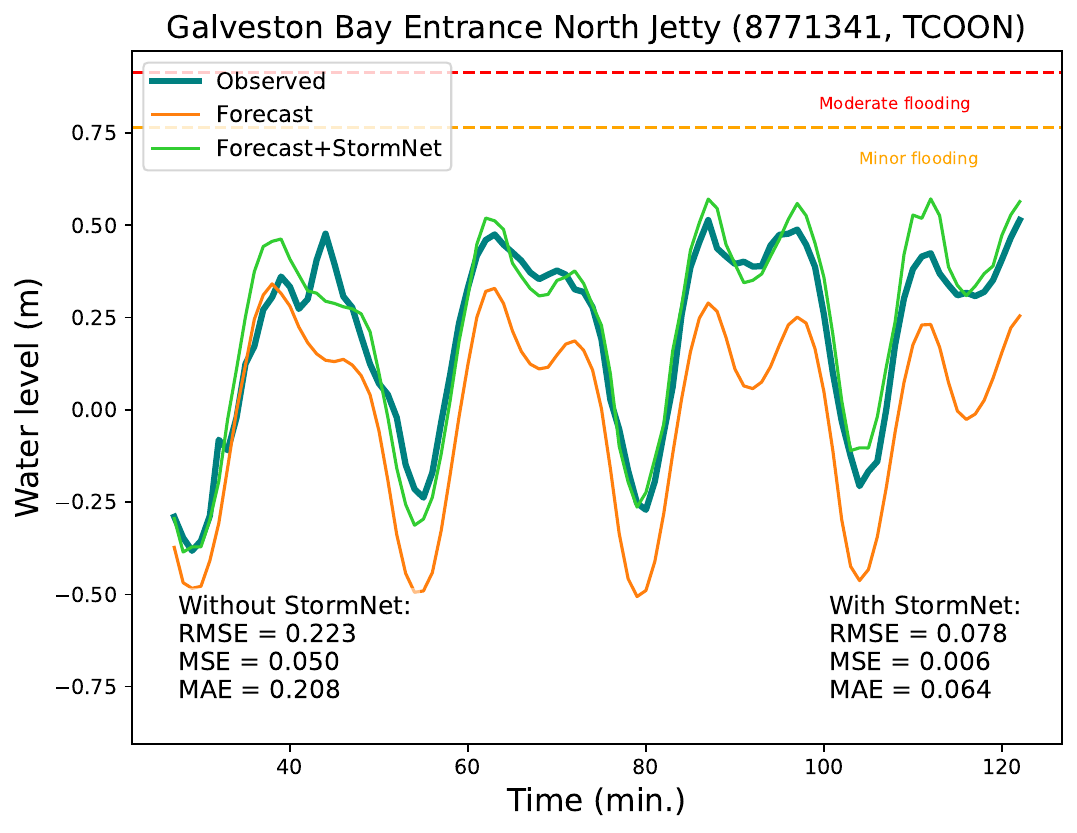}
    \caption{}
\end{subfigure}
\caption{Observed (blue), forecast (orange), and \projectname{}-corrected forecast (green, full lines) water levels for selected gauge stations for 12h-, 48h- and 72h-ahead predictions ((a-c), (d-f), and (g-i), respectively) and the corresponding evaluation metrics for each case. Minor, moderate and major flooding thresholds (shown in yellow, red and purple) are obtained by the National Oceanic and Atmospheric Administration~\cite{NOAAInundationHistory}, and established by the National Weather Service local Weather Forecast Offices (see~\cite{Sweet2018})}
\label{fig:STGNN_stations}
\end{figure}

Figure~\ref{fig:STGNN_stations} demonstrates the performance of \projectname{} in three representative stations for the largest considered prediction windows (i.e., 48 h and 72 h) used for \projectname{}. Cedar Key, FL (8727520, NOAA-NOS, node number 3, see Figure~\ref{fig:graph}) was directly impacted by severe storm surge during landfall, as indicated by the maximum observed water levels of $\sim 2.5$ m. On the other hand, Apalachicola, FL (8728690, NOAA-NOS, node number 4, Figure~\ref{fig:graph}), although in proximity to landfall, was not as severely impacted by Idalia, with a maximum water height of approximately $1.0$ m. Finally, Galveston Bay Entrance North Jetty, TX (8771341, TCOON, node number 12, Figure~\ref{fig:graph}) is located in a region further from landfall and thus did not record any noticeable storm-surge-related events. This choice of stations was made to examine the performance of \projectname{} in regions in relative geographic proximity but influenced by different storm surge severity (i.e., high, medium, low). 

Applying \projectname{} leads to a noticeable improvement in water level forecasts, as can be seen from Figure~\ref{fig:STGNN_stations} it can be seen that applying \projectname{} with a 48 h-ahead prediction window leads to a reduction in RMSE in water level forecasts by 60\%, 65\% and 73\% for the Cedar Key, Apalachicola and Galveston Bay Entrance North Jetty stations, respectively. Increasing the prediction window to 72 h still leads to a reduction of RMSE in water level forecasts by 50, 52 and 59\% for the same stations. Although improvements are slightly more pronounced for stations further from landfall, the model showed consistent performance in all three scenarios covered by the considered stations. \projectname{} was also capable of improving forecasts during peak surge, as it reduced the forecasting error at maximum observed water level in Cedar Key by 35 and 29\% for a 48 h- and 72 h-ahead prediction, respectively. The above demonstrates that \projectname{} was able to predict bias corrections at a reasonable accuracy for prolonged prediction windows and in cases with different storm surge severity.

Furthermore, the minor, moderate and major flooding thresholds, as established by NOAA and the U.S. National Weather Service (NWS)~\cite{NOAAInundationHistory}, are shown in Figure~\ref{fig:STGNN_stations} for the three selected stations. These thresholds are discussed in detail in~\cite{Sweet2018}, and they are set to $0.50 \pm 0.19$ m, $0.80 \pm 0.25$ m and $1.17 \pm 0.39$ m above the local diurnal tide range, amplified by 3-4\%, for minor, moderate, and major flooding, respectively. The three different levels are associated with the anticipated impact of the event, with ``minor" corresponding to mostly disruptive events (often causing no more than a plain ``nuisance" rather than significant damage),``moderate" to damaging, and ``major" to destructive anticipated outcomes. NOAA issues a ``Coastal/Lakeshore Flood Advisory" when minor flooding is ongoing or anticipated within 12 hours, a ``Coastal/Lakeshore Flood Warning" when ``flooding that poses a serious threat to life and property" is ongoing or expected in the next 12 to 24 hours (i.e., the first forecasting cycle), and a ``Coastal/Lakeshore Flood Watch", when severe impacts are expected in at least 12 to 48 hours~\cite{NOAA2017}. From Figure~\ref{fig:STGNN_stations}, it can be seen that uncorrected modeled water levels for the three selected stations tend to underestimate water levels by $\sim 0.25 - 0.35$ m. This leads to modeled values below the minor flooding threshold during landfall in the Apalachicola station, which is exceeded by the observed water levels (``false negative"). A similar situation occurs for Cedar Key at landfall, with the modeled water levels predicted to lie between moderate and major flooding thresholds, but the observed ones actually exceeding the major flooding threshold. Bias-corrected values obtained with \projectname{} appear to correctly predict the exceedance of the minor flooding threshold in the Apalachicola station even with a 72 h window, as well as the exceedance of the major flooding threshold in the Cedar Key station but only for a 12 h-ahead prediction. In the latter case, the exceedance of the major flooding threshold was properly captured up to 15 h ahead. This performance is in line with the NOAA requirements for issuing an advisory/warning/watch, which requires lead time of at least 12 h ahead of the expected event.

\subsection{Comparison with Sequential LSTM-based Model Predictions}
The novelty of \projectname{} lies in its use of a GNN, with a graph representation of nodes constructed based on the correlation of their water level observations (see Section~\ref{ssec:graph-construction}), in conjunction with LSTM layers, forming a spatiotemporal predictive framework. In this section, the performance of this approach is compared with the results of the LSTM-based model of~\cite{giaremis2024storm}, in which interactions between data from different stations were not explicitly modeled in the architecture. 

\begin{figure}[H]
\centering
\begin{subfigure}{0.495\textwidth}
    \includegraphics[width=\textwidth]{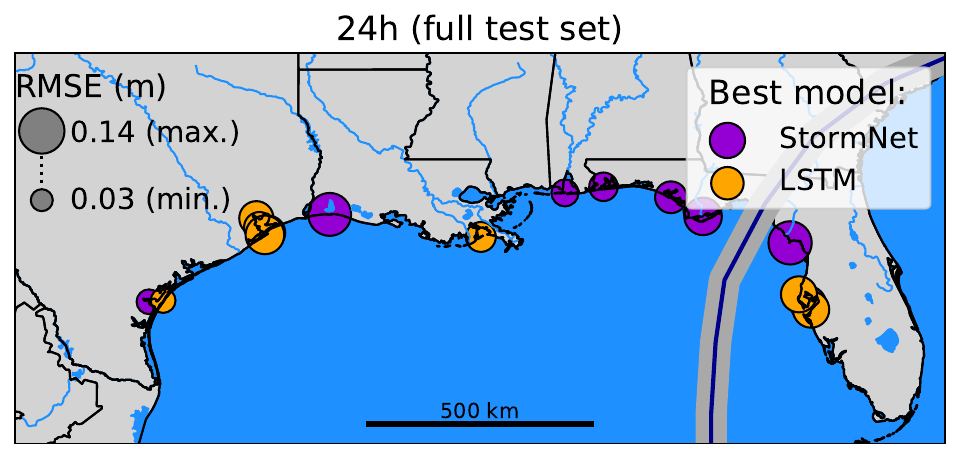}
    \caption{}
\end{subfigure}
\hfill
\begin{subfigure}{0.495\textwidth}
    \includegraphics[width=\textwidth]{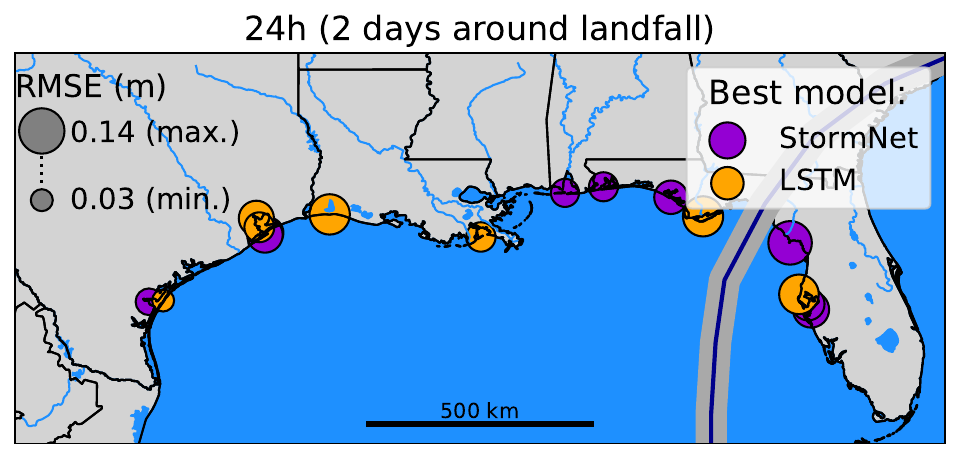}
    \caption{}
\end{subfigure}
\hfill
\begin{subfigure}{0.495\textwidth}
    \includegraphics[width=\textwidth]{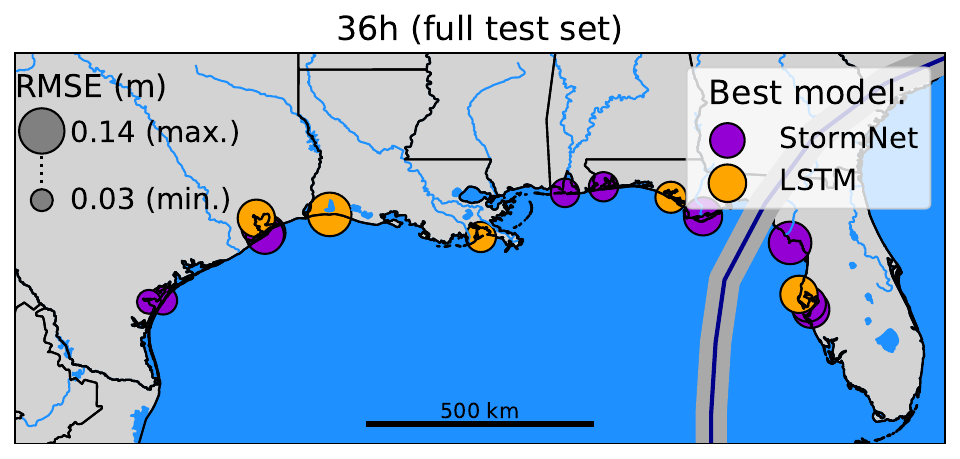}
    \caption{}
\end{subfigure}
\hfill
\begin{subfigure}{0.495\textwidth}
    \includegraphics[width=\textwidth]{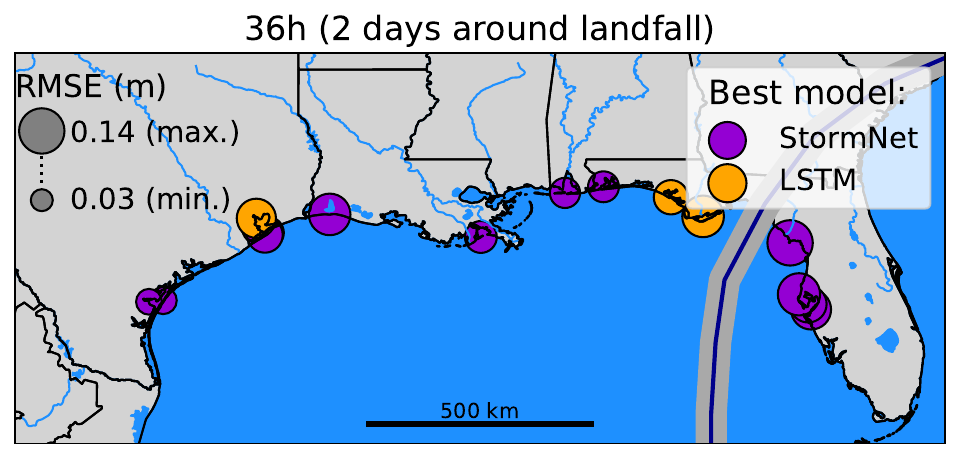}
    \caption{}
\end{subfigure}
\caption{Comparison of bias correction predictions generated with \projectname{} and a sequential LSTM-based approach ("LSTM") \cite{giaremis2024storm} for Hurricane Idalia (2023). Gauge stations in which \projectname{} (LSTM) yields better performance are denoted with purple (yellow), while larger (smaller) data point radii represent greater (lower) values of RMSE for the considered station. Each station is represented only by the model that yields the lowest RMSE (i.e., the best model). The gray-outlined dark blue line indicates the storm track. 24 h- and 36 h-ahead predictions are shown in (a, b) and (c, d), respectively. Predictions for the full test dataset are shown in (a, c) while predictions for a 2-day period around landfall are shown in (b, d).}
\label{fig:STGNN_vs_LSTM_map}
\end{figure}

In Figure~\ref{fig:STGNN_vs_LSTM_map} and Table~\ref{tbl:STGNN_vs_LSTM_tbl}, we compare the results of \projectname{} and the LSTM-based model of~\cite{giaremis2024storm} in bias-correcting water level forecasts in the gauge stations we have considered (Figures~\ref{fig:STGNN_stations} and Table~\ref{tbl:gauge_stations}). We also examined the accuracy of both models for both the full duration of the water level time series and a shorter span of two days around landfall, to further assess their performance specifically during the most intense storm conditions. In Figure~\ref{fig:STGNN_vs_LSTM_map}, the results of the model with the lowest RMSE are presented for each station (with purple denoting \projectname{} and orange the LSTM model of~\cite{giaremis2024storm}). Each station is represented only by the model that yields the lowest RMSE among the two. For applying 24 h-ahead corrections, both models yield quite similar accuracy. For corrections throughout the full timeseries, each model outperforms the other in 8 of the stations, with the LSTM-based model of~\cite{giaremis2024storm} even yielding slightly lower average RMSE across all stations, compared to \projectname{}. \projectname{} corrections show slightly better accuracy for the two-day period around landfall, as applying \projectname{} leads to a lower RMSE for 9 of the 16 stations, with only a small improvement to the average RMSE. The picture is clearer for the 36 h-ahead corrections, as \projectname{} outperforms the LSTM-based model of~\cite{giaremis2024storm} in 11 of 16 stations in the full time series case, and in 13 of 16 stations in the case of the two-day period around landfall, showing approximately 11\% and 12\% lower average RMSE throughout all stations in each case, respectively.



\begin{table}[h!]
\setlength\tabcolsep{1.5pt}
\caption{Comparison of RMSE values for 24h- and 36h-ahead water level bias predictions for the full dataset and a two-day period around landfall for prediction with either \projectname{} or a sequential LSTM-based model~\cite{giaremis2024storm}, for each of the considered gauge stations and Hurricane Idalia (2023). The lowest RMSE value for each case among the two models are indicated with bold. The total amount of stations in which each model yields lower RMSE in each case in noted in the next-to-last row, and the average RMSE across all stations is included in the last row. The mapping of node numbers is displayed in Figure~\ref{fig:graph} and the respective gauge station information is shown in Table~\ref{tbl:gauge_stations}.}

\begin{tabular}{rcccccccc}
                                     & \multicolumn{8}{c}{RMSE (m)}                                                                                                                                                                                                                                     \\
                                     & \multicolumn{4}{c|}{24h}                                                                                                                  & \multicolumn{4}{c}{36h}                                                                                              \\ \hline
\multicolumn{1}{r|}{}                & \multicolumn{2}{c|}{Full test set}                                  & \multicolumn{2}{c|}{2 days around landfall}                         & \multicolumn{2}{c|}{Full test set}                                  & \multicolumn{2}{c}{2 days around landfall}     \\ \hline
\multicolumn{1}{r|}{Node ID}         & \projectname{} & \multicolumn{1}{c|}{LSTM~\cite{giaremis2024storm}} & \projectname{} & \multicolumn{1}{c|}{LSTM~\cite{giaremis2024storm}} & \projectname{} & \multicolumn{1}{c|}{LSTM~\cite{giaremis2024storm}} & \projectname{} & LSTM~\cite{giaremis2024storm} \\
\multicolumn{1}{r|}{0}               & 0.088          & \multicolumn{1}{c|}{\textbf{0.087}}                & \textbf{0.086} & \multicolumn{1}{c|}{0.087}                         & \textbf{0.089} & \multicolumn{1}{c|}{0.093}                         & \textbf{0.106} & 0.119                         \\
\multicolumn{1}{r|}{1}               & \textbf{0.085} & \multicolumn{1}{c|}{0.085}                         & \textbf{0.071} & \multicolumn{1}{c|}{0.081}                         & \textbf{0.083} & \multicolumn{1}{c|}{0.093}                         & \textbf{0.090} & 0.117                         \\
\multicolumn{1}{r|}{2}               & 0.103          & \multicolumn{1}{c|}{\textbf{0.086}}                & 0.109          & \multicolumn{1}{c|}{\textbf{0.103}}                & 0.100          & \multicolumn{1}{c|}{\textbf{0.092}}                & \textbf{0.117} & 0.122                         \\
\multicolumn{1}{r|}{3}               & \textbf{0.123} & \multicolumn{1}{c|}{0.169}                         & \textbf{0.125} & \multicolumn{1}{c|}{0.196}                         & \textbf{0.120} & \multicolumn{1}{c|}{0.179}                         & \textbf{0.136} & 0.200                         \\
\multicolumn{1}{r|}{4}               & \textbf{0.093} & \multicolumn{1}{c|}{0.107}                         & 0.113          & \multicolumn{1}{c|}{\textbf{0.106}}                & \textbf{0.096} & \multicolumn{1}{c|}{0.109}                         & 0.126          & \textbf{0.114}                \\
\multicolumn{1}{r|}{5}               & \textbf{0.066} & \multicolumn{1}{c|}{0.069}                         & \textbf{0.074} & \multicolumn{1}{c|}{0.090}                         & 0.069          & \multicolumn{1}{c|}{\textbf{0.064}}                & 0.083          & \textbf{0.078}                \\
\multicolumn{1}{r|}{6}               & \textbf{0.054} & \multicolumn{1}{c|}{0.058}                         & \textbf{0.056} & \multicolumn{1}{c|}{0.063}                         & \textbf{0.057} & \multicolumn{1}{c|}{0.076}                         & \textbf{0.063} & 0.072                         \\
\multicolumn{1}{r|}{7}               & \textbf{0.048} & \multicolumn{1}{c|}{0.062}                         & \textbf{0.052} & \multicolumn{1}{c|}{0.066}                         & \textbf{0.053} & \multicolumn{1}{c|}{0.071}                         & \textbf{0.060} & 0.070                         \\
\multicolumn{1}{r|}{8}               & 0.069          & \multicolumn{1}{c|}{\textbf{0.059}}                & 0.065          & \multicolumn{1}{c|}{\textbf{0.057}}                & 0.062          & \multicolumn{1}{c|}{\textbf{0.060}}                & \textbf{0.067} & 0.068                         \\
\multicolumn{1}{r|}{9}               & \textbf{0.121} & \multicolumn{1}{c|}{0.130}                         & 0.108          & \multicolumn{1}{c|}{\textbf{0.107}}                & 0.127          & \multicolumn{1}{c|}{\textbf{0.126}}                & \textbf{0.114} & 0.115                         \\
\multicolumn{1}{r|}{10}              & 0.122          & \multicolumn{1}{c|}{\textbf{0.077}}                & 0.132          & \multicolumn{1}{c|}{\textbf{0.081}}                & 0.131          & \multicolumn{1}{c|}{\textbf{0.092}}                & 0.142          & \textbf{0.101}                \\
\multicolumn{1}{r|}{11}              & 0.089          & \multicolumn{1}{c|}{\textbf{0.061}}                & 0.083          & \multicolumn{1}{c|}{\textbf{0.059}}                & \textbf{0.085} & \multicolumn{1}{c|}{0.100}                         & \textbf{0.080} & 0.110                         \\
\multicolumn{1}{r|}{12}              & 0.085          & \multicolumn{1}{c|}{\textbf{0.078}}                & \textbf{0.063} & \multicolumn{1}{c|}{0.065}                         & \textbf{0.081} & \multicolumn{1}{c|}{0.103}                         & \textbf{0.066} & 0.080                         \\
\multicolumn{1}{r|}{13}              & 0.108          & \multicolumn{1}{c|}{\textbf{0.099}}                & \textbf{0.084} & \multicolumn{1}{c|}{0.091}                         & \textbf{0.095} & \multicolumn{1}{c|}{0.127}                         & \textbf{0.084} & 0.127                         \\
\multicolumn{1}{r|}{14}              & 0.054          & \multicolumn{1}{c|}{\textbf{0.040}}                & 0.061          & \multicolumn{1}{c|}{\textbf{0.032}}                & \textbf{0.052} & \multicolumn{1}{c|}{0.055}                         & \textbf{0.049} & 0.050                         \\
\multicolumn{1}{r|}{15}              & \textbf{0.040} & \multicolumn{1}{c|}{0.049}                         & \textbf{0.047} & \multicolumn{1}{c|}{0.054}                         & \textbf{0.038} & \multicolumn{1}{c|}{0.059}                         & \textbf{0.044} & 0.076                         \\ \hline
\multicolumn{1}{r|}{Lowest RMSE in:} & 8              & \multicolumn{1}{c|}{8}                             & 9              & \multicolumn{1}{c|}{7}                             & 11             & \multicolumn{1}{c|}{5}                             & 13             & 3                             \\ \hline
\multicolumn{1}{r|}{Average RMSE:}   & 0.084          & \multicolumn{1}{c|}{0.082}                         & 0.083          & \multicolumn{1}{c|}{0.084}                         & 0.084          & \multicolumn{1}{c|}{0.094}                         & 0.089          & 0.101                         \\
\multicolumn{1}{r|}{}                & $\pm$0.026          & \multicolumn{1}{c|}{$\pm$0.031}                         & $\pm$0.026          & \multicolumn{1}{c|}{$\pm$0.036}                         & $\pm$0.027          & \multicolumn{1}{c|}{0.031}                         & $\pm$0.030          & $\pm$0.035                        
\end{tabular}
\label{tbl:STGNN_vs_LSTM_tbl}
\end{table}

\begin{figure}[H]
\centering
\begin{subfigure}{0.325\textwidth}
    \includegraphics[width=1.05\textwidth]{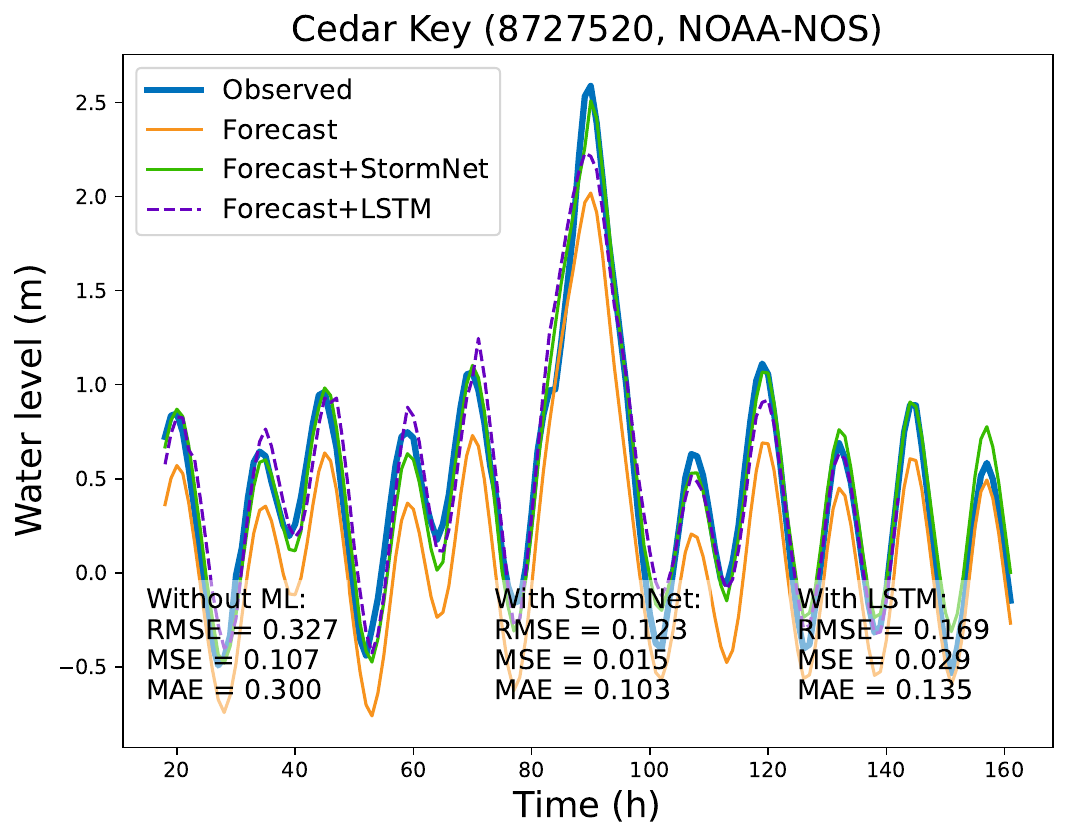}
    \caption{}
\end{subfigure}
\hfill
\begin{subfigure}{0.325\textwidth}
    \centering
    \mbox{\centering Prediction window: 24 h}    \includegraphics[width=1.05\textwidth]{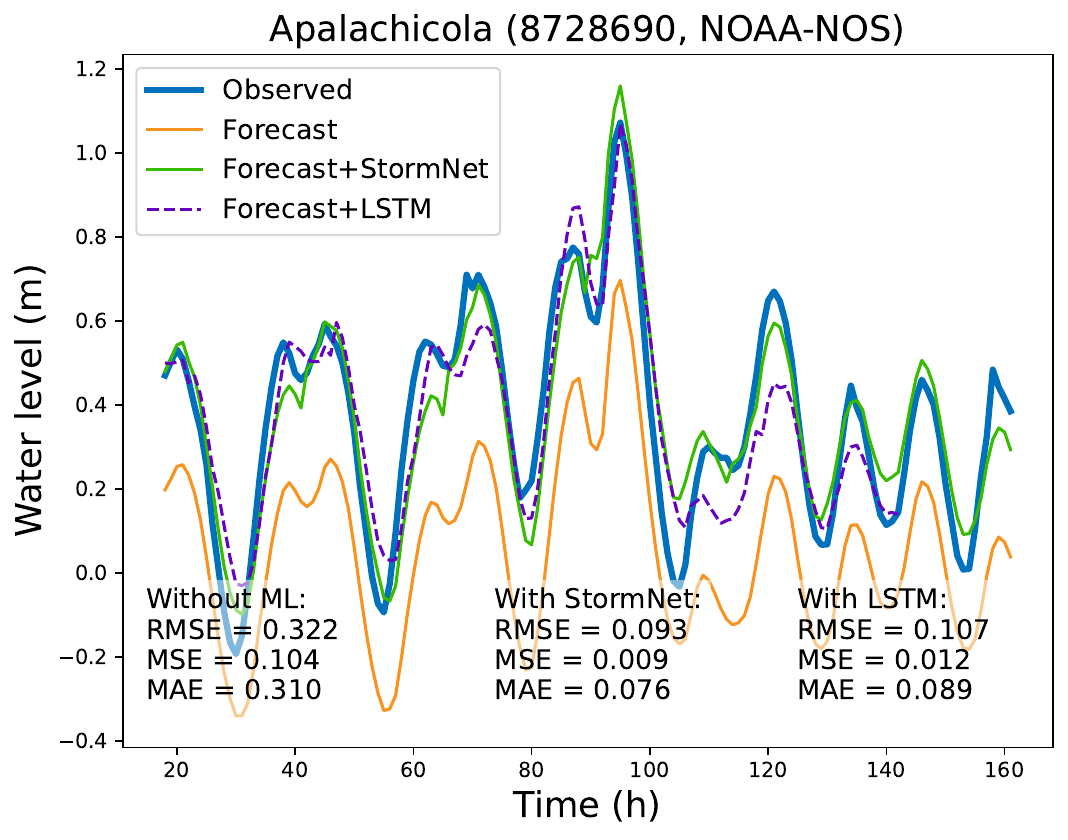}
    \caption{}
\end{subfigure}
\hfill
\begin{subfigure}{0.325\textwidth}
    \includegraphics[width=1.12\textwidth]{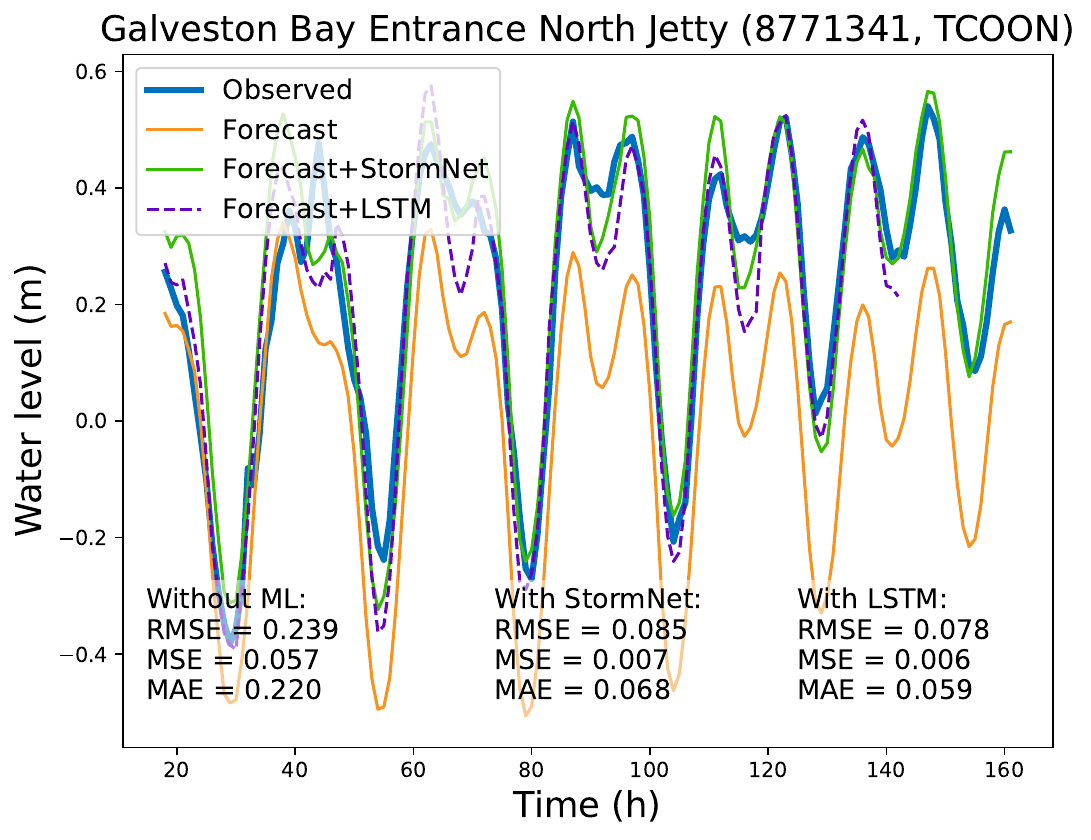}
    \caption{}
\end{subfigure}
\hfill
\begin{subfigure}{0.325\textwidth}
    \includegraphics[width=1.05\textwidth]{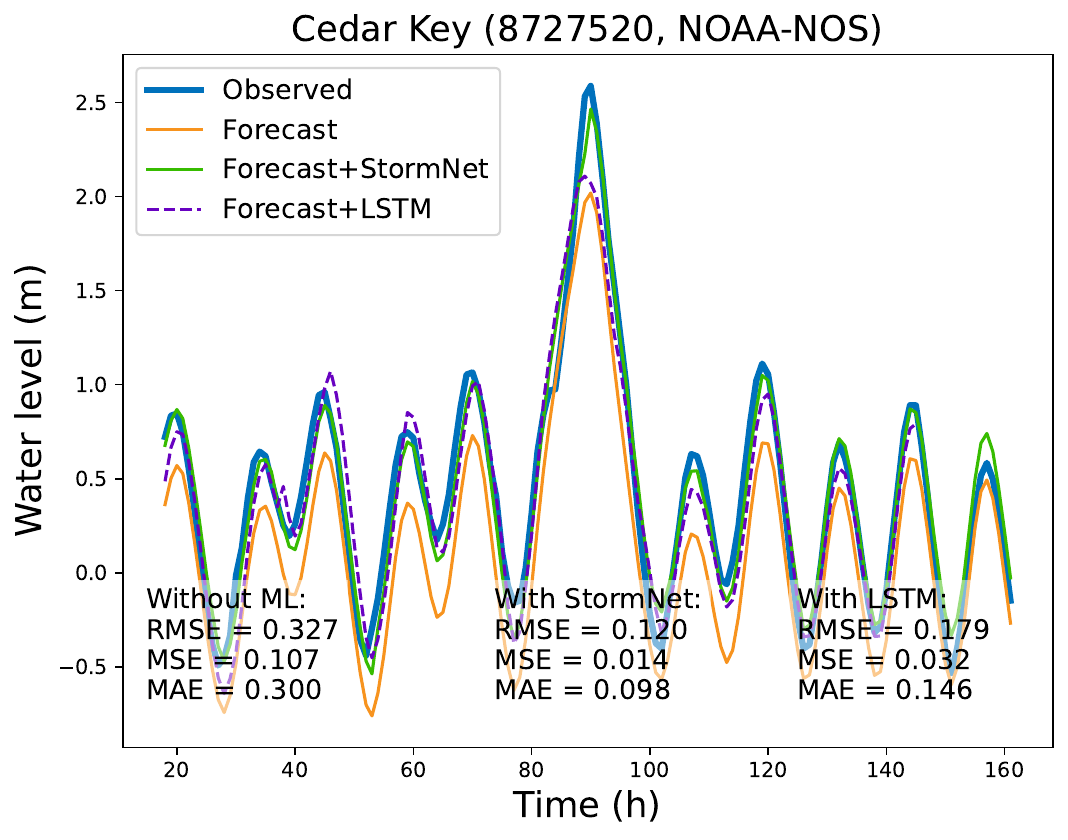}
    \caption{}
\end{subfigure}
\hfill
\begin{subfigure}{0.325\textwidth}
    \centering
    \mbox{\centering Prediction window: 36 h}
    \includegraphics[width=1.05\textwidth]{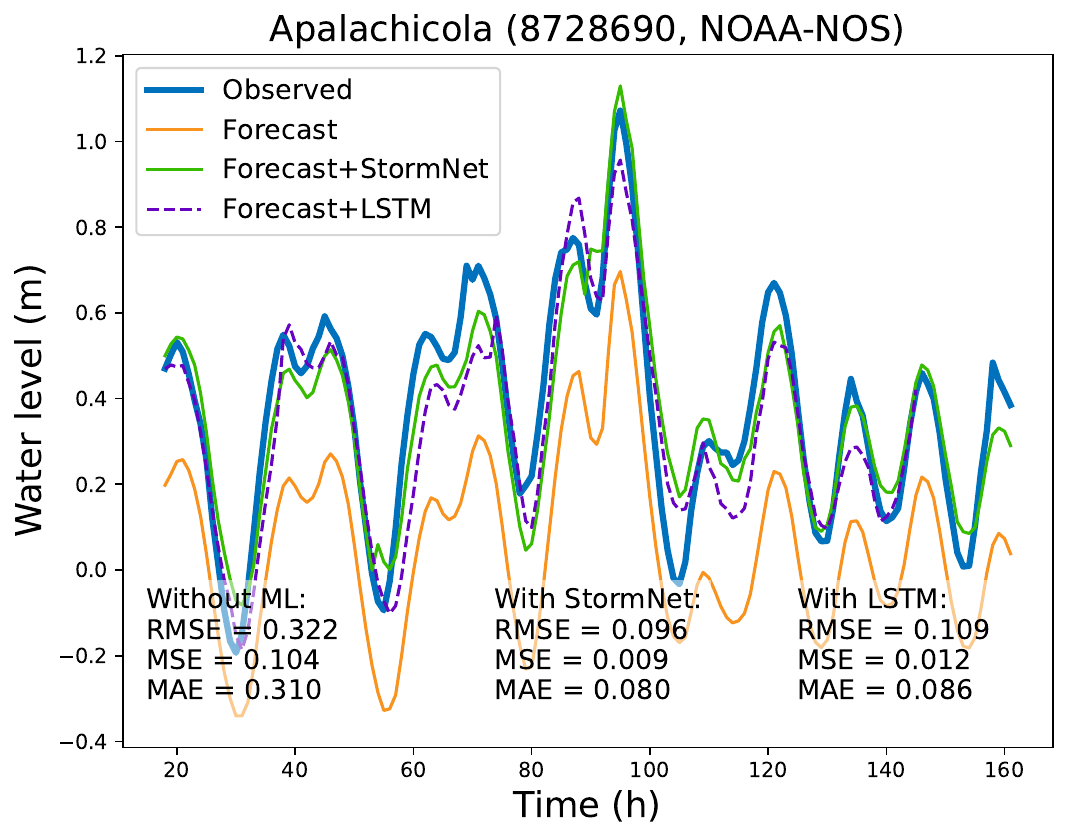}
    \caption{}
\end{subfigure}
\hfill
\begin{subfigure}{0.325\textwidth}
    \includegraphics[width=1.12\textwidth]{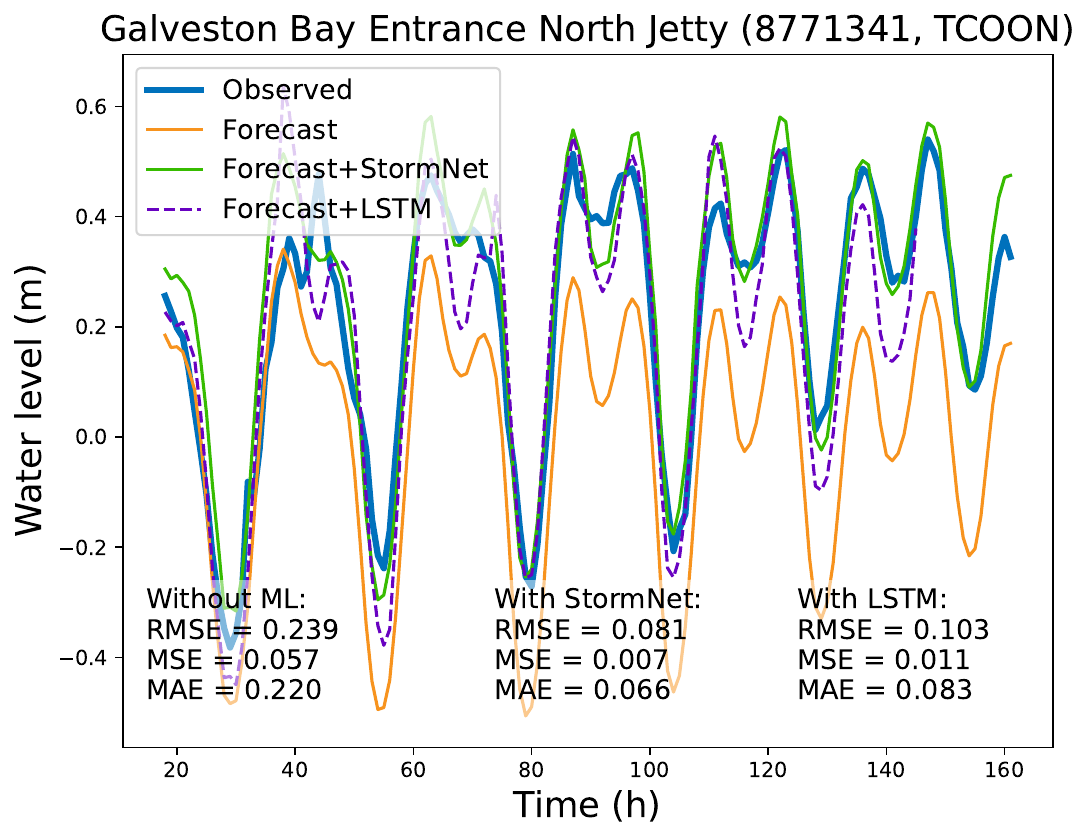}
    \caption{}
\end{subfigure}
\caption{Observed (blue), forecast (orange), \projectname{} ("ST-GNN")-corrected forecast (green, full lines) and LSTM-corrected forecast \cite{giaremis2024storm} (purple, dashed lines) water levels for selected gauge stations, 24 h- and 36 h-ahead predictions ((a-c) and (d-f), respectively) and the corresponding evaluation metrics for each case.}
\label{fig:STGNN_vs_LSTM_stations}
\end{figure}

The performance of the two models in the three stations representing areas of high, medium and, low storm impact examined in the previous section is presented in Figure~\ref{fig:STGNN_vs_LSTM_stations}. \projectname{}-corrected water levels yield consistently lower RMSE compared with their LSTM-corrected counterparts for both prediction windows (24 and 36 h). \projectname{} also captures the peak surge during landfall at the Cedar Key station (8727520, NOAA-NOS, high severity example) better than LSTM. In the Apalachicola station (8728690, NOAA-NOS, medium severity example), applying \projectname{} leads to slightly over-predicting water levels at peak surge, while the LSTM model, although capturing the correct value for 24 h-ahead corrections, slightly underestimate the water level for the 36 h-ahead case. In both cases, though, \projectname{} yields 12\%-13\% lower RMSE throughout the full water level time series. Finally, in the low severity example station (Galveston Bay Entrance North Jetty, 8771341, TCOON), the LSTM model shows slightly lower RMSE than \projectname{} for 24 h-ahead corrections, however, \projectname{} yields a 21\% lower RMSE than LSTM in the 36 h-ahead correction case.

Moreover, another important aspect for the application of such models in OFS is their training time. The training time of \projectname{} was around 5 minutes in all cases, while the LSTM-based model of~\cite{giaremis2024storm} was shown to require 30 minutes to 1 hour to be trained on a similar dataset size on the same hardware, so \projectname{} requires approximately 80\% less training time. This can be important in the context of the implementation of such models in OFS. Both models are intended as post-processing tools, to be pre-trained on historical storm data and produce corrections based on inputs from real-time measurements. However, shorter training times can enable exploring model training in real time with different training sets, comprising storms with similar or different characteristics to a real-world, ongoing scenario.


\subsection{Ablation}

In this section, we estimate the impact of each of the main components of \projectname{} (see Section~\ref{sec:proposed_model}) on overall performance. We do this by evaluating the RMSE of the bias-corrected water levels across all stations using models that include different subsets of components of the full \projectname{} architecture. For this analysis, the prediction window was set to 24 h. The results are presented in Table~\ref{tbl:ablation}. Using only a GAT or a GCN component along with LSTM one leads to the highest RMSE, as expected, with the GAT yielding a lower RMSE than GCN. Using both GAT and GCN with LSTM leads to a reduction of RMSE by 11\% compared to using only GAT. Adding an MLP to the GAT+LSTM architecture improves RMSE by roughly 9\%. Finally, adding the MLP component to the full GAT+GCN+LSTM architecture, i.e., the complete \projectname{} model, yields a further 3\% reduction in RMSE.


\begin{table}[h!]
\centering
\caption{RMSE values of bias correction predictions across all stations produced by using only the specified component subsets of the full \projectname{} model.}
\begin{tabular}{r|cl}
Components       & RMSE (m) &  \\ 
\hline
GAT+LSTM         & 0.097    &  \\
GCN+LSTM         & 0.131    &  \\
GCN+GAT+LSTM     & 0.087    &  \\
MLP+GAT+LSTM     & 0.088    &  \\
MLP+GCN+LSTM     & 0.088    &  \\
MLP+GAT+GCN+LSTM & 0.084    & 
\end{tabular}
\label{tbl:ablation}
\end{table}

\section{Conclusions}
In this work, we introduce and explore the use of \projectname{}, a spatiotemporal graph neural network model, employing convolution and attention components, for bias correction of storm surge numerical forecasts. In our proposed model, graph nodes are used to describe gauge stations, and node connectivity is defined based on correlations between water level observations at each gauge station. We demonstrate the performance of \projectname{} in predicting 6, 9, 12, 15, 18, 24, 36, 48, and 72 h-ahead biases in ADCIRC water level forecasts at 16 gauge stations along the U.S. Gulf Coast region during Hurricane Idalia (2023). The model is trained on water level bias data from these stations from 11 historical hurricanes (see Table~\ref{tbl:hurricanes}), with all data obtained from the Coastal Emergency Risk Assessement Historical Storm Archive~\cite{CERA2023, CERAarchive}. 

Overall, the model is found to consistently yield low RMSE values even at increased prediction windows, with the median RMSE saturating to 0.9 m with prediction window increasing up to 72 h, corresponding to an improvement of more than 50\% compared to the non-corrected water level predictions. The physics-based graph representation enables the model to adequately capture biases in gauge stations located at both severely storm-impacted and unaffected regions. An ablation analysis assessing architectures composed of different component subsets demonstrates the optimality of the design choices adopted for \projectname{}. 
Furthermore, the model is shown to yield similar accuracy compared to the previous LSTM-based bias prediction method of~\cite{giaremis2024storm} for short prediction windows and the considered gauge stations, but lower RMSE by \(10\%\) for prediction windows greater than 24 h. However, it requires shorter training times in all cases, making it promising for operational purposes. 

Note that this approach can only be used to perform bias correction on coordinates with both observed and numerically predicted data available, and further requires comparable amounts of training data to be available for each of the considered locations. Moreover, physical insight is incorporated in the model via the construction of the graph representation, but not on the learning process. Therefore, future work is intended for developing approaches for assessing biases in ungauged locations in a physically meaningful manner.

\projectname{} can be used as a post-processing bias correction module for gauged locations in real-time operational forecasting systems, due to its consistent performance at increased prediction windows and its low inference time. Its ability to consistently alleviate systematic biases in storm conditions can improve the implementation of warning, prevention, and evacuation strategies during such critical conditions.

\backmatter

\bmhead{Supplementary information}

Additional table including the values of the input window length that yield the lowest RMSE for each of the values considered for the prediction window.

\section*{Statements and Declarations}

We acknowledge the support of the Department of Energy (DoE) through the award DE-SC0022320 (MuSiKAL). We also acknowledge the support of the National Science Foundation (NSF) through award 2347449 (NSF POSE: Phase II: Nexus: Harnessing open High Performance Computing (HPC) through HPX). We would further like to thank  Louisiana State University (LSU) and the Center for Computation and Technology at LSU for granting allocations for their high performance computing resources and storage space. The authors have no relevant financial or non-financial interests to disclose.

\section*{Conflict of 
Interest}
On behalf of all authors, the corresponding author states that there is no conflict of interest.


\begin{thebibliography}{101}
\ifx \bisbn   \undefined \def \bisbn  #1{ISBN #1}\fi
\ifx \binits  \undefined \def \binits#1{#1}\fi
\ifx \bauthor  \undefined \def \bauthor#1{#1}\fi
\ifx \batitle  \undefined \def \batitle#1{#1}\fi
\ifx \bjtitle  \undefined \def \bjtitle#1{#1}\fi
\ifx \bvolume  \undefined \def \bvolume#1{\textbf{#1}}\fi
\ifx \byear  \undefined \def \byear#1{#1}\fi
\ifx \bissue  \undefined \def \bissue#1{#1}\fi
\ifx \bfpage  \undefined \def \bfpage#1{#1}\fi
\ifx \blpage  \undefined \def \blpage #1{#1}\fi
\ifx \burl  \undefined \def \burl#1{\textsf{#1}}\fi
\ifx \doiurl  \undefined \def \doiurl#1{\url{https://doi.org/#1}}\fi
\ifx \betal  \undefined \def \betal{\textit{et al.}}\fi
\ifx \binstitute  \undefined \def \binstitute#1{#1}\fi
\ifx \binstitutionaled  \undefined \def \binstitutionaled#1{#1}\fi
\ifx \bctitle  \undefined \def \bctitle#1{#1}\fi
\ifx \beditor  \undefined \def \beditor#1{#1}\fi
\ifx \bpublisher  \undefined \def \bpublisher#1{#1}\fi
\ifx \bbtitle  \undefined \def \bbtitle#1{#1}\fi
\ifx \bedition  \undefined \def \bedition#1{#1}\fi
\ifx \bseriesno  \undefined \def \bseriesno#1{#1}\fi
\ifx \blocation  \undefined \def \blocation#1{#1}\fi
\ifx \bsertitle  \undefined \def \bsertitle#1{#1}\fi
\ifx \bsnm \undefined \def \bsnm#1{#1}\fi
\ifx \bsuffix \undefined \def \bsuffix#1{#1}\fi
\ifx \bparticle \undefined \def \bparticle#1{#1}\fi
\ifx \barticle \undefined \def \barticle#1{#1}\fi
\bibcommenthead
\ifx \bconfdate \undefined \def \bconfdate #1{#1}\fi
\ifx \botherref \undefined \def \botherref #1{#1}\fi
\ifx \url \undefined \def \url#1{\textsf{#1}}\fi
\ifx \bchapter \undefined \def \bchapter#1{#1}\fi
\ifx \bbook \undefined \def \bbook#1{#1}\fi
\ifx \bcomment \undefined \def \bcomment#1{#1}\fi
\ifx \oauthor \undefined \def \oauthor#1{#1}\fi
\ifx \citeauthoryear \undefined \def \citeauthoryear#1{#1}\fi
\ifx \endbibitem  \undefined \def \endbibitem {}\fi
\ifx \bconflocation  \undefined \def \bconflocation#1{#1}\fi
\ifx \arxivurl  \undefined \def \arxivurl#1{\textsf{#1}}\fi
\csname PreBibitemsHook\endcsname

\bibitem[\protect\citeauthoryear{Smith}{2020}]{NCEI}
\begin{botherref}
\oauthor{\bsnm{Smith}, \binits{A.B.}}:
{U.S.} Billion-dollar Weather and Climate Disasters, 1980 - present ({NCEI Accession} 0209268). {NOAA National Centers for Environmental Information}. Dataset
(2020).
\doiurl{10.25921/STKW-7W73} .
\url{https://www.ncei.noaa.gov/access/billions/summary-stats/US/2004-2024}
\end{botherref}
\endbibitem

\bibitem[\protect\citeauthoryear{Calafat et~al.}{2022}]{Calafat2022}
\begin{barticle}
\bauthor{\bsnm{Calafat}, \binits{F.M.}},
\bauthor{\bsnm{Wahl}, \binits{T.}},
\bauthor{\bsnm{Tadesse}, \binits{M.G.}},
\bauthor{\bsnm{Sparrow}, \binits{S.N.}}:
\batitle{Trends in europe storm surge extremes match the rate of sea-level rise}.
\bjtitle{Nature}
\bvolume{603}(\bissue{7903}),
\bfpage{841}--\blpage{845}
(\byear{2022})
\doiurl{10.1038/s41586-022-04426-5}
\end{barticle}
\endbibitem

\bibitem[\protect\citeauthoryear{Wang and Toumi}{2021}]{Wang2021}
\begin{barticle}
\bauthor{\bsnm{Wang}, \binits{S.}},
\bauthor{\bsnm{Toumi}, \binits{R.}}:
\batitle{Recent migration of tropical cyclones toward coasts}.
\bjtitle{Science}
\bvolume{371}(\bissue{6528}),
\bfpage{514}--\blpage{517}
(\byear{2021})
\doiurl{10.1126/science.abb9038}
\end{barticle}
\endbibitem

\bibitem[\protect\citeauthoryear{Hall and Kossin}{2019}]{Hall2019}
\begin{botherref}
\oauthor{\bsnm{Hall}, \binits{T.M.}},
\oauthor{\bsnm{Kossin}, \binits{J.P.}}:
Hurricane stalling along the {North American} coast and implications for rainfall.
npj Climate and Atmospheric Science
\textbf{2}(1)
(2019)
\doiurl{10.1038/s41612-019-0074-8}
\end{botherref}
\endbibitem

\bibitem[\protect\citeauthoryear{Patricola and Wehner}{2018}]{Patricola2018}
\begin{barticle}
\bauthor{\bsnm{Patricola}, \binits{C.M.}},
\bauthor{\bsnm{Wehner}, \binits{M.F.}}:
\batitle{Anthropogenic influences on major tropical cyclone events}.
\bjtitle{Nature}
\bvolume{563}(\bissue{7731}),
\bfpage{339}--\blpage{346}
(\byear{2018})
\doiurl{10.1038/s41586-018-0673-2}
\end{barticle}
\endbibitem

\bibitem[\protect\citeauthoryear{Li and Chakraborty}{2020}]{Li2020}
\begin{barticle}
\bauthor{\bsnm{Li}, \binits{L.}},
\bauthor{\bsnm{Chakraborty}, \binits{P.}}:
\batitle{Slower decay of landfalling hurricanes in a warming world}.
\bjtitle{Nature}
\bvolume{587}(\bissue{7833}),
\bfpage{230}--\blpage{234}
(\byear{2020})
\doiurl{10.1038/s41586-020-2867-7}
\end{barticle}
\endbibitem

\bibitem[\protect\citeauthoryear{Balaguru et~al.}{2024}]{Balaguru2024}
\begin{botherref}
\oauthor{\bsnm{Balaguru}, \binits{K.}},
\oauthor{\bsnm{Chang}, \binits{C.}},
\oauthor{\bsnm{Leung}, \binits{L.R.}},
\oauthor{\bsnm{Foltz}, \binits{G.R.}},
\oauthor{\bsnm{Hagos}, \binits{S.M.}},
\oauthor{\bsnm{Wehner}, \binits{M.F.}},
\oauthor{\bsnm{Kossin}, \binits{J.P.}},
\oauthor{\bsnm{Ting}, \binits{M.}},
\oauthor{\bsnm{Xu}, \binits{W.}}:
A global increase in nearshore tropical cyclone intensification.
Earth‚Äôs Future
\textbf{12}(5)
(2024)
\doiurl{10.1029/2023ef004230}
\end{botherref}
\endbibitem

\bibitem[\protect\citeauthoryear{Li et~al.}{2023}]{Li2023}
\begin{barticle}
\bauthor{\bsnm{Li}, \binits{X.}},
\bauthor{\bsnm{Zhan}, \binits{R.}},
\bauthor{\bsnm{Wang}, \binits{Y.}},
\bauthor{\bsnm{Zhao}, \binits{J.}},
\bauthor{\bsnm{Ding}, \binits{Y.}},
\bauthor{\bsnm{Song}, \binits{K.}}:
\batitle{Recent increase in rapid intensification events of tropical cyclones along {China} coast}.
\bjtitle{Climate Dynamics}
\bvolume{62}(\bissue{1}),
\bfpage{331}--\blpage{344}
(\byear{2023})
\doiurl{10.1007/s00382-023-06917-1}
\end{barticle}
\endbibitem

\bibitem[\protect\citeauthoryear{Shi et~al.}{2025}]{shi2025intensification}
\begin{botherref}
\oauthor{\bsnm{Shi}, \binits{J.}},
\oauthor{\bsnm{Hu}, \binits{C.}},
\oauthor{\bsnm{Cannizzaro}, \binits{J.}},
\oauthor{\bsnm{Barnes}, \binits{B.B.}},
\oauthor{\bsnm{Zhang}, \binits{Y.}},
\oauthor{\bsnm{Lembke}, \binits{C.}},
\oauthor{\bsnm{Le~Henaff}, \binits{M.}}:
Intensification of hurricane idalia by a river plume in the eastern gulf of mexico.
Environmental Research Letters
(2025)
\end{botherref}
\endbibitem

\bibitem[\protect\citeauthoryear{Liu et~al.}{2024}]{Liu2024}
\begin{botherref}
\oauthor{\bsnm{Liu}, \binits{Y.}},
\oauthor{\bsnm{Weisberg}, \binits{R.H.}},
\oauthor{\bsnm{Sorinas}, \binits{L.}},
\oauthor{\bsnm{Law}, \binits{J.A.}},
\oauthor{\bsnm{Nickerson}, \binits{A.K.}}:
Rapid intensification of {Hurricane Ian} in relation to anomalously warm subsurface water on the wide continental shelf.
Geophysical Research Letters
\textbf{52}(1)
(2024)
\doiurl{10.1029/2024gl113192}
\end{botherref}
\endbibitem

\bibitem[\protect\citeauthoryear{Zhu et~al.}{2022}]{Zhu2022}
\begin{barticle}
\bauthor{\bsnm{Zhu}, \binits{Y.-J.}},
\bauthor{\bsnm{Collins}, \binits{J.M.}},
\bauthor{\bsnm{Klotzbach}, \binits{P.J.}},
\bauthor{\bsnm{Schreck}, \binits{C.J.}}:
\batitle{{Hurricane Ida} (2021): Rapid intensification followed by slow inland decay}.
\bjtitle{Bulletin of the American Meteorological Society}
\bvolume{103}(\bissue{10}),
\bfpage{2354}--\blpage{2369}
(\byear{2022})
\doiurl{10.1175/bams-d-21-0240.1}
\end{barticle}
\endbibitem

\bibitem[\protect\citeauthoryear{Kotal et~al.}{2024}]{Kotal2024}
\begin{barticle}
\bauthor{\bsnm{Kotal}, \binits{S.D.}},
\bauthor{\bsnm{Arulalan}, \binits{T.}},
\bauthor{\bsnm{Mohapatra}, \binits{M.}}:
\batitle{Forecasting of tropical cyclones {ASANI} (2022) and {MOCHA} (2023) over the {Bay of Bengal} - real time challenges to forecasters}.
\bjtitle{Tropical Cyclone Research and Review}
\bvolume{13}(\bissue{2}),
\bfpage{88}--\blpage{112}
(\byear{2024})
\doiurl{10.1016/j.tcrr.2024.06.002}
\end{barticle}
\endbibitem

\bibitem[\protect\citeauthoryear{Petilla et~al.}{2025}]{Petilla2025}
\begin{barticle}
\bauthor{\bsnm{Petilla}, \binits{C.E.R.}},
\bauthor{\bsnm{Olaguera}, \binits{L.M.P.}},
\bauthor{\bsnm{Cruz}, \binits{F.A.T.}},
\bauthor{\bsnm{Villarin}, \binits{J.R.T.}},
\bauthor{\bsnm{Fudeyasu}, \binits{H.}},
\bauthor{\bsnm{Yoshida}, \binits{R.}},
\bauthor{\bsnm{Matsumoto}, \binits{J.}}:
\batitle{The unique features of typhoon rai (2021): an observational study}.
\bjtitle{Natural Hazards}
\bvolume{121}(\bissue{7}),
\bfpage{8279}--\blpage{8303}
(\byear{2025})
\doiurl{10.1007/s11069-025-07138-x}
\end{barticle}
\endbibitem

\bibitem[\protect\citeauthoryear{Zhang et~al.}{2019}]{Zhang2019}
\begin{barticle}
\bauthor{\bsnm{Zhang}, \binits{Z.}},
\bauthor{\bsnm{Wang}, \binits{Y.}},
\bauthor{\bsnm{Zhang}, \binits{W.}},
\bauthor{\bsnm{Xu}, \binits{J.}}:
\batitle{Coastal ocean response and its feedback to {Typhoon Hato} (2017) over the {South China Sea}: A numerical study}.
\bjtitle{Journal of Geophysical Research: Atmospheres}
\bvolume{124}(\bissue{24}),
\bfpage{13731}--\blpage{13749}
(\byear{2019})
\doiurl{10.1029/2019jd031377}
\end{barticle}
\endbibitem

\bibitem[\protect\citeauthoryear{Chang and Wu}{2017}]{Chang2017}
\begin{barticle}
\bauthor{\bsnm{Chang}, \binits{C.-C.}},
\bauthor{\bsnm{Wu}, \binits{C.-C.}}:
\batitle{On the processes leading to the rapid intensification of {Typhoon Megi} (2010)}.
\bjtitle{Journal of the Atmospheric Sciences}
\bvolume{74}(\bissue{4}),
\bfpage{1169}--\blpage{1200}
(\byear{2017})
\doiurl{10.1175/jas-d-16-0075.1}
\end{barticle}
\endbibitem

\bibitem[\protect\citeauthoryear{Wu et~al.}{2015}]{Wu2015}
\begin{barticle}
\bauthor{\bsnm{Wu}, \binits{L.}},
\bauthor{\bsnm{Su}, \binits{H.}},
\bauthor{\bsnm{Fovell}, \binits{R.G.}},
\bauthor{\bsnm{Dunkerton}, \binits{T.J.}},
\bauthor{\bsnm{Wang}, \binits{Z.}},
\bauthor{\bsnm{Kahn}, \binits{B.H.}}:
\batitle{Impact of environmental moisture on tropical cyclone intensification}.
\bjtitle{Atmospheric Chemistry and Physics}
\bvolume{15}(\bissue{24}),
\bfpage{14041}--\blpage{14053}
(\byear{2015})
\doiurl{10.5194/acp-15-14041-2015}
\end{barticle}
\endbibitem

\bibitem[\protect\citeauthoryear{Zagrodnik and Jiang}{2014}]{Zagrodnik2014}
\begin{barticle}
\bauthor{\bsnm{Zagrodnik}, \binits{J.P.}},
\bauthor{\bsnm{Jiang}, \binits{H.}}:
\batitle{Rainfall, convection, and latent heating distributions in rapidly intensifying tropical cyclones}.
\bjtitle{Journal of the Atmospheric Sciences}
\bvolume{71}(\bissue{8}),
\bfpage{2789}--\blpage{2809}
(\byear{2014})
\doiurl{10.1175/jas-d-13-0314.1}
\end{barticle}
\endbibitem

\bibitem[\protect\citeauthoryear{Wu et~al.}{2025}]{Wu2025}
\begin{botherref}
\oauthor{\bsnm{Wu}, \binits{X.}},
\oauthor{\bsnm{Hoffmann}, \binits{L.}},
\oauthor{\bsnm{Wright}, \binits{C.J.}},
\oauthor{\bsnm{Hindley}, \binits{N.P.}},
\oauthor{\bsnm{Alexander}, \binits{M.J.}},
\oauthor{\bsnm{Wang}, \binits{X.}},
\oauthor{\bsnm{Chen}, \binits{B.}},
\oauthor{\bsnm{Wang}, \binits{Y.}},
\oauthor{\bsnm{Li}, \binits{M.}}:
Mechanisms linking stratospheric gravity wave activity to hurricane intensification: Insights from model simulation of {Hurricane Joaquin}.
Geophysical Research Letters
\textbf{52}(10)
(2025)
\doiurl{10.1029/2024gl113531}
\end{botherref}
\endbibitem

\bibitem[\protect\citeauthoryear{Yang et~al.}{2024}]{Yang2024}
\begin{barticle}
\bauthor{\bsnm{Yang}, \binits{S.}},
\bauthor{\bsnm{Shin}, \binits{D.}},
\bauthor{\bsnm{Cocke}, \binits{S.}},
\bauthor{\bsnm{Nam}, \binits{C.C.}},
\bauthor{\bsnm{Bourassa}, \binits{M.}},
\bauthor{\bsnm{Cha}, \binits{D.-H.}},
\bauthor{\bsnm{Kim}, \binits{B.-M.}}:
\batitle{Unveiling the pivotal influence of sea spray heat fluxes on hurricane rapid intensification}.
\bjtitle{Environmental Research Letters}
\bvolume{19}(\bissue{11}),
\bfpage{114058}
(\byear{2024})
\doiurl{10.1088/1748-9326/ad7ee0}
\end{barticle}
\endbibitem

\bibitem[\protect\citeauthoryear{Kim et~al.}{2024}]{Kim2024}
\begin{botherref}
\oauthor{\bsnm{Kim}, \binits{J.-H.}},
\oauthor{\bsnm{Ham}, \binits{Y.-G.}},
\oauthor{\bsnm{Kim}, \binits{D.}},
\oauthor{\bsnm{Li}, \binits{T.}},
\oauthor{\bsnm{Ma}, \binits{C.}}:
Improvement in forecasting short-term tropical cyclone intensity change and their rapid intensification using deep learning.
Artificial Intelligence for the Earth Systems
\textbf{3}(2)
(2024)
\doiurl{10.1175/aies-d-23-0052.1}
\end{botherref}
\endbibitem

\bibitem[\protect\citeauthoryear{Cangialosi et~al.}{2020}]{Cangialosi2020}
\begin{barticle}
\bauthor{\bsnm{Cangialosi}, \binits{J.P.}},
\bauthor{\bsnm{Blake}, \binits{E.}},
\bauthor{\bsnm{DeMaria}, \binits{M.}},
\bauthor{\bsnm{Penny}, \binits{A.}},
\bauthor{\bsnm{Latto}, \binits{A.}},
\bauthor{\bsnm{Rappaport}, \binits{E.}},
\bauthor{\bsnm{Tallapragada}, \binits{V.}}:
\batitle{Recent progress in tropical cyclone intensity forecasting at the {National Hurricane Center}}.
\bjtitle{Weather and Forecasting}
\bvolume{35}(\bissue{5}),
\bfpage{1913}--\blpage{1922}
(\byear{2020})
\doiurl{10.1175/waf-d-20-0059.1}
\end{barticle}
\endbibitem

\bibitem[\protect\citeauthoryear{Trabing and Bell}{2020}]{Trabing2020}
\begin{barticle}
\bauthor{\bsnm{Trabing}, \binits{B.C.}},
\bauthor{\bsnm{Bell}, \binits{M.M.}}:
\batitle{Understanding error distributions of hurricane intensity forecasts during rapid intensity changes}.
\bjtitle{Weather and Forecasting}
\bvolume{35}(\bissue{6}),
\bfpage{2219}--\blpage{2234}
(\byear{2020})
\doiurl{10.1175/waf-d-19-0253.1}
\end{barticle}
\endbibitem

\bibitem[\protect\citeauthoryear{Cyriac et~al.}{2018}]{Cyriac2018}
\begin{barticle}
\bauthor{\bsnm{Cyriac}, \binits{R.}},
\bauthor{\bsnm{Dietrich}, \binits{J.C.}},
\bauthor{\bsnm{Fleming}, \binits{J.G.}},
\bauthor{\bsnm{Blanton}, \binits{B.O.}},
\bauthor{\bsnm{Kaiser}, \binits{C.}},
\bauthor{\bsnm{Dawson}, \binits{C.N.}},
\bauthor{\bsnm{Luettich}, \binits{R.A.}}:
\batitle{Variability in coastal flooding predictions due to forecast errors during {Hurricane} {Arthur}}.
\bjtitle{Coastal Engineering}
\bvolume{137},
\bfpage{59}--\blpage{78}
(\byear{2018})
\doiurl{10.1016/j.coastaleng.2018.02.008}
\end{barticle}
\endbibitem

\bibitem[\protect\citeauthoryear{Turner et~al.}{2024}]{Turner2024}
\begin{barticle}
\bauthor{\bsnm{Turner}, \binits{I.L.}},
\bauthor{\bsnm{Leaman}, \binits{C.K.}},
\bauthor{\bsnm{Harley}, \binits{M.D.}},
\bauthor{\bsnm{Thran}, \binits{M.C.}},
\bauthor{\bsnm{David}, \binits{D.R.}},
\bauthor{\bsnm{Splinter}, \binits{K.D.}},
\bauthor{\bsnm{Matheen}, \binits{N.}},
\bauthor{\bsnm{Hansen}, \binits{J.E.}},
\bauthor{\bsnm{Cuttler}, \binits{M.V.W.}},
\bauthor{\bsnm{Greenslade}, \binits{D.J.M.}},
\bauthor{\bsnm{Zieger}, \binits{S.}},
\bauthor{\bsnm{Lowe}, \binits{R.J.}}:
\batitle{A framework for national-scale coastal storm hazards early warning}.
\bjtitle{Coastal Engineering}
\bvolume{192},
\bfpage{104571}
(\byear{2024})
\doiurl{10.1016/j.coastaleng.2024.104571}
\end{barticle}
\endbibitem

\bibitem[\protect\citeauthoryear{Penny et~al.}{2023}]{Penny2023}
\begin{barticle}
\bauthor{\bsnm{Penny}, \binits{A.B.}},
\bauthor{\bsnm{Alaka}, \binits{L.}},
\bauthor{\bsnm{Taylor}, \binits{A.A.}},
\bauthor{\bsnm{Booth}, \binits{W.}},
\bauthor{\bsnm{DeMaria}, \binits{M.}},
\bauthor{\bsnm{Fritz}, \binits{C.}},
\bauthor{\bsnm{Rhome}, \binits{J.}}:
\batitle{Operational storm surge forecasting at the national hurricane center: The case for probabilistic guidance and the evaluation of improved storm size forecasts used to define the wind forcing}.
\bjtitle{Weather and Forecasting}
\bvolume{38}(\bissue{12}),
\bfpage{2461}--\blpage{2479}
(\byear{2023})
\doiurl{10.1175/waf-d-22-0209.1}
\end{barticle}
\endbibitem

\bibitem[\protect\citeauthoryear{Suh et~al.}{2015}]{Suh2015}
\begin{barticle}
\bauthor{\bsnm{Suh}, \binits{S.W.}},
\bauthor{\bsnm{Lee}, \binits{H.Y.}},
\bauthor{\bsnm{Kim}, \binits{H.J.}},
\bauthor{\bsnm{Fleming}, \binits{J.G.}}:
\batitle{An efficient early warning system for typhoon storm surge based on time-varying advisories by coupled {ADCIRC} and {SWAN}}.
\bjtitle{Ocean Dynamics}
\bvolume{65}(\bissue{5}),
\bfpage{617}--\blpage{646}
(\byear{2015})
\doiurl{10.1007/s10236-015-0820-3}
\end{barticle}
\endbibitem

\bibitem[\protect\citeauthoryear{Westerink et~al.}{1992}]{Westerink1992}
\begin{barticle}
\bauthor{\bsnm{Westerink}, \binits{J.J.}},
\bauthor{\bsnm{Luettich}, \binits{R.A.}},
\bauthor{\bsnm{Baptists}, \binits{A.M.}},
\bauthor{\bsnm{Scheffner}, \binits{N.W.}},
\bauthor{\bsnm{Farrar}, \binits{P.}}:
\batitle{Tide and storm surge predictions using finite element model}.
\bjtitle{Journal of Hydraulic Engineering}
\bvolume{118}(\bissue{10}),
\bfpage{1373}--\blpage{1390}
(\byear{1992})
\doiurl{10.1061/(asce)0733-9429(1992)118:10(1373)}
\end{barticle}
\endbibitem

\bibitem[\protect\citeauthoryear{Booij et~al.}{1999}]{Booij1999}
\begin{barticle}
\bauthor{\bsnm{Booij}, \binits{N.}},
\bauthor{\bsnm{Ris}, \binits{R.C.}},
\bauthor{\bsnm{Holthuijsen}, \binits{L.H.}}:
\batitle{A third‚Äêgeneration wave model for coastal regions: 1. model description and validation}.
\bjtitle{Journal of Geophysical Research: Oceans}
\bvolume{104}(\bissue{C4}),
\bfpage{7649}--\blpage{7666}
(\byear{1999})
\doiurl{10.1029/98jc02622}
\end{barticle}
\endbibitem

\bibitem[\protect\citeauthoryear{Dietrich et~al.}{2011}]{Dietrich2011}
\begin{barticle}
\bauthor{\bsnm{Dietrich}, \binits{J.C.}},
\bauthor{\bsnm{Zijlema}, \binits{M.}},
\bauthor{\bsnm{Westerink}, \binits{J.J.}},
\bauthor{\bsnm{Holthuijsen}, \binits{L.H.}},
\bauthor{\bsnm{Dawson}, \binits{C.}},
\bauthor{\bsnm{Luettich}, \binits{R.A.}},
\bauthor{\bsnm{Jensen}, \binits{R.E.}},
\bauthor{\bsnm{Smith}, \binits{J.M.}},
\bauthor{\bsnm{Stelling}, \binits{G.S.}},
\bauthor{\bsnm{Stone}, \binits{G.W.}}:
\batitle{Modeling hurricane waves and storm surge using integrally-coupled, scalable computations}.
\bjtitle{Coastal Engineering}
\bvolume{58}(\bissue{1}),
\bfpage{45}--\blpage{65}
(\byear{2011})
\doiurl{10.1016/j.coastaleng.2010.08.001}
\end{barticle}
\endbibitem

\bibitem[\protect\citeauthoryear{{CERA - Coastal Emergency Risk Accessment}}{2025}]{CERA}
\begin{botherref}
\oauthor{\bsnm{{CERA - Coastal Emergency Risk Accessment}}}:
{Center for Computation and Technology at Louisiana State University}.
\url{https://cera.coastalrisk.live/}
(2025)
\end{botherref}
\endbibitem

\bibitem[\protect\citeauthoryear{Chen et~al.}{2025}]{Chen2025}
\begin{barticle}
\bauthor{\bsnm{Chen}, \binits{F.}},
\bauthor{\bsnm{Yang}, \binits{W.}},
\bauthor{\bsnm{Xiao}, \binits{L.}},
\bauthor{\bsnm{Xia}, \binits{X.}},
\bauthor{\bsnm{Ding}, \binits{K.}},
\bauthor{\bsnm{Sun}, \binits{Z.}}:
\batitle{An exploratory assessment of a submarine topographic characteristic index for predicting extreme flow velocities: A case study of {Typhoon In---Fa} in the {Zhoushan Sea} area}.
\bjtitle{Journal of Marine Science and Engineering}
\bvolume{13}(\bissue{5}),
\bfpage{864}
(\byear{2025})
\doiurl{10.3390/jmse13050864}
\end{barticle}
\endbibitem

\bibitem[\protect\citeauthoryear{Pringle et~al.}{2021}]{Pringle2021}
\begin{barticle}
\bauthor{\bsnm{Pringle}, \binits{W.J.}},
\bauthor{\bsnm{Wirasaet}, \binits{D.}},
\bauthor{\bsnm{Roberts}, \binits{K.J.}},
\bauthor{\bsnm{Westerink}, \binits{J.J.}}:
\batitle{Global storm tide modeling with {ADCIRC} v55: unstructured mesh design and performance}.
\bjtitle{Geoscientific Model Development}
\bvolume{14}(\bissue{2}),
\bfpage{1125}--\blpage{1145}
(\byear{2021})
\doiurl{10.5194/gmd-14-1125-2021}
\end{barticle}
\endbibitem

\bibitem[\protect\citeauthoryear{Khani and Dawson}{2023}]{Khani2023}
\begin{botherref}
\oauthor{\bsnm{Khani}, \binits{S.}},
\oauthor{\bsnm{Dawson}, \binits{C.N.}}:
A gradient based subgrid‚Äêscale parameterization for ocean mesoscale eddies.
Journal of Advances in Modeling Earth Systems
\textbf{15}(2)
(2023)
\doiurl{10.1029/2022ms003356}
\end{botherref}
\endbibitem

\bibitem[\protect\citeauthoryear{Blakely et~al.}{2022}]{Blakely2022}
\begin{botherref}
\oauthor{\bsnm{Blakely}, \binits{C.P.}},
\oauthor{\bsnm{Ling}, \binits{G.}},
\oauthor{\bsnm{Pringle}, \binits{W.J.}},
\oauthor{\bsnm{Contreras}, \binits{M.T.}},
\oauthor{\bsnm{Wirasaet}, \binits{D.}},
\oauthor{\bsnm{Westerink}, \binits{J.J.}},
\oauthor{\bsnm{Moghimi}, \binits{S.}},
\oauthor{\bsnm{Seroka}, \binits{G.}},
\oauthor{\bsnm{Shi}, \binits{L.}},
\oauthor{\bsnm{Myers}, \binits{E.}},
\oauthor{\bsnm{Owensby}, \binits{M.}},
\oauthor{\bsnm{Massey}, \binits{C.}}:
Dissipation and bathymetric sensitivities in an unstructured mesh global tidal model.
Journal of Geophysical Research: Oceans
\textbf{127}(5)
(2022)
\doiurl{10.1029/2021jc018178}
\end{botherref}
\endbibitem

\bibitem[\protect\citeauthoryear{Loveland et~al.}{2024}]{Loveland2024}
\begin{barticle}
\bauthor{\bsnm{Loveland}, \binits{M.}},
\bauthor{\bsnm{Meixner}, \binits{J.}},
\bauthor{\bsnm{Valseth}, \binits{E.}},
\bauthor{\bsnm{Dawson}, \binits{C.}}:
\batitle{Efficacy of reduced order source terms for a coupled wave-circulation model in the {Gulf} of {Mexico}}.
\bjtitle{Ocean Modelling}
\bvolume{190},
\bfpage{102387}
(\byear{2024})
\doiurl{10.1016/j.ocemod.2024.102387}
\end{barticle}
\endbibitem

\bibitem[\protect\citeauthoryear{Dawson et~al.}{2024}]{Dawson2024}
\begin{botherref}
\oauthor{\bsnm{Dawson}, \binits{C.}},
\oauthor{\bsnm{Loveland}, \binits{M.}},
\oauthor{\bsnm{Pachev}, \binits{B.}},
\oauthor{\bsnm{Proft}, \binits{J.}},
\oauthor{\bsnm{Valseth}, \binits{E.}}:
{SWEMniCS}: a software toolbox for modeling coastal ocean circulation, storm surges, inland, and compound flooding.
npj Natural Hazards
\textbf{1}(1)
(2024)
\doiurl{10.1038/s44304-024-00036-5}
\end{botherref}
\endbibitem

\bibitem[\protect\citeauthoryear{Bernier et~al.}{2024}]{Bernier2024}
\begin{barticle}
\bauthor{\bsnm{Bernier}, \binits{N.B.}},
\bauthor{\bsnm{Hemer}, \binits{M.}},
\bauthor{\bsnm{Mori}, \binits{N.}},
\bauthor{\bsnm{Appendini}, \binits{C.M.}},
\bauthor{\bsnm{Breivik}, \binits{O.}},
\bauthor{\bsnm{Camargo}, \binits{R.}},
\bauthor{\bsnm{Casas-Prat}, \binits{M.}},
\bauthor{\bsnm{Duong}, \binits{T.M.}},
\bauthor{\bsnm{Haigh}, \binits{I.D.}},
\bauthor{\bsnm{Howard}, \binits{T.}},
\bauthor{\bsnm{Hernaman}, \binits{V.}},
\bauthor{\bsnm{Huizy}, \binits{O.}},
\bauthor{\bsnm{Irish}, \binits{J.L.}},
\bauthor{\bsnm{Kirezci}, \binits{E.}},
\bauthor{\bsnm{Kohno}, \binits{N.}},
\bauthor{\bsnm{Lee}, \binits{J.-W.}},
\bauthor{\bsnm{McInnes}, \binits{K.L.}},
\bauthor{\bsnm{Meyer}, \binits{E.M.I.}},
\bauthor{\bsnm{Marcos}, \binits{M.}},
\bauthor{\bsnm{Marsooli}, \binits{R.}},
\bauthor{\bsnm{Martin~Oliva}, \binits{A.}},
\bauthor{\bsnm{Menendez}, \binits{M.}},
\bauthor{\bsnm{Moghimi}, \binits{S.}},
\bauthor{\bsnm{Muis}, \binits{S.}},
\bauthor{\bsnm{Polton}, \binits{J.A.}},
\bauthor{\bsnm{Pringle}, \binits{W.J.}},
\bauthor{\bsnm{Ranasinghe}, \binits{R.}},
\bauthor{\bsnm{Saillour}, \binits{T.}},
\bauthor{\bsnm{Smith}, \binits{G.}},
\bauthor{\bsnm{Tadesse}, \binits{M.G.}},
\bauthor{\bsnm{Swail}, \binits{V.}},
\bauthor{\bsnm{Tomoya}, \binits{S.}},
\bauthor{\bsnm{Voukouvalas}, \binits{E.}},
\bauthor{\bsnm{Wahl}, \binits{T.}},
\bauthor{\bsnm{Wang}, \binits{P.}},
\bauthor{\bsnm{Weisse}, \binits{R.}},
\bauthor{\bsnm{Westerink}, \binits{J.J.}},
\bauthor{\bsnm{Young}, \binits{I.}},
\bauthor{\bsnm{Zhang}, \binits{Y.J.}}:
\batitle{Storm surges and extreme sea levels: Review, establishment of model intercomparison and coordination of surge climate projection efforts ({SurgeMIP}).}
\bjtitle{Weather and Climate Extremes}
\bvolume{45},
\bfpage{100689}
(\byear{2024})
\doiurl{10.1016/j.wace.2024.100689}
\end{barticle}
\endbibitem

\bibitem[\protect\citeauthoryear{Loveland et~al.}{2021}]{Loveland2021}
\begin{botherref}
\oauthor{\bsnm{Loveland}, \binits{M.}},
\oauthor{\bsnm{Kiaghadi}, \binits{A.}},
\oauthor{\bsnm{Dawson}, \binits{C.N.}},
\oauthor{\bsnm{Rifai}, \binits{H.S.}},
\oauthor{\bsnm{Misra}, \binits{S.}},
\oauthor{\bsnm{Mosser}, \binits{H.}},
\oauthor{\bsnm{Parola}, \binits{A.}}:
Developing a modeling framework to simulate compound flooding: When storm surge interacts with riverine flow.
Frontiers in Climate
\textbf{2}
(2021)
\doiurl{10.3389/fclim.2020.609610}
\end{botherref}
\endbibitem

\bibitem[\protect\citeauthoryear{Wei et~al.}{2024}]{Wei2024}
\begin{botherref}
\oauthor{\bsnm{Wei}, \binits{W.}},
\oauthor{\bsnm{Huang}, \binits{S.}},
\oauthor{\bsnm{Qin}, \binits{H.}},
\oauthor{\bsnm{Yu}, \binits{L.}},
\oauthor{\bsnm{Mu}, \binits{L.}}:
Storm surge risk assessment and sensitivity analysis based on multiple criteria decision-making methods: a case study of {Huizhou} city.
Frontiers in Marine Science
\textbf{11}
(2024)
\doiurl{10.3389/fmars.2024.1364929}
\end{botherref}
\endbibitem

\bibitem[\protect\citeauthoryear{Zhang et~al.}{2023}]{Zhang2023}
\begin{barticle}
\bauthor{\bsnm{Zhang}, \binits{Z.}},
\bauthor{\bsnm{Lu}, \binits{Y.}},
\bauthor{\bsnm{Hu}, \binits{D.}},
\bauthor{\bsnm{Guo}, \binits{F.}},
\bauthor{\bsnm{Yu}, \binits{Z.}},
\bauthor{\bsnm{Song}, \binits{Z.}},
\bauthor{\bsnm{Chen}, \binits{P.}},
\bauthor{\bsnm{Wu}, \binits{J.}},
\bauthor{\bsnm{Huang}, \binits{W.}}:
\batitle{A cross-scale modeling framework for simulating typhoon-induced compound floods and assessing the emergency response in urban regions}.
\bjtitle{Ocean \& Coastal Management}
\bvolume{245},
\bfpage{106863}
(\byear{2023})
\doiurl{10.1016/j.ocecoaman.2023.106863}
\end{barticle}
\endbibitem

\bibitem[\protect\citeauthoryear{Huang et~al.}{2021}]{Huang2021}
\begin{barticle}
\bauthor{\bsnm{Huang}, \binits{W.}},
\bauthor{\bsnm{Yin}, \binits{K.}},
\bauthor{\bsnm{Ghorbanzadeh}, \binits{M.}},
\bauthor{\bsnm{Ozguven}, \binits{E.}},
\bauthor{\bsnm{Xu}, \binits{S.}},
\bauthor{\bsnm{Vijayan}, \binits{L.}}:
\batitle{Integrating storm surge modeling with traffic data analysis to evaluate the effectiveness of hurricane evacuation}.
\bjtitle{Frontiers of Structural and Civil Engineering}
\bvolume{15}(\bissue{6}),
\bfpage{1301}--\blpage{1316}
(\byear{2021})
\doiurl{10.1007/s11709-021-0765-1}
\end{barticle}
\endbibitem

\bibitem[\protect\citeauthoryear{\"{O}zkan et~al.}{2025}]{Ozkan2025}
\begin{botherref}
\oauthor{\bsnm{\"{O}zkan}, \binits{F.N.}},
\oauthor{\bsnm{Verlaan}, \binits{M.}},
\oauthor{\bsnm{Muis}, \binits{S.}},
\oauthor{\bsnm{Zijl}, \binits{F.}}:
Sensitivity of global storm surge modelling to sea surface drag.
Ocean Dynamics
\textbf{75}(8)
(2025)
\doiurl{10.1007/s10236-025-01713-3}
\end{botherref}
\endbibitem

\bibitem[\protect\citeauthoryear{Mu√±oz et~al.}{2022}]{Munoz2022}
\begin{barticle}
\bauthor{\bsnm{Mu√±oz}, \binits{D.F.}},
\bauthor{\bsnm{Abbaszadeh}, \binits{P.}},
\bauthor{\bsnm{Moftakhari}, \binits{H.}},
\bauthor{\bsnm{Moradkhani}, \binits{H.}}:
\batitle{Accounting for uncertainties in compound flood hazard assessment: The value of data assimilation}.
\bjtitle{Coastal Engineering}
\bvolume{171},
\bfpage{104057}
(\byear{2022})
\doiurl{10.1016/j.coastaleng.2021.104057}
\end{barticle}
\endbibitem

\bibitem[\protect\citeauthoryear{Torres et~al.}{2019}]{Torres2019}
\begin{botherref}
\oauthor{\bsnm{Torres}, \binits{M.J.}},
\oauthor{\bsnm{Reza~Hashemi}, \binits{M.}},
\oauthor{\bsnm{Hayward}, \binits{S.}},
\oauthor{\bsnm{Spaulding}, \binits{M.}},
\oauthor{\bsnm{Ginis}, \binits{I.}},
\oauthor{\bsnm{Grilli}, \binits{S.T.}}:
Role of hurricane wind models in accurate simulation of storm surge and waves.
Journal of Waterway, Port, Coastal, and Ocean Engineering
\textbf{145}(1)
(2019)
\doiurl{10.1061/(asce)ww.1943-5460.0000496}
\end{botherref}
\endbibitem

\bibitem[\protect\citeauthoryear{Gallien et~al.}{2018}]{Gallien2018}
\begin{barticle}
\bauthor{\bsnm{Gallien}, \binits{T.W.}},
\bauthor{\bsnm{Kalligeris}, \binits{N.}},
\bauthor{\bsnm{Delisle}, \binits{M.-P.C.}},
\bauthor{\bsnm{Tang}, \binits{B.-X.}},
\bauthor{\bsnm{Lucey}, \binits{J.T.D.}},
\bauthor{\bsnm{Winters}, \binits{M.A.}}:
\batitle{Coastal flood modeling challenges in defended urban backshores}.
\bjtitle{Geosciences}
\bvolume{8}(\bissue{12}),
\bfpage{450}
(\byear{2018})
\doiurl{10.3390/geosciences8120450}
\end{barticle}
\endbibitem

\bibitem[\protect\citeauthoryear{Ferreira et~al.}{2014}]{Ferreira2014}
\begin{barticle}
\bauthor{\bsnm{Ferreira}, \binits{C.M.}},
\bauthor{\bsnm{Irish}, \binits{J.L.}},
\bauthor{\bsnm{Olivera}, \binits{F.}}:
\batitle{Uncertainty in hurricane surge simulation due to land cover specification}.
\bjtitle{Journal of Geophysical Research: Oceans}
\bvolume{119}(\bissue{3}),
\bfpage{1812}--\blpage{1827}
(\byear{2014})
\doiurl{10.1002/2013jc009604}
\end{barticle}
\endbibitem

\bibitem[\protect\citeauthoryear{Asher et~al.}{2019}]{Asher2019}
\begin{barticle}
\bauthor{\bsnm{Asher}, \binits{T.G.}},
\bauthor{\bsnm{Luettich~Jr.}, \binits{R.A.}},
\bauthor{\bsnm{Fleming}, \binits{J.G.}},
\bauthor{\bsnm{Blanton}, \binits{B.O.}}:
\batitle{Low frequency water level correction in storm surge models using data assimilation}.
\bjtitle{Ocean Modelling}
\bvolume{144},
\bfpage{101483}
(\byear{2019})
\doiurl{10.1016/j.ocemod.2019.101483}
\end{barticle}
\endbibitem

\bibitem[\protect\citeauthoryear{Gonzalez et~al.}{2019}]{Gonzalez2019}
\begin{botherref}
\oauthor{\bsnm{Gonzalez}, \binits{V.M.}},
\oauthor{\bsnm{Nadal-Caraballo}, \binits{N.C.}},
\oauthor{\bsnm{Melby}, \binits{J.A.}},
\oauthor{\bsnm{Cialone}, \binits{M.A.}}:
Quantification of uncertainty in probabilistic storm surge models: Literature review.
Technical report,
U.S. Army Corps of Engineers, Engineer Research and Development Center
(2019).
\url{https://chs.erdc.dren.mil/Library/References/CHS_PCHA_Publications/Reports/SR-19-1_Gonzalez_et_al_2019_UncertaintyInSurgeModels.pdf}
\end{botherref}
\endbibitem

\bibitem[\protect\citeauthoryear{Resio et~al.}{2012}]{Resio2012}
\begin{barticle}
\bauthor{\bsnm{Resio}, \binits{D.T.}},
\bauthor{\bsnm{Irish}, \binits{J.L.}},
\bauthor{\bsnm{Westerink}, \binits{J.J.}},
\bauthor{\bsnm{Powell}, \binits{N.J.}}:
\batitle{The effect of uncertainty on estimates of hurricane surge hazards}.
\bjtitle{Natural Hazards}
\bvolume{66}(\bissue{3}),
\bfpage{1443}--\blpage{1459}
(\byear{2012})
\doiurl{10.1007/s11069-012-0315-1}
\end{barticle}
\endbibitem

\bibitem[\protect\citeauthoryear{Sweet et~al.}{2018}]{Sweet2018}
\begin{botherref}
\oauthor{\bsnm{Sweet}, \binits{W.V.}},
\oauthor{\bsnm{Obeysekera}, \binits{J.T.B.}},
\oauthor{\bsnm{Marra}, \binits{J.J.}},
\oauthor{\bsnm{Dusek}, \binits{G.}}:
Patterns and projections of high tide flooding along the {U.S.} coastline using a common impact threshold.
(2018)
\doiurl{10.7289/V5/TR-NOS-COOPS-086}
\end{botherref}
\endbibitem

\bibitem[\protect\citeauthoryear{Feng et~al.}{2023}]{Feng2023}
\begin{barticle}
\bauthor{\bsnm{Feng}, \binits{J.}},
\bauthor{\bsnm{Li}, \binits{D.}},
\bauthor{\bsnm{Dang}, \binits{W.}},
\bauthor{\bsnm{Zhao}, \binits{L.}}:
\batitle{Changes in storm surges based on a bias-adjusted reconstruction dataset from 1900 to 2010}.
\bjtitle{Journal of Hydrology}
\bvolume{617},
\bfpage{128759}
(\byear{2023})
\end{barticle}
\endbibitem

\bibitem[\protect\citeauthoryear{Resio et~al.}{2017}]{Resio2017}
\begin{barticle}
\bauthor{\bsnm{Resio}, \binits{D.T.}},
\bauthor{\bsnm{J.~Powell}, \binits{N.}},
\bauthor{\bsnm{A.~Cialone}, \binits{M.}},
\bauthor{\bsnm{Das}, \binits{H.S.}},
\bauthor{\bsnm{Westerink}, \binits{J.J.}}:
\batitle{Quantifying impacts of forecast uncertainties on predicted storm surges}.
\bjtitle{Natural Hazards}
\bvolume{88}(\bissue{3}),
\bfpage{1423}--\blpage{1449}
(\byear{2017})
\doiurl{10.1007/s11069-017-2924-1}
\end{barticle}
\endbibitem

\bibitem[\protect\citeauthoryear{Butler et~al.}{2012}]{Butler2012}
\begin{barticle}
\bauthor{\bsnm{Butler}, \binits{T.}},
\bauthor{\bsnm{Altaf}, \binits{M.U.}},
\bauthor{\bsnm{Dawson}, \binits{C.}},
\bauthor{\bsnm{Hoteit}, \binits{I.}},
\bauthor{\bsnm{Luo}, \binits{X.}},
\bauthor{\bsnm{Mayo}, \binits{T.}}:
\batitle{Data assimilation within the {Advanced Circulation (ADCIRC)} modeling framework for hurricane storm surge forecasting}.
\bjtitle{Monthly Weather Review}
\bvolume{140}(\bissue{7}),
\bfpage{2215}--\blpage{2231}
(\byear{2012})
\doiurl{10.1175/mwr-d-11-00118.1}
\end{barticle}
\endbibitem

\bibitem[\protect\citeauthoryear{Muis et~al.}{2016}]{Muis2016}
\begin{botherref}
\oauthor{\bsnm{Muis}, \binits{S.}},
\oauthor{\bsnm{Verlaan}, \binits{M.}},
\oauthor{\bsnm{Winsemius}, \binits{H.C.}},
\oauthor{\bsnm{Aerts}, \binits{J.C.J.H.}},
\oauthor{\bsnm{Ward}, \binits{P.J.}}:
A global reanalysis of storm surges and extreme sea levels.
Nature Communications
\textbf{7}(1)
(2016)
\doiurl{10.1038/ncomms11969}
\end{botherref}
\endbibitem

\bibitem[\protect\citeauthoryear{Tadesse and Wahl}{2021}]{Tadesse2021}
\begin{botherref}
\oauthor{\bsnm{Tadesse}, \binits{M.G.}},
\oauthor{\bsnm{Wahl}, \binits{T.}}:
A database of global storm surge reconstructions.
Scientific Data
\textbf{8}(1)
(2021)
\doiurl{10.1038/s41597-021-00906-x}
\end{botherref}
\endbibitem

\bibitem[\protect\citeauthoryear{Kaiser et~al.}{2023}]{CERA2023}
\begin{botherref}
\oauthor{\bsnm{Kaiser}, \binits{C.}},
\oauthor{\bsnm{Dawson}, \binits{C.N.}},
\oauthor{\bsnm{Nikidis}, \binits{E.}},
\oauthor{\bsnm{Fleming}, \binits{J.G.}}:
{ADCIRC/SWAN} Hindcasts for Historical Storms 2003-2022.
Designsafe-CI
(2023).
\doiurl{10.17603/DS2-B5GH-CE94} .
\url{https://www.designsafe-ci.org/data/browser/public/designsafe.storage.published/PRJ-3932/#details-5508251847528869395-242ac117-0001-012}
\end{botherref}
\endbibitem

\bibitem[\protect\citeauthoryear{Haigh et~al.}{2022}]{Haigh2022}
\begin{barticle}
\bauthor{\bsnm{Haigh}, \binits{I.D.}},
\bauthor{\bsnm{Marcos}, \binits{M.}},
\bauthor{\bsnm{Talke}, \binits{S.A.}},
\bauthor{\bsnm{Woodworth}, \binits{P.L.}},
\bauthor{\bsnm{Hunter}, \binits{J.R.}},
\bauthor{\bsnm{Hague}, \binits{B.S.}},
\bauthor{\bsnm{Arns}, \binits{A.}},
\bauthor{\bsnm{Bradshaw}, \binits{E.}},
\bauthor{\bsnm{Thompson}, \binits{P.}}:
\batitle{<scp>gesla</scp> version 3: A major update to the global higher‚Äêfrequency sea‚Äêlevel dataset}.
\bjtitle{Geoscience Data Journal}
\bvolume{10}(\bissue{3}),
\bfpage{293}--\blpage{314}
(\byear{2022})
\doiurl{10.1002/gdj3.174}
\end{barticle}
\endbibitem

\bibitem[\protect\citeauthoryear{Soci et~al.}{2024}]{Soci2024}
\begin{barticle}
\bauthor{\bsnm{Soci}, \binits{C.}},
\bauthor{\bsnm{Hersbach}, \binits{H.}},
\bauthor{\bsnm{Simmons}, \binits{A.}},
\bauthor{\bsnm{Poli}, \binits{P.}},
\bauthor{\bsnm{Bell}, \binits{B.}},
\bauthor{\bsnm{Berrisford}, \binits{P.}},
\bauthor{\bsnm{Hor√°nyi}, \binits{A.}},
\bauthor{\bsnm{Mu√±oz‚ÄêSabater}, \binits{J.}},
\bauthor{\bsnm{Nicolas}, \binits{J.}},
\bauthor{\bsnm{Radu}, \binits{R.}},
\bauthor{\bsnm{Schepers}, \binits{D.}},
\bauthor{\bsnm{Villaume}, \binits{S.}},
\bauthor{\bsnm{Haimberger}, \binits{L.}},
\bauthor{\bsnm{Woollen}, \binits{J.}},
\bauthor{\bsnm{Buontempo}, \binits{C.}},
\bauthor{\bsnm{Th√©paut}, \binits{J.}}:
\batitle{The {ERA5} global reanalysis from 1940 to 2022}.
\bjtitle{Quarterly Journal of the Royal Meteorological Society}
\bvolume{150}(\bissue{764}),
\bfpage{4014}--\blpage{4048}
(\byear{2024})
\doiurl{10.1002/qj.4803}
\end{barticle}
\endbibitem

\bibitem[\protect\citeauthoryear{Muis et~al.}{2020}]{Muis2020}
\begin{botherref}
\oauthor{\bsnm{Muis}, \binits{S.}},
\oauthor{\bsnm{Apecechea}, \binits{M.I.}},
\oauthor{\bsnm{Dullaart}, \binits{J.}},
\oauthor{\bsnm{Lima~Rego}, \binits{J.}},
\oauthor{\bsnm{Madsen}, \binits{K.S.}},
\oauthor{\bsnm{Su}, \binits{J.}},
\oauthor{\bsnm{Yan}, \binits{K.}},
\oauthor{\bsnm{Verlaan}, \binits{M.}}:
A high-resolution global dataset of extreme sea levels, tides, and storm surges, including future projections.
Frontiers in Marine Science
\textbf{7}
(2020)
\doiurl{10.3389/fmars.2020.00263}
\end{botherref}
\endbibitem

\bibitem[\protect\citeauthoryear{}{}]{noaaHURDATReanalysis}
\begin{botherref}
{H}{U}{R}{D}{A}{T} {R}e-analysis --- aoml.noaa.gov.
\url{https://www.aoml.noaa.gov/hrd/hurdat/Data_Storm.html}.
[Accessed 03-10-2025]
\end{botherref}
\endbibitem

\bibitem[\protect\citeauthoryear{}{2025}]{CERAarchive}
\begin{botherref}
{C}{E}{R}{A} - {H}istorical {S}torm {A}rchive --- historicalstorms.coastalrisk.live.
\url{https://historicalstorms.coastalrisk.live/}.
[Accessed 13-10-2025]
(2025)
\end{botherref}
\endbibitem

\bibitem[\protect\citeauthoryear{Lu et~al.}{2021}]{Lu2021}
\begin{barticle}
\bauthor{\bsnm{Lu}, \binits{X.}},
\bauthor{\bsnm{Yu}, \binits{H.}},
\bauthor{\bsnm{Ying}, \binits{M.}},
\bauthor{\bsnm{Zhao}, \binits{B.}},
\bauthor{\bsnm{Zhang}, \binits{S.}},
\bauthor{\bsnm{Lin}, \binits{L.}},
\bauthor{\bsnm{Bai}, \binits{L.}},
\bauthor{\bsnm{Wan}, \binits{R.}}:
\batitle{Western north pacific tropical cyclone database created by the china meteorological administration}.
\bjtitle{Advances in Atmospheric Sciences}
\bvolume{38}(\bissue{4}),
\bfpage{690}--\blpage{699}
(\byear{2021})
\doiurl{10.1007/s00376-020-0211-7}
\end{barticle}
\endbibitem

\bibitem[\protect\citeauthoryear{KITAMOTO et~al.}{2023}]{Kitamoto2023}
\begin{bchapter}
\bauthor{\bsnm{KITAMOTO}, \binits{A.}},
\bauthor{\bsnm{HWANG}, \binits{J.}},
\bauthor{\bsnm{VUILLOD}, \binits{B.}},
\bauthor{\bsnm{GAUTIER}, \binits{L.}},
\bauthor{\bsnm{TIAN}, \binits{Y.}},
\bauthor{\bsnm{CLANUWAT}, \binits{T.}}:
\bctitle{Digital typhoon: Long-term satellite image dataset for the spatio-temporal modeling of tropical cyclones}.
In: \bbtitle{NeurIPS 2023 Datasets and Benchmarks (Spotlight)}
(\byear{2023})
\end{bchapter}
\endbibitem

\bibitem[\protect\citeauthoryear{Zhao et~al.}{2025}]{Zhao2025}
\begin{botherref}
\oauthor{\bsnm{Zhao}, \binits{J.}},
\oauthor{\bsnm{Cerrone}, \binits{A.}},
\oauthor{\bsnm{Valseth}, \binits{E.}},
\oauthor{\bsnm{Westerink}, \binits{L.}},
\oauthor{\bsnm{Dawson}, \binits{C.}}:
Storm surge in color: Rgb-encoded physics-aware deep learning for storm surge forecasting.
arXiv preprint arXiv:2506.21743
(2025)
\end{botherref}
\endbibitem

\bibitem[\protect\citeauthoryear{Han et~al.}{2025}]{Han2025}
\begin{barticle}
\bauthor{\bsnm{Han}, \binits{L.}},
\bauthor{\bsnm{Lu}, \binits{W.}},
\bauthor{\bsnm{Dong}, \binits{C.}}:
\batitle{{XAI} helps in storm surge forecasts: A case study for the southeastern chinese coasts}.
\bjtitle{Journal of Marine Science and Engineering}
\bvolume{13}(\bissue{5}),
\bfpage{896}
(\byear{2025})
\doiurl{10.3390/jmse13050896}
\end{barticle}
\endbibitem

\bibitem[\protect\citeauthoryear{Saviz~Naeini et~al.}{2025}]{SavizNaeini2025}
\begin{barticle}
\bauthor{\bsnm{Saviz~Naeini}, \binits{S.}},
\bauthor{\bsnm{Snaiki}, \binits{R.}},
\bauthor{\bsnm{Wu}, \binits{T.}}:
\batitle{Advancing spatio-temporal storm surge prediction with hierarchical deep neural networks}.
\bjtitle{Natural Hazards}
\bvolume{121}(\bissue{14}),
\bfpage{16317}--\blpage{16344}
(\byear{2025})
\doiurl{10.1007/s11069-025-07428-4}
\end{barticle}
\endbibitem

\bibitem[\protect\citeauthoryear{Zhu et~al.}{2025}]{Zhu2025}
\begin{barticle}
\bauthor{\bsnm{Zhu}, \binits{Z.}},
\bauthor{\bsnm{Wang}, \binits{Z.}},
\bauthor{\bsnm{Dong}, \binits{C.}},
\bauthor{\bsnm{Yu}, \binits{M.}},
\bauthor{\bsnm{Xie}, \binits{H.}},
\bauthor{\bsnm{Cao}, \binits{X.}},
\bauthor{\bsnm{Han}, \binits{L.}},
\bauthor{\bsnm{Qi}, \binits{J.}}:
\batitle{Physics informed neural network modelling for storm surge forecasting ‚Äî a case study in the {Bohai Sea}, {China}}.
\bjtitle{Coastal Engineering}
\bvolume{197},
\bfpage{104686}
(\byear{2025})
\doiurl{10.1016/j.coastaleng.2024.104686}
\end{barticle}
\endbibitem

\bibitem[\protect\citeauthoryear{Sreeraj et~al.}{2025}]{Sreeraj2025}
\begin{barticle}
\bauthor{\bsnm{Sreeraj}, \binits{P.}},
\bauthor{\bsnm{Swapna}, \binits{P.}},
\bauthor{\bsnm{Singh}, \binits{M.}},
\bauthor{\bsnm{Krishnan}, \binits{R.}}:
\batitle{Improved storm surge prediction and extreme sea level future projections in the indian ocean using deep learning}.
\bjtitle{Environmental Research Letters}
\bvolume{20}(\bissue{8}),
\bfpage{084058}
(\byear{2025})
\doiurl{10.1088/1748-9326/ade9e0}
\end{barticle}
\endbibitem

\bibitem[\protect\citeauthoryear{Huang et~al.}{2024}]{Huang2024}
\begin{barticle}
\bauthor{\bsnm{Huang}, \binits{S.}},
\bauthor{\bsnm{Nie}, \binits{H.}},
\bauthor{\bsnm{Jiao}, \binits{J.}},
\bauthor{\bsnm{Chen}, \binits{H.}},
\bauthor{\bsnm{Xie}, \binits{Z.}}:
\batitle{Tidal level prediction model based on {VMD-LSTM} neural network}.
\bjtitle{Water}
\bvolume{16}(\bissue{17}),
\bfpage{2452}
(\byear{2024})
\doiurl{10.3390/w16172452}
\end{barticle}
\endbibitem

\bibitem[\protect\citeauthoryear{Shi et~al.}{2024}]{Shi2024}
\begin{barticle}
\bauthor{\bsnm{Shi}, \binits{X.}},
\bauthor{\bsnm{Chen}, \binits{P.}},
\bauthor{\bsnm{Ye}, \binits{Z.}},
\bauthor{\bsnm{Zhang}, \binits{X.}},
\bauthor{\bsnm{Wang}, \binits{W.}}:
\batitle{Tide level prediction during typhoons based on variable topology in graph convolution recurrent neural networks}.
\bjtitle{Ocean Engineering}
\bvolume{312},
\bfpage{119228}
(\byear{2024})
\doiurl{10.1016/j.oceaneng.2024.119228}
\end{barticle}
\endbibitem

\bibitem[\protect\citeauthoryear{Pachev et~al.}{2023}]{Pachev2023}
\begin{barticle}
\bauthor{\bsnm{Pachev}, \binits{B.}},
\bauthor{\bsnm{Arora}, \binits{P.}},
\bauthor{\bsnm{del-Castillo-Negrete}, \binits{C.}},
\bauthor{\bsnm{Valseth}, \binits{E.}},
\bauthor{\bsnm{Dawson}, \binits{C.}}:
\batitle{A framework for flexible peak storm surge prediction}.
\bjtitle{Coastal Engineering}
\bvolume{186},
\bfpage{104406}
(\byear{2023})
\doiurl{10.1016/j.coastaleng.2023.104406}
\end{barticle}
\endbibitem

\bibitem[\protect\citeauthoryear{Dotse et~al.}{2023}]{Dotse2023}
\begin{barticle}
\bauthor{\bsnm{Dotse}, \binits{S.-Q.}},
\bauthor{\bsnm{Larbi}, \binits{I.}},
\bauthor{\bsnm{Limantol}, \binits{A.M.}},
\bauthor{\bsnm{De~Silva}, \binits{L.C.}}:
\batitle{A review of the application of hybrid machine learning models to improve rainfall prediction}.
\bjtitle{Modeling Earth Systems and Environment}
\bvolume{10}(\bissue{1}),
\bfpage{19}--\blpage{44}
(\byear{2023})
\doiurl{10.1007/s40808-023-01835-x}
\end{barticle}
\endbibitem

\bibitem[\protect\citeauthoryear{Wei et~al.}{2025}]{Wei2025}
\begin{botherref}
\oauthor{\bsnm{Wei}, \binits{C.}},
\oauthor{\bsnm{Zhao}, \binits{X.}},
\oauthor{\bsnm{Liu}, \binits{Y.}},
\oauthor{\bsnm{Yang}, \binits{P.}},
\oauthor{\bsnm{Zhou}, \binits{Z.}},
\oauthor{\bsnm{Chen}, \binits{Y.}}:
Bias analysis and correction of {ERA5} reanalysis in the context of tropical cyclones.
Journal of Geophysical Research: Atmospheres
\textbf{130}(2)
(2025)
\doiurl{10.1029/2024jd042737}
\end{botherref}
\endbibitem

\bibitem[\protect\citeauthoryear{Li et~al.}{2025}]{Li2025}
\begin{barticle}
\bauthor{\bsnm{Li}, \binits{R.}},
\bauthor{\bsnm{Guilloteau}, \binits{C.}},
\bauthor{\bsnm{Foufoula-Georgiou}, \binits{E.}}:
\batitle{Added value of environmental variables for satellite precipitation retrieval: A temporal coevolution perspective and a machine learning integration assessment}.
\bjtitle{Geophysical Research Letters}
\bvolume{52}(\bissue{11}),
\bfpage{2025}--\blpage{116048}
(\byear{2025})
\end{barticle}
\endbibitem

\bibitem[\protect\citeauthoryear{Cerrone et~al.}{2025}]{Cerrone2025}
\begin{barticle}
\bauthor{\bsnm{Cerrone}, \binits{A.R.}},
\bauthor{\bsnm{Westerink}, \binits{L.G.}},
\bauthor{\bsnm{Ling}, \binits{G.}},
\bauthor{\bsnm{Blakely}, \binits{C.P.}},
\bauthor{\bsnm{Wirasaet}, \binits{D.}},
\bauthor{\bsnm{Dawson}, \binits{C.}},
\bauthor{\bsnm{Westerink}, \binits{J.J.}}:
\batitle{Correcting physics-based global tide and storm water level forecasts with the temporal fusion transformer}.
\bjtitle{Ocean Modelling}
\bvolume{195},
\bfpage{102509}
(\byear{2025})
\doiurl{10.1016/j.ocemod.2025.102509}
\end{barticle}
\endbibitem

\bibitem[\protect\citeauthoryear{Carneiro-Barros et~al.}{2025}]{CarneiroBarros2025}
\begin{barticle}
\bauthor{\bsnm{Carneiro-Barros}, \binits{J.E.}},
\bauthor{\bsnm{Majidi}, \binits{A.G.}},
\bauthor{\bsnm{Plomaritis}, \binits{T.}},
\bauthor{\bsnm{Fazeres-Ferradosa}, \binits{T.}},
\bauthor{\bsnm{Rosa-Santos}, \binits{P.}},
\bauthor{\bsnm{Taveira-Pinto}, \binits{F.}}:
\batitle{Coastal flooding hazards in northern {Portugal}: A practical large-scale evaluation of total water levels and swash regimes}.
\bjtitle{Water}
\bvolume{17}(\bissue{10}),
\bfpage{1478}
(\byear{2025})
\doiurl{10.3390/w17101478}
\end{barticle}
\endbibitem

\bibitem[\protect\citeauthoryear{Giaremis et~al.}{2024}]{giaremis2024storm}
\begin{barticle}
\bauthor{\bsnm{Giaremis}, \binits{S.}},
\bauthor{\bsnm{Nader}, \binits{N.}},
\bauthor{\bsnm{Dawson}, \binits{C.}},
\bauthor{\bsnm{Kaiser}, \binits{C.}},
\bauthor{\bsnm{Nikidis}, \binits{E.}},
\bauthor{\bsnm{Kaiser}, \binits{H.}}:
\batitle{Storm surge modeling in the {AI} era: Using {LSTM}-based machine learning for enhancing forecasting accuracy}.
\bjtitle{Coastal Engineering}
\bvolume{191},
\bfpage{104532}
(\byear{2024})
\end{barticle}
\endbibitem

\bibitem[\protect\citeauthoryear{Tedesco et~al.}{2024}]{Tedesco2024}
\begin{barticle}
\bauthor{\bsnm{Tedesco}, \binits{P.}},
\bauthor{\bsnm{Rabault}, \binits{J.}},
\bauthor{\bsnm{S√¶tra}, \binits{M.L.}},
\bauthor{\bsnm{Kristensen}, \binits{N.M.}},
\bauthor{\bsnm{Aarnes}, \binits{O.J.}},
\bauthor{\bsnm{Breivik}, \binits{√.}},
\bauthor{\bsnm{Mauritzen}, \binits{C.}},
\bauthor{\bsnm{S√¶tra}, \binits{√.}}:
\batitle{Bias correction of operational storm surge forecasts using neural networks}.
\bjtitle{Ocean Modelling}
\bvolume{188},
\bfpage{102334}
(\byear{2024})
\doiurl{10.1016/j.ocemod.2024.102334}
\end{barticle}
\endbibitem

\bibitem[\protect\citeauthoryear{Liao et~al.}{2024}]{Liao2024}
\begin{barticle}
\bauthor{\bsnm{Liao}, \binits{J.}},
\bauthor{\bsnm{Li}, \binits{Y.}},
\bauthor{\bsnm{Li}, \binits{J.}},
\bauthor{\bsnm{Li}, \binits{S.}},
\bauthor{\bsnm{Peng}, \binits{S.}}:
\batitle{A two-module bias-correction model for sea wave hindcasting based on the long-short term memory neural network}.
\bjtitle{Ocean Engineering}
\bvolume{311},
\bfpage{118827}
(\byear{2024})
\doiurl{10.1016/j.oceaneng.2024.118827}
\end{barticle}
\endbibitem

\bibitem[\protect\citeauthoryear{Zhang et~al.}{2024a}]{Zhang2024}
\begin{botherref}
\oauthor{\bsnm{Zhang}, \binits{S.}},
\oauthor{\bsnm{Harrop}, \binits{B.}},
\oauthor{\bsnm{Leung}, \binits{L.R.}},
\oauthor{\bsnm{Charalampopoulos}, \binits{A.}},
\oauthor{\bsnm{Barthel~Sorensen}, \binits{B.}},
\oauthor{\bsnm{Xu}, \binits{W.}},
\oauthor{\bsnm{Sapsis}, \binits{T.}}:
A machine learning bias correction on large‚Äêscale environment of high‚Äêimpact weather systems in {E3SM} atmosphere model.
Journal of Advances in Modeling Earth Systems
\textbf{16}(8)
(2024)
\doiurl{10.1029/2023ms004138}
\end{botherref}
\endbibitem

\bibitem[\protect\citeauthoryear{Zhang et~al.}{2024b}]{Zhang2024b}
\begin{barticle}
\bauthor{\bsnm{Zhang}, \binits{W.}},
\bauthor{\bsnm{Sun}, \binits{Y.}},
\bauthor{\bsnm{Wu}, \binits{Y.}},
\bauthor{\bsnm{Dong}, \binits{J.}},
\bauthor{\bsnm{Song}, \binits{X.}},
\bauthor{\bsnm{Gao}, \binits{Z.}},
\bauthor{\bsnm{Pang}, \binits{R.}},
\bauthor{\bsnm{Guoan}, \binits{B.}}:
\batitle{A deep-learning real-time bias correction method for significant wave height forecasts in the {Western North Pacific}}.
\bjtitle{Ocean Modelling}
\bvolume{187},
\bfpage{102289}
(\byear{2024})
\doiurl{10.1016/j.ocemod.2023.102289}
\end{barticle}
\endbibitem

\bibitem[\protect\citeauthoryear{Kao et~al.}{2024}]{Kao2024}
\begin{barticle}
\bauthor{\bsnm{Kao}, \binits{Y.-C.}},
\bauthor{\bsnm{Tsou}, \binits{H.-E.}},
\bauthor{\bsnm{Chen}, \binits{C.-J.}}:
\batitle{Development of multi-source weighted-ensemble precipitation: Influence of bias correction based on recurrent convolutional neural networks}.
\bjtitle{Journal of Hydrology}
\bvolume{629},
\bfpage{130621}
(\byear{2024})
\doiurl{10.1016/j.jhydrol.2024.130621}
\end{barticle}
\endbibitem

\bibitem[\protect\citeauthoryear{Liu et~al.}{2023}]{Liu2023}
\begin{botherref}
\oauthor{\bsnm{Liu}, \binits{G.}},
\oauthor{\bsnm{Bracco}, \binits{A.}},
\oauthor{\bsnm{Brajard}, \binits{J.}}:
Systematic bias correction in ocean mesoscale forecasting using machine learning.
Journal of Advances in Modeling Earth Systems
\textbf{15}(11)
(2023)
\doiurl{10.1029/2022ms003426}
\end{botherref}
\endbibitem

\bibitem[\protect\citeauthoryear{Yan et~al.}{2018}]{Yan2018}
\begin{botherref}
\oauthor{\bsnm{Yan}, \binits{S.}},
\oauthor{\bsnm{Xiong}, \binits{Y.}},
\oauthor{\bsnm{Lin}, \binits{D.}}:
Spatial Temporal Graph Convolutional Networks for Skeleton-Based Action Recognition.
arXiv
(2018).
\doiurl{10.48550/ARXIV.1801.07455} .
\url{https://arxiv.org/abs/1801.07455}
\end{botherref}
\endbibitem

\bibitem[\protect\citeauthoryear{Zhang et~al.}{2019}]{Zhang2019b}
\begin{barticle}
\bauthor{\bsnm{Zhang}, \binits{C.}},
\bauthor{\bsnm{Yu}, \binits{J.J.Q.}},
\bauthor{\bsnm{Liu}, \binits{Y.}}:
\batitle{Spatial-temporal graph attention networks: A deep learning approach for traffic forecasting}.
\bjtitle{IEEE Access}
\bvolume{7},
\bfpage{166246}--\blpage{166256}
(\byear{2019})
\doiurl{10.1109/access.2019.2953888}
\end{barticle}
\endbibitem

\bibitem[\protect\citeauthoryear{Jiang et~al.}{2024}]{Jiang2024}
\begin{barticle}
\bauthor{\bsnm{Jiang}, \binits{W.}},
\bauthor{\bsnm{Zhang}, \binits{J.}},
\bauthor{\bsnm{Li}, \binits{Y.}},
\bauthor{\bsnm{Zhang}, \binits{D.}},
\bauthor{\bsnm{Hu}, \binits{G.}},
\bauthor{\bsnm{Gao}, \binits{H.}},
\bauthor{\bsnm{Duan}, \binits{Z.}}:
\batitle{Advancing storm surge forecasting from scarce observation data: A causal-inference based spatio-temporal graph neural network approach}.
\bjtitle{Coastal Engineering}
\bvolume{190},
\bfpage{104512}
(\byear{2024})
\doiurl{10.1016/j.coastaleng.2024.104512}
\end{barticle}
\endbibitem

\bibitem[\protect\citeauthoryear{Kazadi et~al.}{2024}]{Kazadi2024}
\begin{botherref}
\oauthor{\bsnm{Kazadi}, \binits{A.}},
\oauthor{\bsnm{Doss-Gollin}, \binits{J.}},
\oauthor{\bsnm{Sebastian}, \binits{A.}},
\oauthor{\bsnm{Silva}, \binits{A.}}:
{FloodGNN}-{GRU}: a spatio-temporal graph neural network for flood prediction.
Environmental Data Science
\textbf{3}
(2024)
\doiurl{10.1017/eds.2024.19}
\end{botherref}
\endbibitem

\bibitem[\protect\citeauthoryear{Lam et~al.}{2023}]{Lam2023}
\begin{barticle}
\bauthor{\bsnm{Lam}, \binits{R.}},
\bauthor{\bsnm{Sanchez-Gonzalez}, \binits{A.}},
\bauthor{\bsnm{Willson}, \binits{M.}},
\bauthor{\bsnm{Wirnsberger}, \binits{P.}},
\bauthor{\bsnm{Fortunato}, \binits{M.}},
\bauthor{\bsnm{Alet}, \binits{F.}},
\bauthor{\bsnm{Ravuri}, \binits{S.}},
\bauthor{\bsnm{Ewalds}, \binits{T.}},
\bauthor{\bsnm{Eaton-Rosen}, \binits{Z.}},
\bauthor{\bsnm{Hu}, \binits{W.}},
\bauthor{\bsnm{Merose}, \binits{A.}},
\bauthor{\bsnm{Hoyer}, \binits{S.}},
\bauthor{\bsnm{Holland}, \binits{G.}},
\bauthor{\bsnm{Vinyals}, \binits{O.}},
\bauthor{\bsnm{Stott}, \binits{J.}},
\bauthor{\bsnm{Pritzel}, \binits{A.}},
\bauthor{\bsnm{Mohamed}, \binits{S.}},
\bauthor{\bsnm{Battaglia}, \binits{P.}}:
\batitle{Learning skillful medium-range global weather forecasting}.
\bjtitle{Science}
\bvolume{382}(\bissue{6677}),
\bfpage{1416}--\blpage{1421}
(\byear{2023})
\doiurl{10.1126/science.adi2336}
\end{barticle}
\endbibitem

\bibitem[\protect\citeauthoryear{Wu et~al.}{2023}]{Wu2023}
\begin{botherref}
\oauthor{\bsnm{Wu}, \binits{B.}},
\oauthor{\bsnm{Chen}, \binits{W.}},
\oauthor{\bsnm{Wang}, \binits{W.}},
\oauthor{\bsnm{Peng}, \binits{B.}},
\oauthor{\bsnm{Sun}, \binits{L.}},
\oauthor{\bsnm{Chen}, \binits{L.}}:
WeatherGNN: Exploiting Meteo- and Spatial-Dependencies for Local Numerical Weather Prediction Bias-Correction.
arXiv
(2023).
\doiurl{10.48550/ARXIV.2310.05517} .
\url{https://arxiv.org/abs/2310.05517}
\end{botherref}
\endbibitem

\bibitem[\protect\citeauthoryear{Nader et~al.}{2015}]{nader2015classification}
\begin{bchapter}
\bauthor{\bsnm{Nader}, \binits{N.}},
\bauthor{\bsnm{Hassan}, \binits{M.}},
\bauthor{\bsnm{Falou}, \binits{W.}},
\bauthor{\bsnm{Diab}, \binits{A.}},
\bauthor{\bsnm{Al-Omar}, \binits{S.}},
\bauthor{\bsnm{Khalil}, \binits{M.}},
\bauthor{\bsnm{Marque}, \binits{C.}}:
\bctitle{Classification of pregnancy and labor contractions using a graph theory based analysis}.
In: \bbtitle{2015 37th Annual International Conference of the IEEE Engineering in Medicine and Biology Society (EMBC)},
pp. \bfpage{2876}--\blpage{2879}
(\byear{2015}).
\bcomment{IEEE}
\end{bchapter}
\endbibitem

\bibitem[\protect\citeauthoryear{Nader et~al.}{2016}]{nader2016node}
\begin{bchapter}
\bauthor{\bsnm{Nader}, \binits{N.}},
\bauthor{\bsnm{Hassan}, \binits{M.}},
\bauthor{\bsnm{Falou}, \binits{W.}},
\bauthor{\bsnm{Marque}, \binits{C.}},
\bauthor{\bsnm{Khalil}, \binits{M.}}:
\bctitle{A node-wise analysis of the uterine muscle networks for pregnancy monitoring}.
In: \bbtitle{2016 38th Annual International Conference of the IEEE Engineering in Medicine and Biology Society (EMBC)},
pp. \bfpage{712}--\blpage{715}
(\byear{2016}).
\bcomment{IEEE}
\end{bchapter}
\endbibitem

\bibitem[\protect\citeauthoryear{Al-Omar et~al.}{2015}]{al2015detecting}
\begin{bchapter}
\bauthor{\bsnm{Al-Omar}, \binits{S.}},
\bauthor{\bsnm{Diab}, \binits{A.}},
\bauthor{\bsnm{Nader}, \binits{N.}},
\bauthor{\bsnm{Khalil}, \binits{M.}},
\bauthor{\bsnm{Karlsson}, \binits{B.}},
\bauthor{\bsnm{Marque}, \binits{C.}}:
\bctitle{Detecting labor using graph theory on connectivity matrices of uterine emg}.
In: \bbtitle{2015 37th Annual International Conference of the IEEE Engineering in Medicine and Biology Society (EMBC)},
pp. \bfpage{2195}--\blpage{2198}
(\byear{2015}).
\bcomment{IEEE}
\end{bchapter}
\endbibitem

\bibitem[\protect\citeauthoryear{Pedregosa et~al.}{2011}]{pedregosa2011scikit}
\begin{barticle}
\bauthor{\bsnm{Pedregosa}, \binits{F.}},
\bauthor{\bsnm{Varoquaux}, \binits{G.}},
\bauthor{\bsnm{Gramfort}, \binits{A.}},
\bauthor{\bsnm{Michel}, \binits{V.}},
\bauthor{\bsnm{Thirion}, \binits{B.}},
\bauthor{\bsnm{Grisel}, \binits{O.}},
\bauthor{\bsnm{Blondel}, \binits{M.}},
\bauthor{\bsnm{Prettenhofer}, \binits{P.}},
\bauthor{\bsnm{Weiss}, \binits{R.}},
\bauthor{\bsnm{Dubourg}, \binits{V.}}, \betal:
\batitle{Scikit-learn: Machine learning in python}.
\bjtitle{the Journal of machine Learning research}
\bvolume{12},
\bfpage{2825}--\blpage{2830}
(\byear{2011})
\end{barticle}
\endbibitem

\bibitem[\protect\citeauthoryear{G{\'e}ron}{2022}]{geron2022hands}
\begin{bbook}
\bauthor{\bsnm{G{\'e}ron}, \binits{A.}}:
\bbtitle{Hands-on Machine Learning with Scikit-Learn, Keras, and TensorFlow}.
\bpublisher{" O'Reilly Media, Inc."}, \blocation{???}
(\byear{2022})
\end{bbook}
\endbibitem

\bibitem[\protect\citeauthoryear{}{}]{SaffirSimpson}
\begin{botherref}
{S}affir-{S}impson {H}urricane {S}cale --- weather.gov.
\url{https://www.weather.gov/mfl/saffirsimpson}
\end{botherref}
\endbibitem

\bibitem[\protect\citeauthoryear{Vaswani et~al.}{2017}]{vaswani2017attention}
\begin{botherref}
\oauthor{\bsnm{Vaswani}, \binits{A.}},
\oauthor{\bsnm{Shazeer}, \binits{N.}},
\oauthor{\bsnm{Parmar}, \binits{N.}},
\oauthor{\bsnm{Uszkoreit}, \binits{J.}},
\oauthor{\bsnm{Jones}, \binits{L.}},
\oauthor{\bsnm{Gomez}, \binits{A.N.}},
\oauthor{\bsnm{Kaiser}, \binits{{\L}.}},
\oauthor{\bsnm{Polosukhin}, \binits{I.}}:
Attention is all you need.
Advances in neural information processing systems
\textbf{30}
(2017)
\end{botherref}
\endbibitem

\bibitem[\protect\citeauthoryear{Kingma and Ba}{2014}]{Kingma2014}
\begin{botherref}
\oauthor{\bsnm{Kingma}, \binits{D.P.}},
\oauthor{\bsnm{Ba}, \binits{J.}}:
Adam: A Method for Stochastic Optimization.
arXiv
(2014).
\doiurl{10.48550/ARXIV.1412.6980}
\end{botherref}
\endbibitem

\bibitem[\protect\citeauthoryear{Cangialosi and Alaka}{2023}]{Cangialosi2023}
\begin{botherref}
\oauthor{\bsnm{Cangialosi}, \binits{J.P.}},
\oauthor{\bsnm{Alaka}, \binits{L.}}:
National Hurricane Center Tropical Cyclone Report - Hurricane Idalia (AL102023), 26‚Äì31 August 2023.
\url{https://www.nhc.noaa.gov/data/tcr/AL102023_Idalia.pdf}.
National Hurricane Center Tropical Cyclone Report
(2023)
\end{botherref}
\endbibitem

\bibitem[\protect\citeauthoryear{{NHC - National Hurricane Center and Central Pacific National Center}}{2025}]{NHC}
\begin{botherref}
\oauthor{\bsnm{{NHC - National Hurricane Center and Central Pacific National Center}}}:
{NHC} Active Tropical Cyclones.
\url{https://www.nhc.noaa.gov/cyclones/}
(2025)
\end{botherref}
\endbibitem

\bibitem[\protect\citeauthoryear{}{}]{NOAAInundationHistory}
\begin{botherref}
{I}nundation {H}istory - {N}{O}{A}{A} {T}ides \& {C}urrents --- tidesandcurrents.noaa.gov.
\url{https://tidesandcurrents.noaa.gov/inundationdb/inundation.html}.
[Accessed 15-10-2025]
\end{botherref}
\endbibitem

\bibitem[\protect\citeauthoryear{{D}epartment~of {C}ommerce}{2025}]{NOAA2017}
\begin{botherref}
\oauthor{\bsnm{{C}ommerce}, \binits{N.W.S.} \bsuffix{{N}ational {O}ceanic \& {A}tmospheric~{A}dministration}}:
{N}ational {W}eather {S}ervice {I}nstruction 10-320. {S}urf {Z}one {F}orecast and {C}oastal/{L}akeshore {H}azard {S}ervices.
\url{https://www.weather.gov/media/directives/010_pdfs/pd01003020curr.pdf}.
[Accessed 16-10-2025]
(2025)
\end{botherref}
\endbibitem

\end{thebibliography}


\end{document}